\documentclass[10pt,journal,compsoc]{IEEEtran}

\usepackage{cite}
\usepackage{url}
\usepackage{amssymb}
\usepackage{subcaption}
\usepackage{booktabs}
\makeindex

\usepackage[pagebackref,breaklinks,colorlinks]{hyperref}

\usepackage{graphicx}
\DeclareGraphicsExtensions{.pdf,.jpg,.png}

\usepackage[pagebackref,breaklinks,colorlinks]{hyperref}
\usepackage{overpic}

\usepackage[table]{xcolor}

\usepackage{silence}
\newcommand{\tablestyle}[2]{\small\setlength{\tabcolsep}{#1}\renewcommand{\arraystretch}{#2}\centering}

\usepackage{xspace}
\makeatletter
\DeclareRobustCommand\onedot{\futurelet\@let@token\@onedot}
\def\@onedot{\ifx\@let@token.\else.\null\fi\xspace}

\def\eg{\emph{e.g}\onedot} 
\def\ie{\emph{i.e}\onedot}

\def\etal{\emph{et al}\onedot}

\makeatother

\def\sArt{state-of-the-art~}
\def\RNumber{7~}
\def\SNumber{33~}

\usepackage{geometry}
\newtheorem{assumption}{Assumption}

\renewcommand{\tablename}{Table }

\newcommand{\figref}[1]{Fig.~\ref{#1}}
\newcommand{\tabref}[1]{Tab.~\ref{#1}}
\newcommand{\equref}[1]{(\ref{#1})}
\newcommand{\secref}[1]{Sec.~\ref{#1}}

\newcommand{\myPara}[1]{\vspace{.05in}\noindent\textbf{#1.}}

\graphicspath{{./Imgs/}}

\begin{document}

\title{MS-NeRF: Multi-Space Neural Radiance Fields}

\author{Ze-Xin~Yin, Peng-Yi~Jiao, Jiaxiong~Qiu, 
    Ming-Ming~Cheng, %~\IEEEmembership{Senior Member,~IEEE,}
    and~Bo~Ren %~\IEEEmembership{Member,~IEEE}%
    \IEEEcompsocitemizethanks{
      \IEEEcompsocthanksitem Z.X. Yin, P.Y. Jiao, M.M. Cheng and B. Ren are with the TBI Center \& VCIP,
        Nankai University.
      \IEEEcompsocthanksitem J. Qiu is with Horizon Robotics, Beijing, China.
      \IEEEcompsocthanksitem Bo Ren is the corresponding author (rb@nankai.edu.cn).
      \IEEEcompsocthanksitem A preliminary version of this work appeared at 
            CVPR~\cite{Yin_2023_CVPR}.
    }% <-this % stops a space   %    \thanks{}
}

\markboth{IEEE Transactions on Pattern Analysis and Machine Intelligence  \ \ \ \ \ \ \ \ \ \ \ \ \ \ \ \ \ \ \ \ \ \ \ \ \ \ \ \ \ \ \ \ \ \ \ 
\ \ \ \ \ \ \ \ \ \ \ \ \ \ \ \ \ \ \ \ \ \ \ \ \ \ \ \ \ \ \ \ \ \ \  \ \ \ \ \ 
DOI:~\href{https://doi.org/10.1109/TPAMI.2025.3540074}{10.1109/TPAMI.2025.3540074}\quad\quad\quad\quad}%
{Yin \MakeLowercase{\textit{et al.}}: MS-NeRF: Multi-Space Neural Radiance Fields}
%\pubid{0000--0000/00\$00.00~\copyright~2009 IEEE}

\IEEEcompsoctitleabstractindextext{%
  \begin{abstract}
    Existing Neural Radiance Fields (NeRF) methods suffer from 
    the existence of reflective objects, 
    often resulting in blurry or distorted rendering.
    Instead of calculating a single radiance field, 
    we propose a multi-space neural radiance field (MS-NeRF) that
    represents the scene using a group of feature fields 
    in parallel sub-spaces, 
    which leads to a better understanding of the neural network toward 
    the existence of reflective and refractive objects. 
    Our multi-space scheme works as an enhancement to existing NeRF methods, 
    with only small computational overheads needed for training and 
    inferring the extra-space outputs. 
    We design different multi-space modules for representative MLP-based and grid-based 
    NeRF methods, which improve Mip-NeRF 360 by 4.15 dB in PSNR with 0.5\% extra parameters
    and further improve TensoRF by 2.71 dB with 0.046\% extra parameters on reflective regions
    without degrading the rendering quality on other regions.
    We further construct a novel dataset consisting of 
    \SNumber synthetic scenes and \RNumber real captured scenes 
    with complex reflection and refraction,
    where we design complex camera paths 
    to fully benchmark the robustness of NeRF-based methods.
    Extensive experiments show that our approach significantly outperforms 
    the existing single-space NeRF methods for rendering high-quality scenes 
    concerned with complex light paths through mirror-like objects.
    The source code, dataset, and results are available via our project page:
    \url{https://zx-yin.github.io/msnerf/}.
  \end{abstract}

}

\maketitle

\IEEEdisplaynotcompsoctitleabstractindextext
\IEEEpeerreviewmaketitle

%%%%%%%%%%%%%%%%%intro%%%%%%%%%%%%%%%%%%
\section{Introduction}\label{sec:intro}

\IEEEPARstart{N}{eural} Radiance Fields (NeRF)~\cite{mildenhall2021nerf} 
and its variants 
are refreshing the community of neural rendering and 3D reconstruction, 
and the potential for more promising applications is still under exploration.
NeRF represents scenes as continuous radiance fields stored by 
simple Multi-layer Perceptrons~(MLPs) and renders novel views 
by integrating the densities and radiance, 
which are queried from the MLPs by points sampled along the ray 
from the camera to the image plane. 
Since its first presentation~\cite{mildenhall2021nerf}, 
many efforts have been investigated to enhance the method, 
such as extending to unbounded scenes~\cite{zhang2020nerf++,barron2022mip}, 
handling moving objects~\cite{pumarola2021d, park2021nerfies, tretschk2021non}, 
or reconstructing from pictures in the wild
\cite{martin2021nerf, chen2022hallucinated, zhang2021ners, sun2022neural}.

\newcommand{\addImg}[1]{\includegraphics[width=.49\linewidth]{360_mirror_through/#1}}

\begin{figure}[t]
  \centering
  \addImg{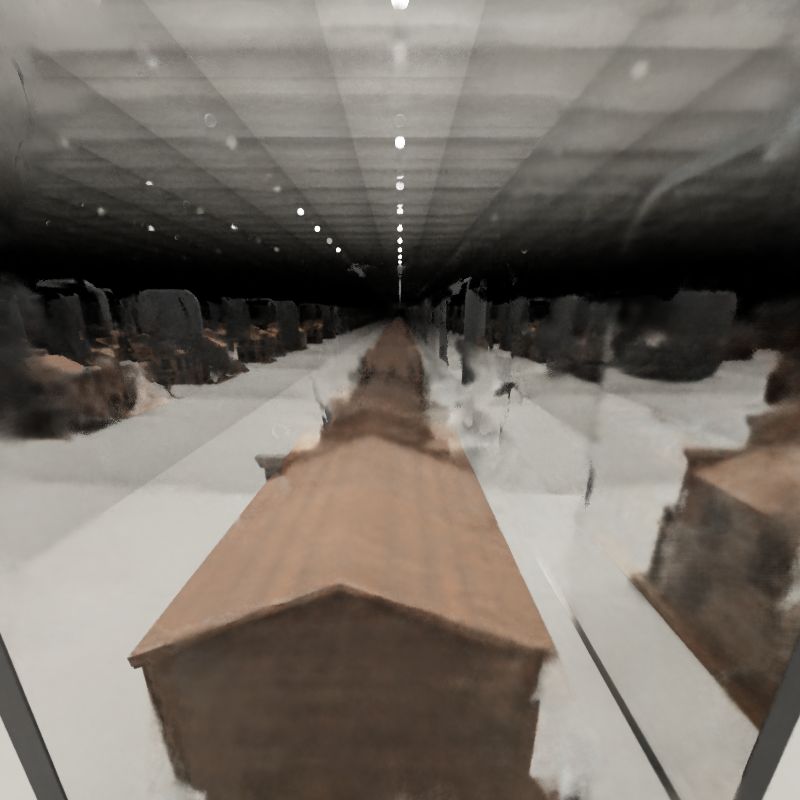} \hspace{-6pt} \addImg{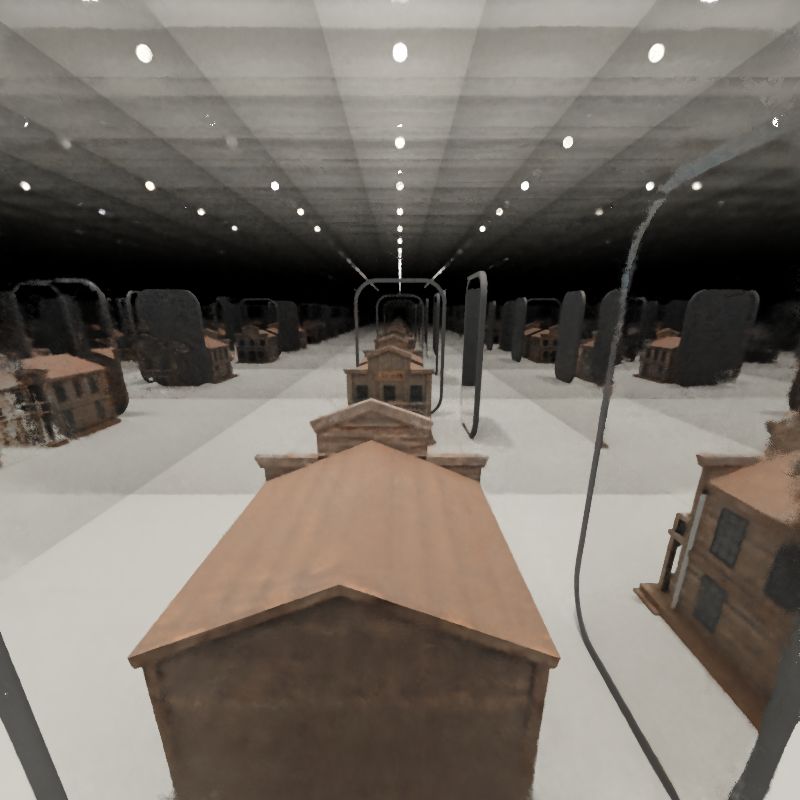} \\ \vspace{1pt}
  \addImg{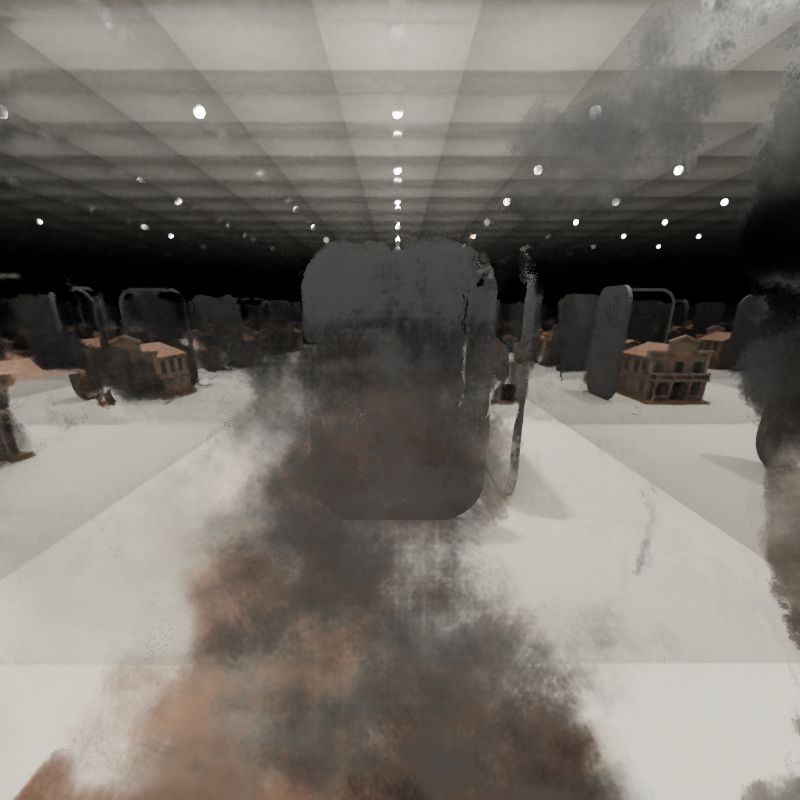} \hspace{-6pt} \addImg{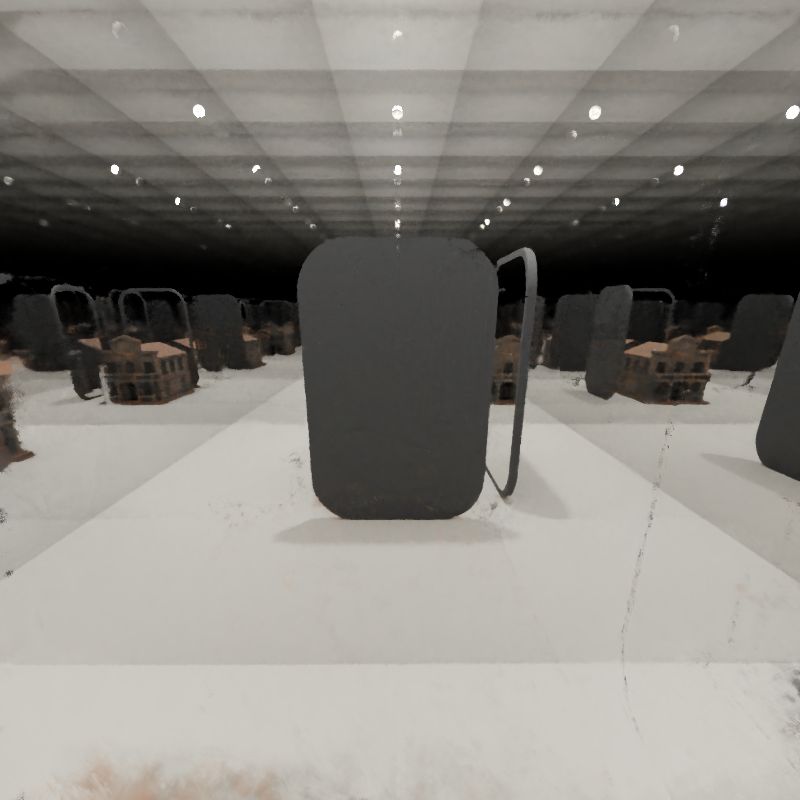} \\ 
  ~ (a) Mip-NeRF 360 \hspace{28pt} (b) MS-Mip-NeRF 360 \\ \vspace{-2pt}
  \caption{These are test views from the novel mirror-passing-through path.
    The first row is in front of the mirror, 
    while the last row is behind the mirror.
  }\label{fig:mirror_through}
\end{figure}

However, rendering scenes with mirrors is still a challenging task 
for \sArt NeRF-like methods. 
One of the principle assumptions for the NeRF method is the 
multi-view consistency property of the target scenes
\cite{jain2022robustifying, lin2021barf, tewari2022advances, bian2022nopenerf}.
When there are mirrors in the space, 
if one allows the viewpoints to move 360-degree around the scene, 
there is no consistency between the front and back views of a mirror, 
since the mirror surface and its reflected virtual image are only visible 
from a small range of views. 
As a result, it is often required to manually label the reflective surfaces 
in order to avoid falling into sub-optimal convergences~\cite{guo2022nerfren}.

In this paper, we propose a novel multi-space NeRF method 
to allow the automatic handling of mirror-like objects in the 360-degree 
high-fidelity rendering of scenes without any manual labeling. 
Instead of regarding the Euclidean scene space as a single space, 
we treat it as composed of multiple virtual sub-spaces, 
whose composition changes according to location and view direction. 
We show that our approach using multi-space decomposition 
leads to successful handlements of complex reflections and refractions
where the multi-view consistency is heavily violated in the 
Euclidean real space. 
Furthermore, we show that the above benefits can be achieved 
by designing a low-cost multi-space module and 
replacing the original output layer with it. 
Therefore, our multi-space approach serves as a general enhancement 
to the NeRF-based backbone, 
equipping most NeRF-like methods with the ability to model complex reflection 
and refraction, as shown in \figref{fig:mirror_through}. 

Existing datasets have not paid enough attention to the 360-degree rendering of 
scenes containing mirror-like objects, 
such as RFFR~\cite{guo2022nerfren} just has forward-facing scenes, 
and the Shiny dataset in~\cite{wizadwongsa2021nex} with small 
viewpoints changes and cannot exhibit view-dependent effects 
in large angle scale. 
Therefore, we construct a novel dataset dedicated to evaluation for 
the 360-degree high-fidelity rendering of scenes 
containing complex reflections and refractions. 
In this dataset, 
we collect \SNumber synthesized scenes and \RNumber captured real-world scenes. 
Each synthesized scene consists of 120 images captured in the 360-degree 
circle path around reflective or refractive objects, 
with 100 randomly split for training, 
10 for validation, and 10 for evaluation. 
Furthermore, we design more challenging paths for 10 of the synthesized scenes
to benchmark the robustness of NeRF-based methods, 
including 360-degree spiral paths where cameras gradually spiral up 
from the equator to the pole and novel mirror-passing-through paths 
where cameras move through the mirrors back and force,
each of which have 100 training views and 200 testing views 
following the convention of NeRF dataset~\cite{mildenhall2021nerf}.
Each real-world scene is captured randomly around scenes with reflective 
and refractive objects, 
consisting of 62 to 118 images, and organized under the convention of 
LLFF~\cite{mildenhall2019local}.

We then demonstrate the superiority of our approach 
by comparisons, using three representative baseline models, 
\ie, NeRF~\cite{mildenhall2021nerf}, 
Mip-NeRF~\cite{barron2021mip}, and Mip-NeRF 360~\cite{barron2022mip}, 
with and without our multi-space module. 
Besides, we investigate the grid-based acceleration methods, 
and we propose a hybrid multi-space module based on classic methods, 
\ie, TensoRF~\cite{Chen2022ECCV} and iNGP~\cite{InstantNGP},
to demonstrate the compatibility of our scheme.
3D reconstruction methods have a stronger dependence on multi-view consistency;
therefore, we experimentally integrate our multi-space module with 
a classic NeRF-based reconstruction method, \ie, NeuS~\cite{wang2021neus}, 
and the rendering results indicate
that our module is also beneficial to reconstruction methods.
Experiments show that our approach not only improves performance 
by a large margin on scenes with reflection and refraction
but also exhibits robustness on methods not specialized in rendering.
Our main contributions are as follows:
\begin{itemize}
  \item We propose a multi-space NeRF method that automatically 
    handles mirror-like objects in 360-degree high-fidelity scene rendering, 
    achieving significant improvements over the existing representative baselines
    both quantitatively and qualitatively.
  \item We design a lightweight module that can equip most NeRF-like
    methods with the ability to model reflection and refraction. 
  \item We propose a hybrid multi-space scheme for TensoRF and iNGP, exhibiting 
    the compatibility of our scheme with grid-based NeRF methods. 
  \item We construct a dataset dedicated to evaluation for 
    the 360-degree high-fidelity rendering of scenes containing complex 
    reflections and refractions, 
    including $33$ synthesized scenes and $7$ real captured scenes,
    with challenging camera paths.
\end{itemize}
%%%%%%%%%%%%%%%%%intro%%%%%%%%%%%%%%%%%%

%%%%%%%%%%%%%%%%%related%%%%%%%%%%%%%%%%%%
\section{Related Work}\label{sec:related}

\myPara{Coordinate-based novel view synthesis}
NeRF~\cite{mildenhall2021nerf} has bridged the gap between computer vision 
and computer graphics, 
and reveals a promising way to render high-quality photorealistic scenes 
with only posed images.
The insights and the generalization ability of this scheme also facilitate 
various tasks, \ie, 
3D reconstruction~\cite{wang2021neus, oechsle2021unisurf, yariv2021volume, Yu2022SDFStudio, yariv2023bakedsdf, li2023neuralangelo, Qiu_2023_CVPR}, 
3D-aware generation~\cite{chan2022efficient, niemeyer2021giraffe, jain2022zero, poole2022dreamfusion, wang2023prolificdreamer}, 
3D-aware edition~\cite{wang2022clip, yuan2022nerf, instructnerf2023}, 
and avatar reconstruction and manipulation
\cite{zheng2022avatar, jiang2022selfrecon, feng2022capturing, cao2023dreamavatar}.
Researchers have made great efforts to enhance this scheme.
Mip-NeRF~\cite{barron2021mip} enhances the anti-aliasing ability of NeRF 
by featuring 3D conical frustum using integrated positional encoding.
\cite{mildenhall2022nerf, huang2022hdr} adapt this scheme to HDR images.
\cite{zhang2020nerf++, barron2022mip} extend NeRF and its variants to 
unbounded scenes.
\cite{wang2021neus, oechsle2021unisurf, yariv2021volume, yariv2023bakedsdf, li2023neuralangelo} 
construct the relationship between the SDF and 
the density in volumetric rendering of NeRF for 3D reconstruction.
There are also many works trying to speed up the training and inference speed 
using explicit or hybrid representations
\cite{InstantNGP, yu2021plenoctrees, garbin2021fastnerf, reiser2021kilonerf, sun2022direct, chen2022mobilenerf, Chen2022ECCV}.

Glossy materials with high specular have a great influence on 
NeRF-like methods. 
\cite{verbin2022refnerf} is inspired by precomputation-based 
techniques~\cite{ramamoorthi2009precomputation} in computer graphics 
to represent and render view-dependent specular and reflection, 
but it fails to handle mirror-like reflective surfaces 
because the virtual images cannot be treated as textures.
Guo \etal~\cite{guo2022nerfren} propose to decompose reflective surfaces 
into a transmitted part and reflected part, 
which is the most relevant work to ours.
However, such decomposition cannot handle 360-degree views 
with mirror-like objects, 
because the virtual images have no difference from real objects 
until the viewpoint moves beyond a certain angle.
Zeng \etal~\cite{zeng2023mirror-nerf} incorporates the ray tracing scheme into 
NeRF to model the reflection, but there also lacks views behind the mirrors.

Another line of work similar to ours is multiple neural radiance fields, 
but they do so for different purposes~\cite{niemeyer2021giraffe, reiser2021kilonerf, yang2022recursive, yang2021learning, guo2020object}.
\cite{niemeyer2021giraffe} uses object-level neural radiance fields 
for 3D-aware generation and composition.
\cite{reiser2021kilonerf, yang2022recursive} uses multiple small MLPs 
for efficient rendering.
\cite{yang2021learning, guo2020object} uses multiple object-level 
neural radiance fields for 3D scene decomposition and edition.

\myPara{Commonly used datasets}
Researchers have introduced or constructed many different datasets 
to facilitate the development of NeRF-based methods in various tasks.
Mildenhall \etal~\cite{mildenhall2021nerf} collect a dataset containing 
eight rendered sets of posed images about eight objects separately, 
and eight real captured forward-facing scenes with the camera poses 
and intrinsics estimated by COLMAP~\cite{schonberger2016structure}.
Nevertheless, these scenes lack reflection and refraction, 
which are very common. 
Wizadwongsa \etal ~\cite{wizadwongsa2021nex} propose a dataset, namely 
Shiny, that contains eight more challenging scenes to test 
NeRF-like methods on view-dependent effects, 
but they are captured in a roughly forward-facing manner. 
Verbin \etal ~\cite{verbin2022refnerf} create a dataset of six glossy objects, 
namely Shiny Blender, 
which are rendered under similar conditions as done in NeRF
to test methods in modeling more complex materials.
For unbounded scenes, Barron \etal ~\cite{barron2022mip} construct 
a dataset consisting of 5 outdoor scenes and 4 indoor scenes, 
while Zhang \etal~\cite{zhang2020nerf++} adopt Tanks and Temples 
(T\&T) dataset~\cite{knapitsch2017tanks} and 
the Light Field dataset~\cite{yucer2016efficient}.
Bemana1 \etal~\cite{bemana2022eikonal} capture a dataset consisting 
of refractive objects, 
which is composed of four scenes with cameras moving in a large range.
Guo \etal~\cite{guo2022nerfren} collect six forward-facing scenes 
with reflective and semi-transparent materials, 
which is, to date, the most relevant dataset to ours, 
but ours is much more challenging.
DTU dataset~\cite{jensen2014large} and 
BlendedMVS dataset~\cite{yao2020blendedmvs} are commonly used as benchmarks 
for the evaluation of 3D reconstruction. 

%%%%%%%%%%%%%%%%%related%%%%%%%%%%%%%%%%%%

%%%%%%%%%%%%%%%%%method%%%%%%%%%%%%%%%%%%
\section{Method}
\label{sec:method}

\begin{figure}[t]
  \centering
  \includegraphics[width=0.9\linewidth]{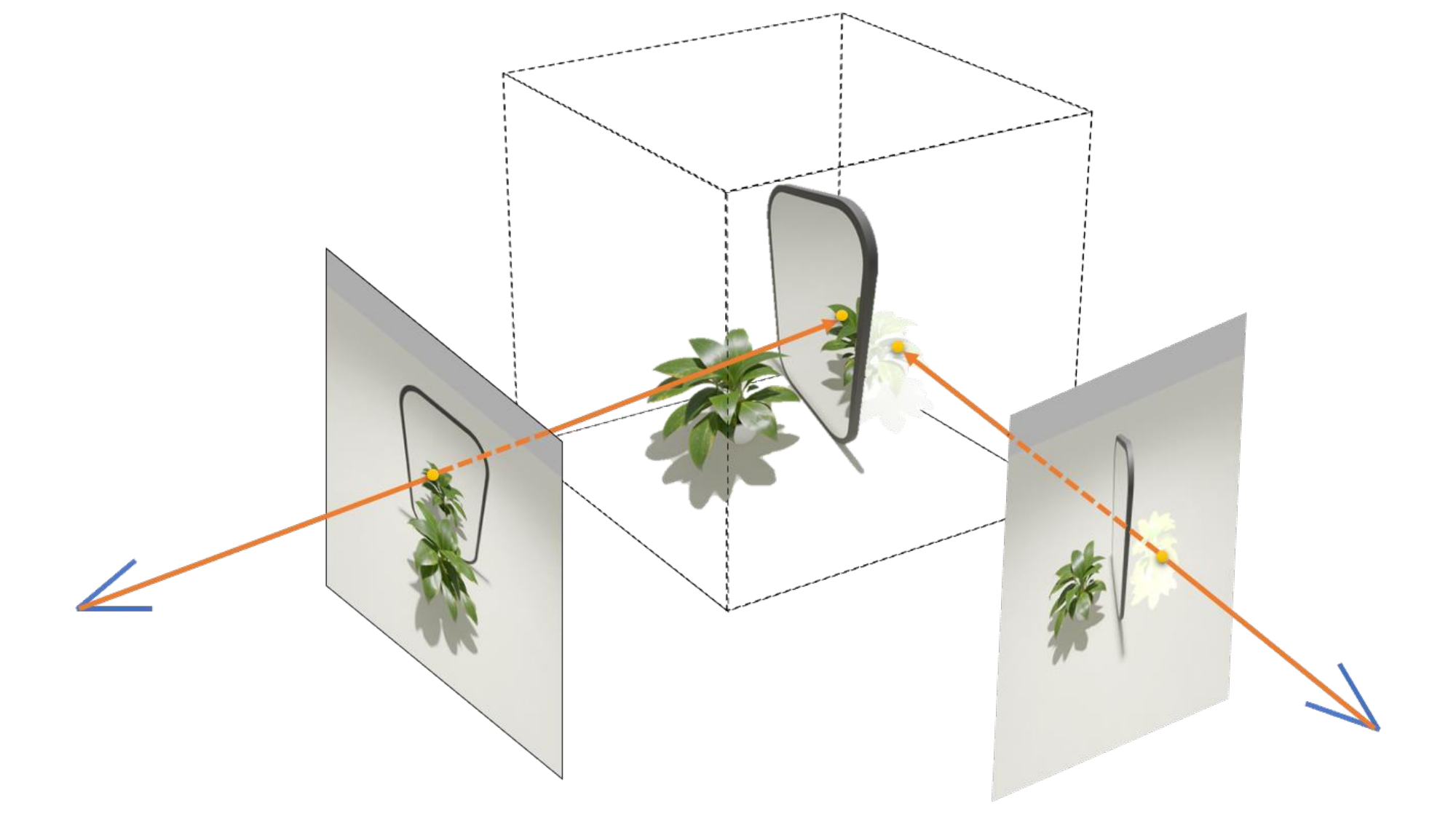}
  \vspace{-10pt}
  \caption{The virtual image created by the mirror is visible only in 
    a small range of views, 
    which violates the multi-view consistency.
  }
  \vspace{-5pt}
  \label{fig:multiview}
\end{figure}

\subsection{Preliminaries: Neural Radiance Fields}
\label{subsec:pre}

Neural Radiance Fields~(NeRF) \cite{mildenhall2021nerf} encodes a scene 
in the form of continuous volumetric fields into the weights of 
a multilayer perceptron~(MLP), 
and adopts the absorption-only model in the traditional volumetric rendering 
to synthesize novel views.
The training process only requires a sparse set of posed images
and casts rays $\mathbf{r}(t)=\mathbf{o}+t\mathbf{d}$ through the scene, 
where $\mathbf{o}\in \mathbb{R}^3$ is the camera center and 
$\mathbf{d}\in \mathbb{R}^3$ is the view direction, 
and the rays can be calculated by intrinsics and poses from the training data.
Given these rays, NeRF samples a set of 3D points 
$\{\mathbf{p}_i=\mathbf{o}+t_i\mathbf{d}\}$ by the distance to 
the camera $t_i$ in the Euclidean space and projects these points 
to a higher dimensional space using the following function:
\begin{equation}
  \gamma({\rm \mathbf{p}})=[{\rm sin}({\rm \mathbf{p}}), {\rm cos}({\rm \mathbf{p}}), ..., {\rm sin}(2^{L-1}{\rm \mathbf{p}}), {\rm cos}(2^{L-1}{\rm \mathbf{p}})]
  \label{eq:PosEnc}
\end{equation}
where $L$ is a hyperparameter and $\mathbf{p}$ is a sampled point. 

Given the projected features $\{\gamma({\rm \mathbf{p}_i})\}$ and 
the ray direction $\mathbf{d}$, the MLP outputs the densities $\{\sigma_i\}$ 
and colors $\{\mathbf{c}_i\}$, 
which are used to estimate the color $\mathbf{C}(\mathbf{r})$ of the ray 
using the quadrature rule reviewed by Max~\cite{max1995optical}:
\begin{equation}\label{eq:intergel}
  \hat{\mathbf{C}}(\mathbf{r})=\sum_{i=1}^N T_i(1-{\rm exp}(-\sigma_i \delta_i))\mathbf{c}_i    
\end{equation}
with $T_i = {\rm exp}(-\sum_{j=1}^{i-1}\sigma_j\delta_j)$ and 
$\delta_i = t_i - t_{i-1}$.
Since the equation is differentiable, 
the model parameters can be optimized directly by Mean Squared Error~(MSE) loss:
\begin{equation}\label{eq:loss}
  \mathcal{L} = \frac{1}{|\mathcal{R}|}\sum_{r\in \mathcal{R}}||\hat{\mathbf{C}}(\mathbf{r})-\mathbf{C}(\mathbf{r})||_2
\end{equation}
where $\mathcal{R}$ is a training batch of rays. 
Besides, NeRF also adopts a hierarchical sampling strategy to sample more 
points where higher weights are accumulated. 
With these designs, NeRF achieves \sArt photorealistic results of novel view 
synthesis in most cases.

\renewcommand{\addImg}[2]{
  \begin{subfigure}{0.49\linewidth}
    \centering
	  \includegraphics[width=\textwidth]{toy_scene/#1}
    \caption{#2}
    \label{fig:#1}
  \end{subfigure}%
}

\begin{figure}[t]
  \centering
  \addImg{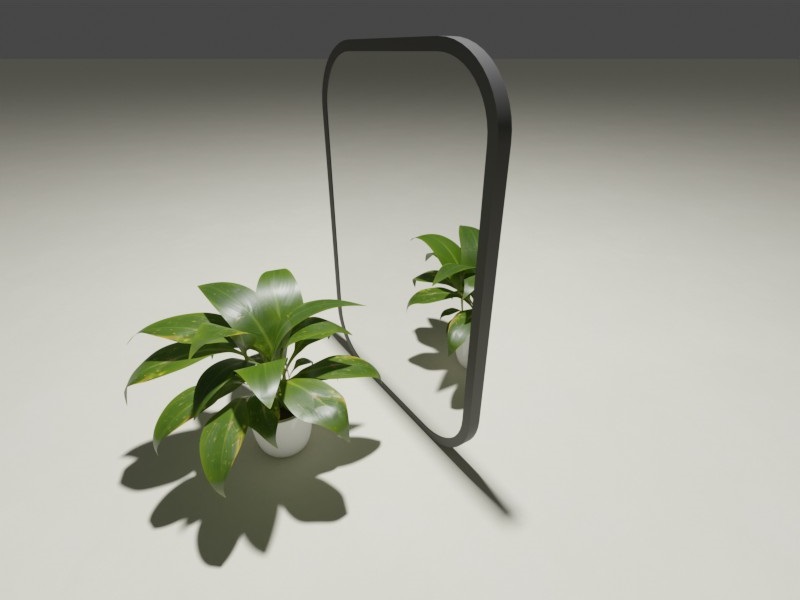}{Training view in scene A.} \hspace{-8pt}
  \addImg{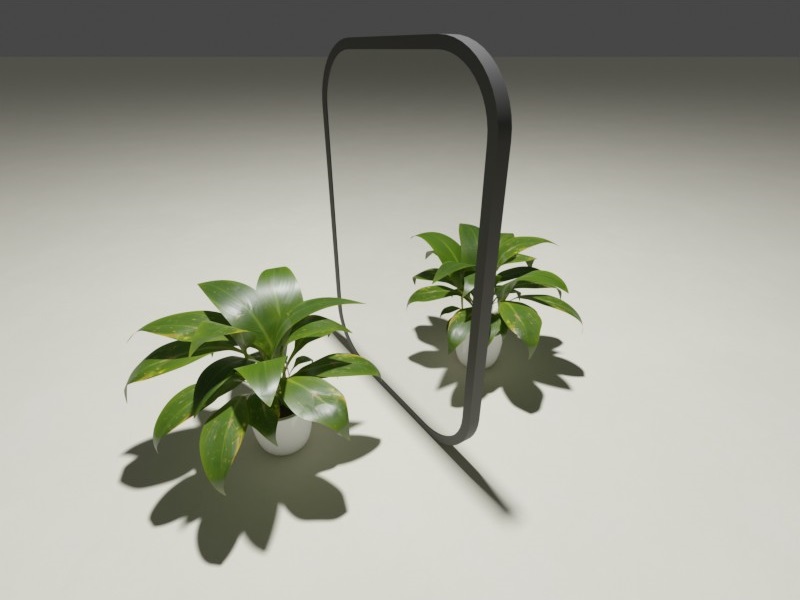}{Training view in scene B.} \\
  \addImg{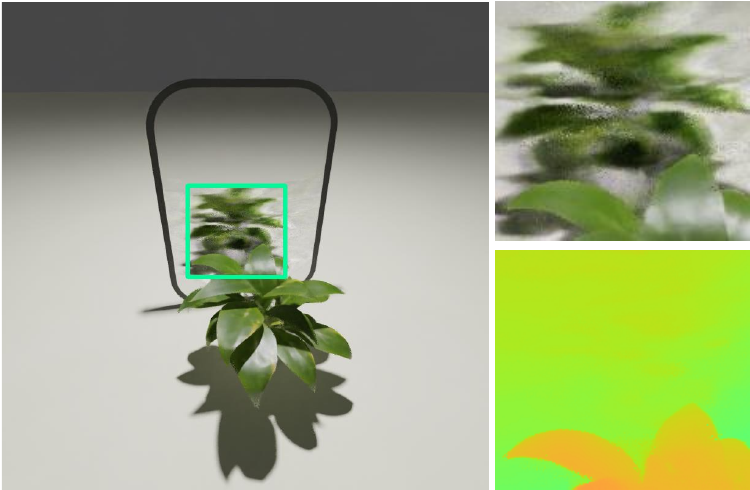}{Render view in scene A.} \hspace{-8pt}
  \addImg{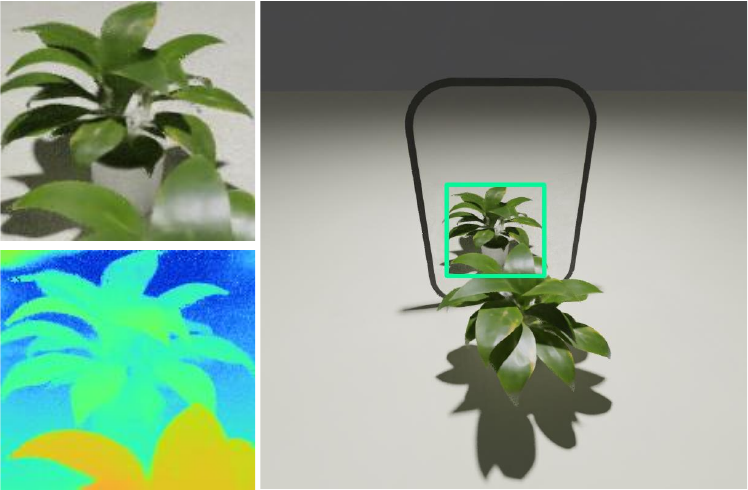}{Render view in scene B.}
  \caption{The first row is training view examples in the two scenes. 
    In scene A, there is only a plant in front of a mirror, 
    while in scene B we carefully place another plant to match the exact 
    position where the virtual image lies.
    The second row is test views with rendered depth from the 
    vanilla NeRF trained on the toy scenes. 
    As demonstrated, NeRF can avoid the trap of treating reflected images 
    as textures when the `virtual image' satisfies multi-view consistency.
  }\label{fig:toy}
\end{figure}

\begin{figure*}[t]
  \centering
  \begin{subfigure}{0.155\linewidth}
        \centering
    	  \includegraphics[width=\textwidth]{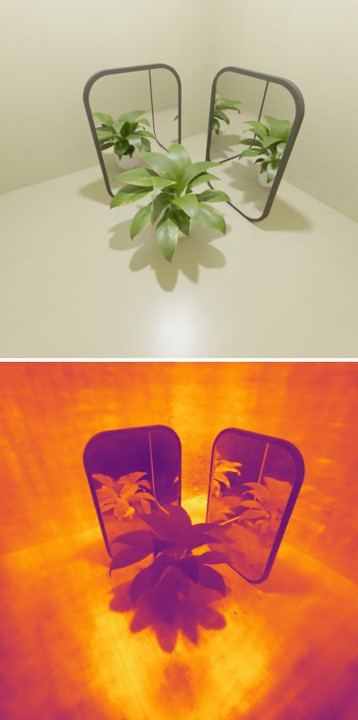}
  \end{subfigure}
  \hfill
  \begin{subfigure}{0.838\linewidth}
    \centering
    \includegraphics[width=\textwidth]{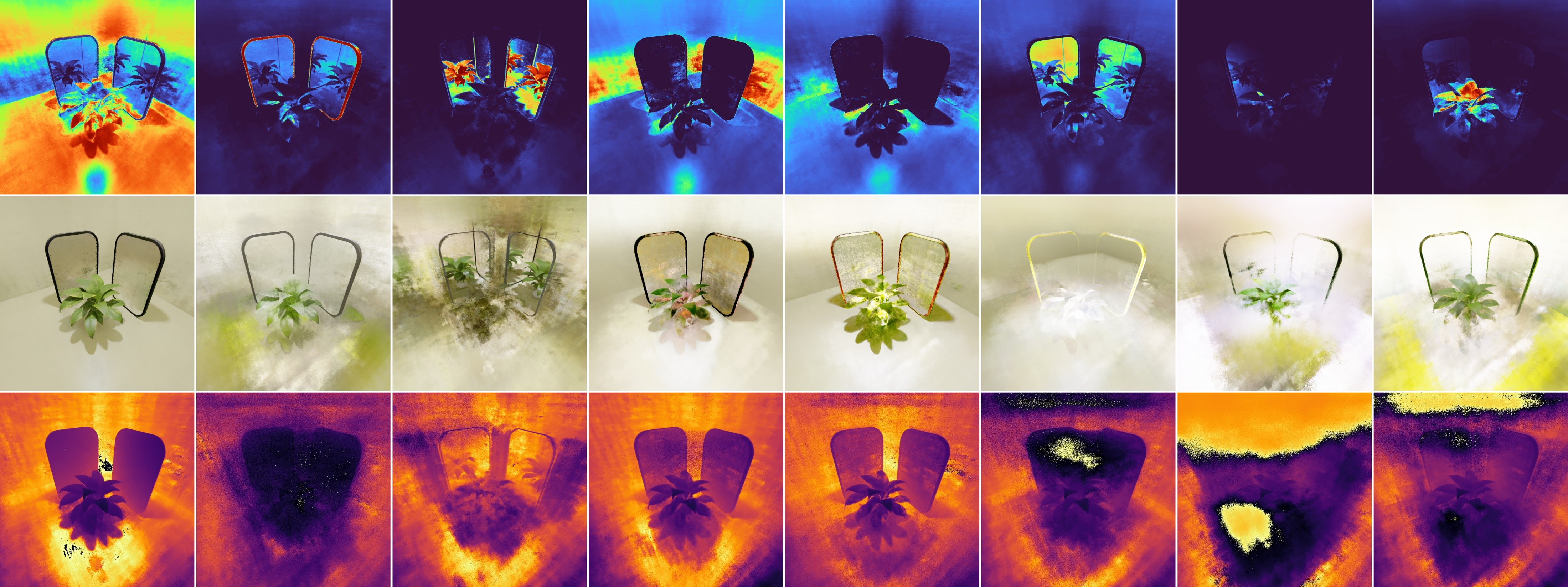}
  \end{subfigure}
  \vfill
  \vspace{2pt}
  \begin{subfigure}{0.155\linewidth}
    \centering
    \includegraphics[width=\textwidth]{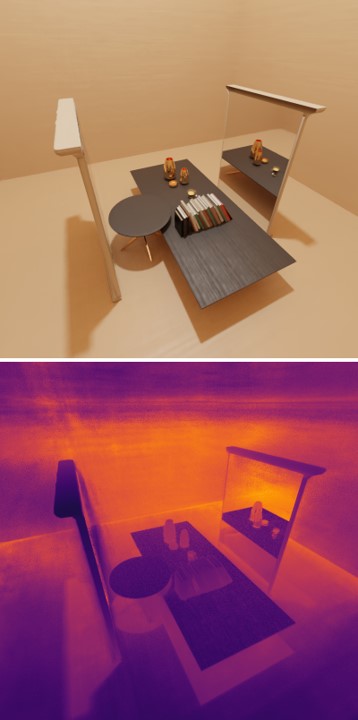}
    \caption{Composed}
    \label{fig:subspace_comp}
  \end{subfigure}
  \hfill
  \begin{subfigure}{0.838\linewidth}
    \centering
    \includegraphics[width=\textwidth]{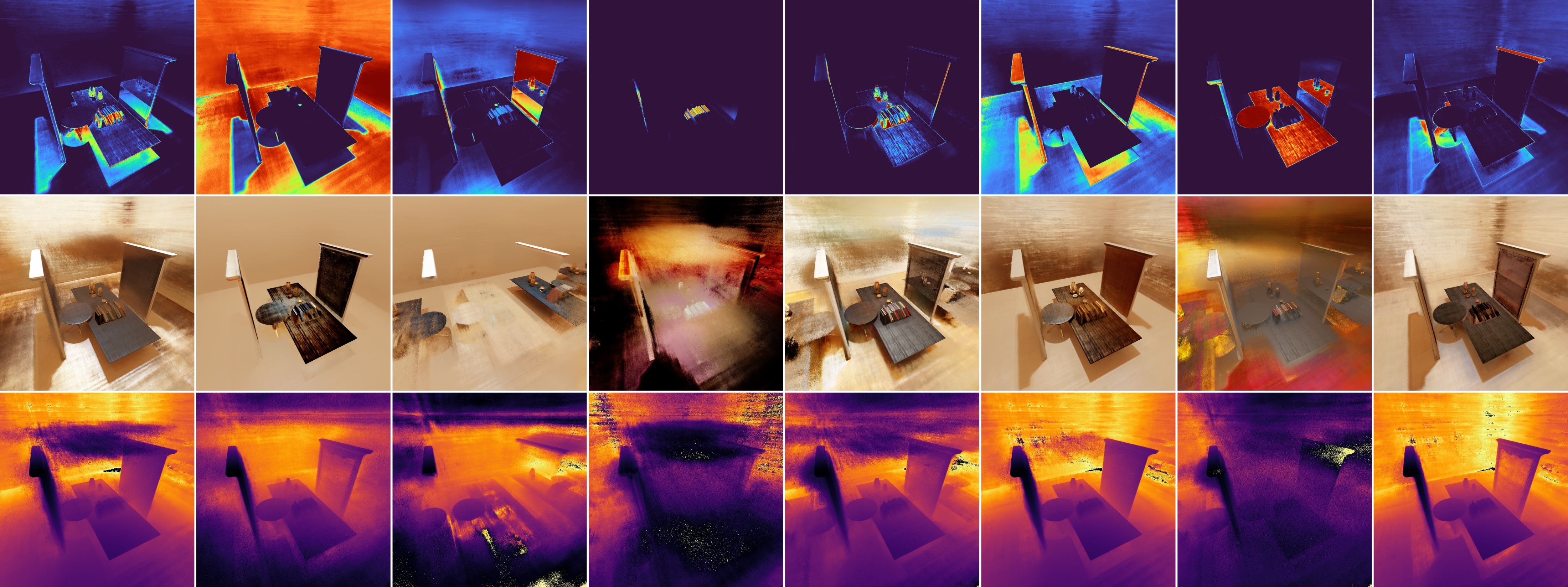}
    \caption{Weight maps, RGB images and depth maps of sub-spaces.}
    \label{fig:subspace_dec}
  \end{subfigure}
  \caption{
  We visualize composed RGB and depth maps of novel views and the decoded images with the corresponding weights 
  and depth maps of all sub-spaces 
  from our $\textrm{MS-NeRF}_B$ model in \secref{sec:main_res}. The results show that our method successfully 
  decomposes virtual images into certain sub-spaces.}
  \label{fig:subspace}
\end{figure*}

\newcommand{\addDepthImg}[1]{\includegraphics[width=.24\linewidth]{light_depth/#1}}

\begin{figure*}[t]
  \centering
  \addDepthImg{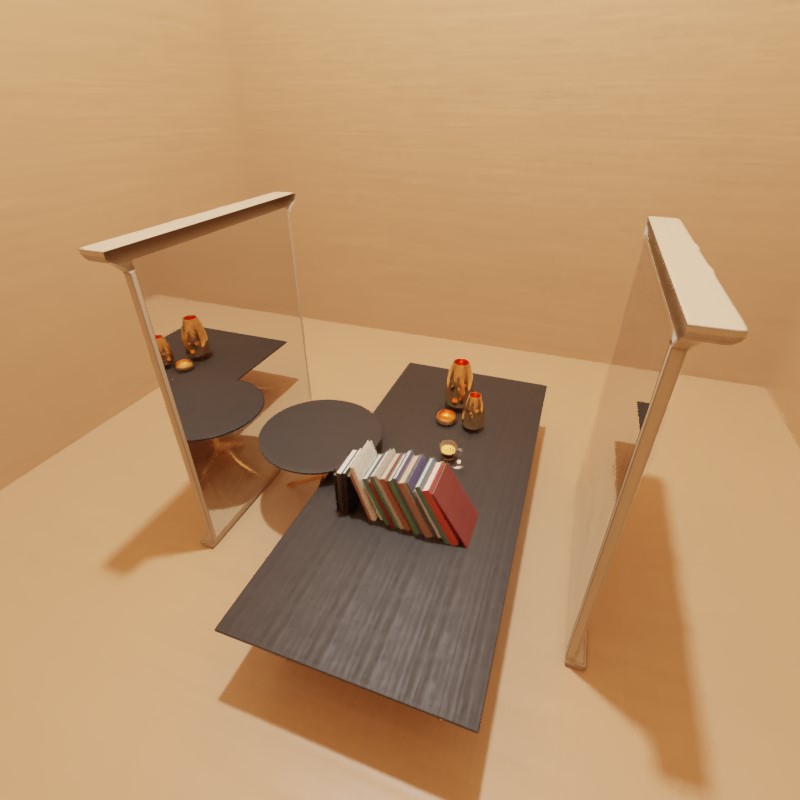} \hspace{-4pt} \addDepthImg{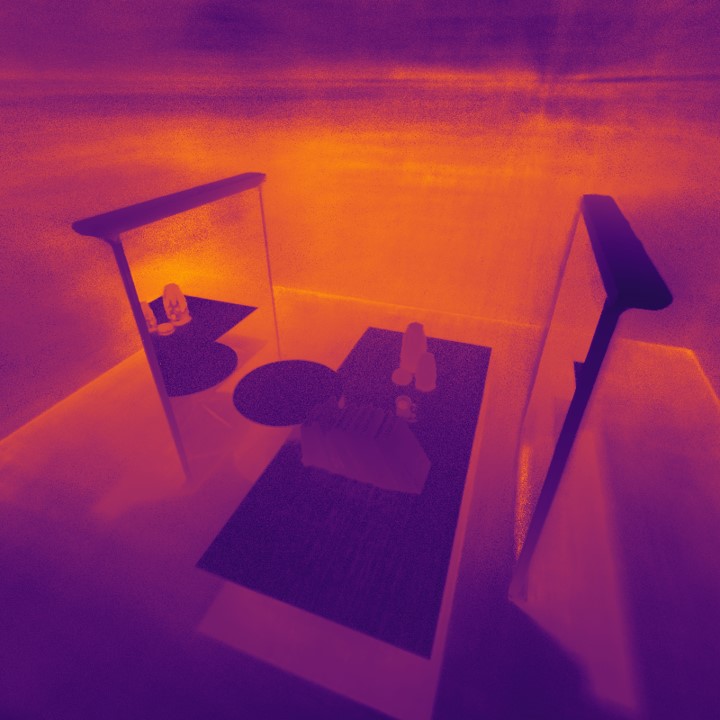} \hspace{-4pt}
  \addDepthImg{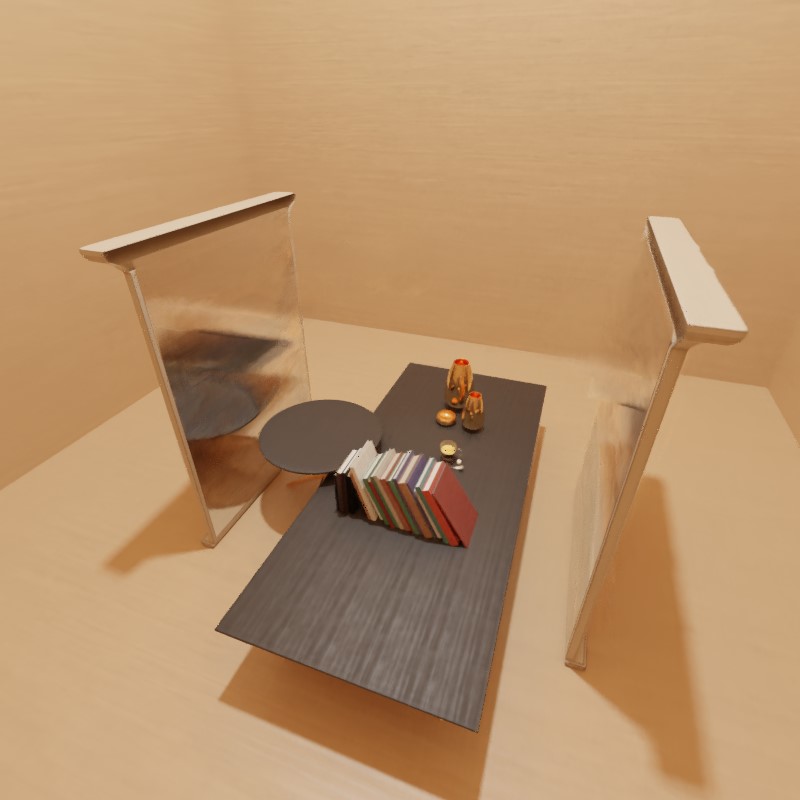} \hspace{-4pt} \addDepthImg{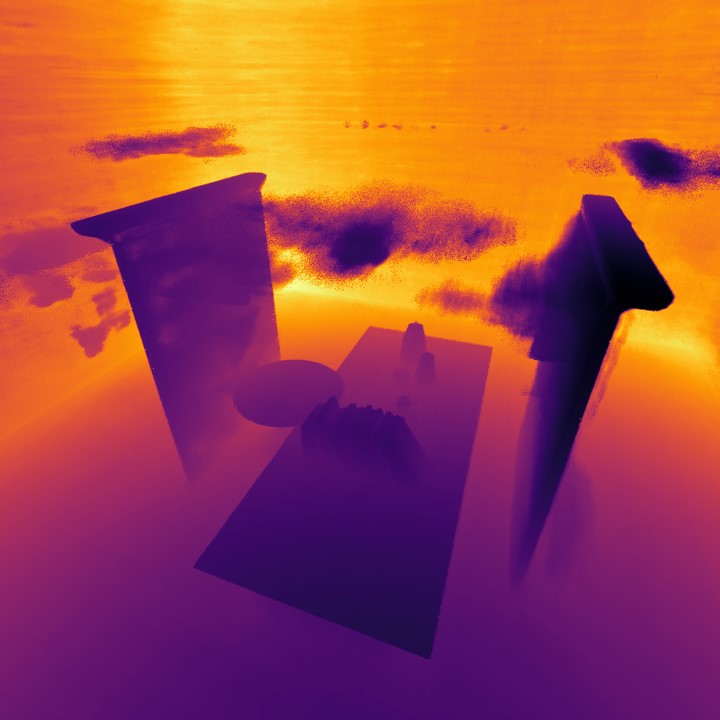} \\ \vspace{2pt} 
  \addDepthImg{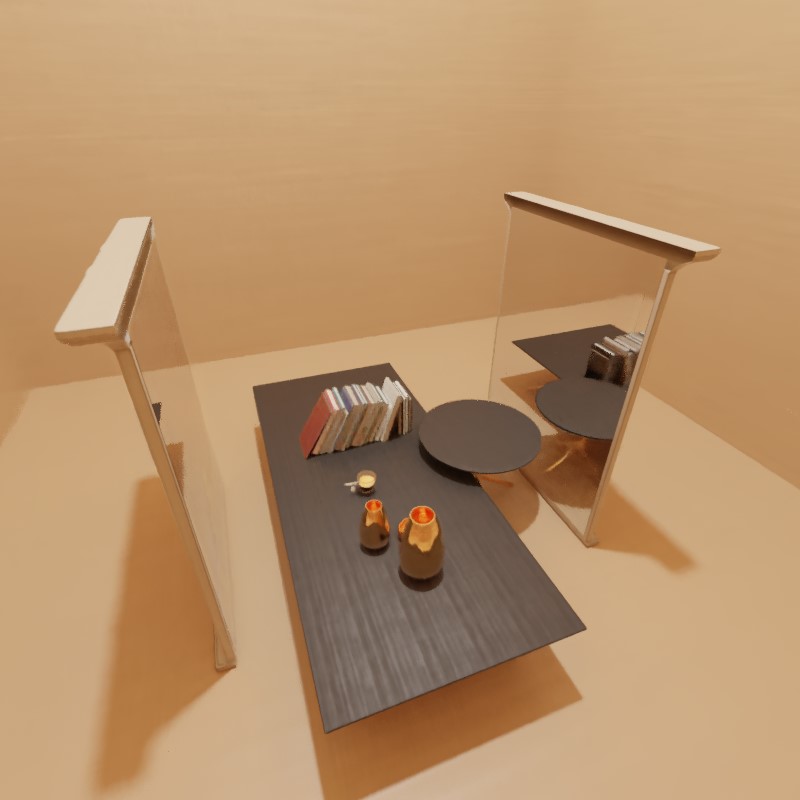} \hspace{-4pt} \addDepthImg{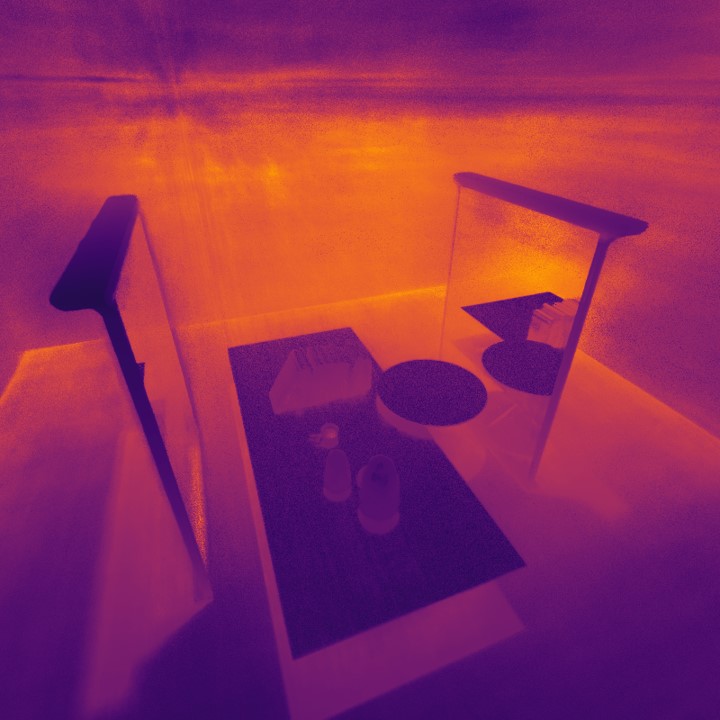} \hspace{-4pt}
  \addDepthImg{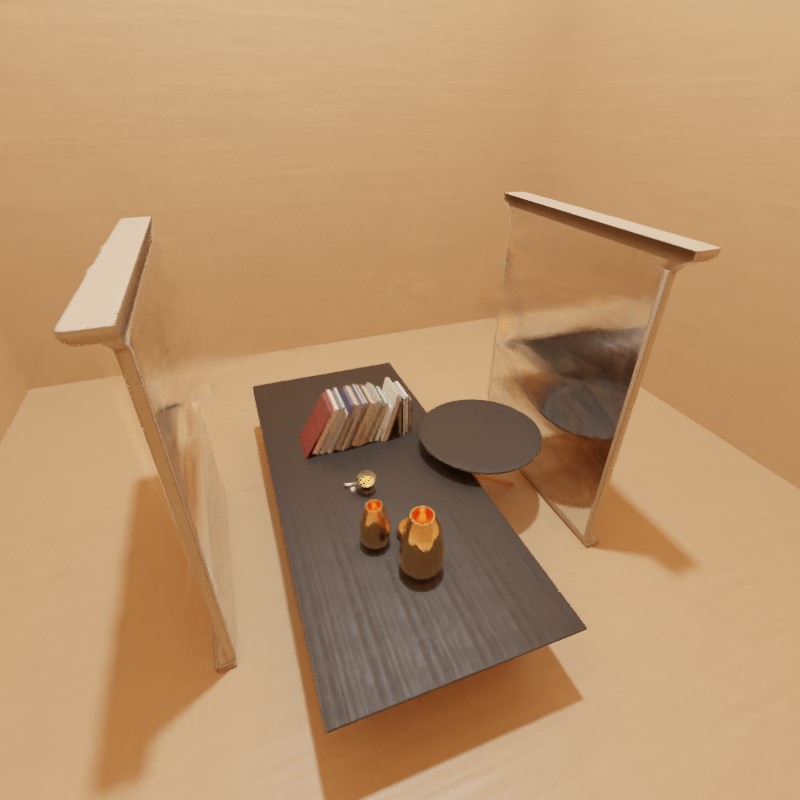} \hspace{-4pt} \addDepthImg{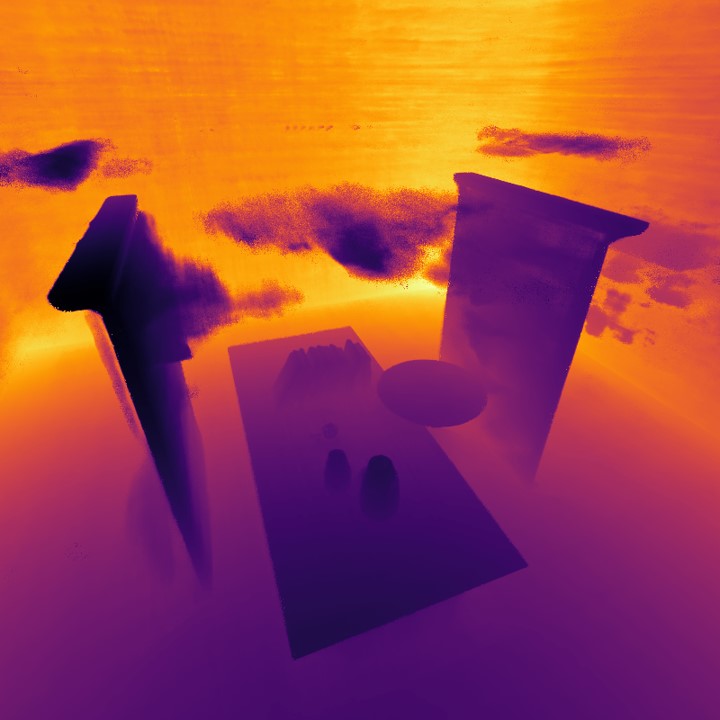} \\
  (a) \hspace{110pt} (b) \hspace{110pt} (c) \hspace{110pt} (d) \\ \vspace{-2pt}
  \caption{{
  (a) render result by $\textrm{MS-NeRF}_B$.
  (b) visualization of depth maps rendered by $\textrm{MS-NeRF}_B$.
  (c) render result by NeRF.
  (d) visualization of depth maps rendered by NeRF.
  The visualization from \secref{sec:main_res} indicates that 
  our MS module understands the light transport at the occurrence of reflections, and the common parts can also be rendered correctly.}}
  \label{fig:light_trans}
\end{figure*}

\subsection{Multi-space Neural Radiance Field}
\label{subsec:hyp}

The volumetric rendering equation and the continuous representation 
ability of MLPs do guarantee the success of NeRF-based methods 
in novel view synthesis, 
but as pointed out by previous works
\cite{jain2022robustifying, lin2021barf, guo2022nerfren}, 
there is also an unignorable property hidden in the training process 
that helps the convergence, which is the multi-view consistency. 
However, the multi-view consistency can be easily violated by 
any reflective surfaces. 
An example is shown in \figref{fig:multiview}, 
when looking in front of a mirror one can observe the reflective 
virtual image as if there were an object behind it, 
but when looking from a side or backward, 
there is actually nothing behind the mirror. 
In practice, this means there will be completely conflictive training 
batches violating the fitting process of MLP.

To experimentally demonstrate the importance of multi-view consistency and its influence on the conventional NeRF network structure, We create two 360-degree toy scenes using an open source software Blender~\cite{blender}, each of which consists of 100 training images and 10 test images, training view examples are shown in \figref{fig:train_a} and \figref{fig:train_b}. The only difference between the two scenes is that we place a mirror-posed real object behind the mirror in the latter scene, but not in the former one.
We train the vanilla NeRF separately on these toy scenes under the same setting and render some views from the test set as in \figref{fig:render_a} and \figref{fig:render_b}, which clearly shows that the vanished virtual image (\ie, violation to the multi-view consistency) in some views leads the model to suboptimal results in reflection-related regions and produces blur in rendering. 
Interestingly, the conventional NeRF is still trying to fulfill the multi-view consistency assumption in the process. 
From the depth map in \figref{fig:render_a}, we can easily conclude that the conventional NeRF treats the viewed virtual image as a ``texture'' on the reflective surface, achieving a compromise between its principle assumption and the conflicts in training data, although the compromise leads to false understandings and worse rendering results of the real scenes.
Contrary to the conventional NeRF, inspired by the common perspective in Physics and Computer Graphics that reflective light can be viewed as ``directly emitted'' from its mirror-symmetric direction, from a possible ``virtual source inside the virtual space in the mirror,'' we build our novel multi-space NeRF approach on the following assumption:

\begin{assumption}
At the existence of reflection and refraction, the real Euclidean scene space can be decomposed into multiple virtual sub-spaces. Each sub-space satisfies the multi-view consistency property.
\end{assumption}
It follows that the composition weights of the sub-spaces can change according to the spatial location and the view direction. Thus all sub-space contributes dynamically to the final render result. 
In this way, the violation of the multi-view consistency in real Euclidean space when there is a reflective surface can be overcome by placing the virtual images in certain sub-spaces only visible from certain views, as shown in \figref{fig:subspace}. 
The depth map shown in \figref{fig:light_trans} further confirms the insight 
that our multi-space module equips the conventional NeRF with the ability 
of understanding the possible ``virtual source inside the virtual space in the mirror'',
on the contrary, ordinary NeRF even fails to model complex reflections as textures.

\begin{figure}[t]
  \begin{subfigure}{0.49\linewidth}
    \includegraphics[width=\linewidth]{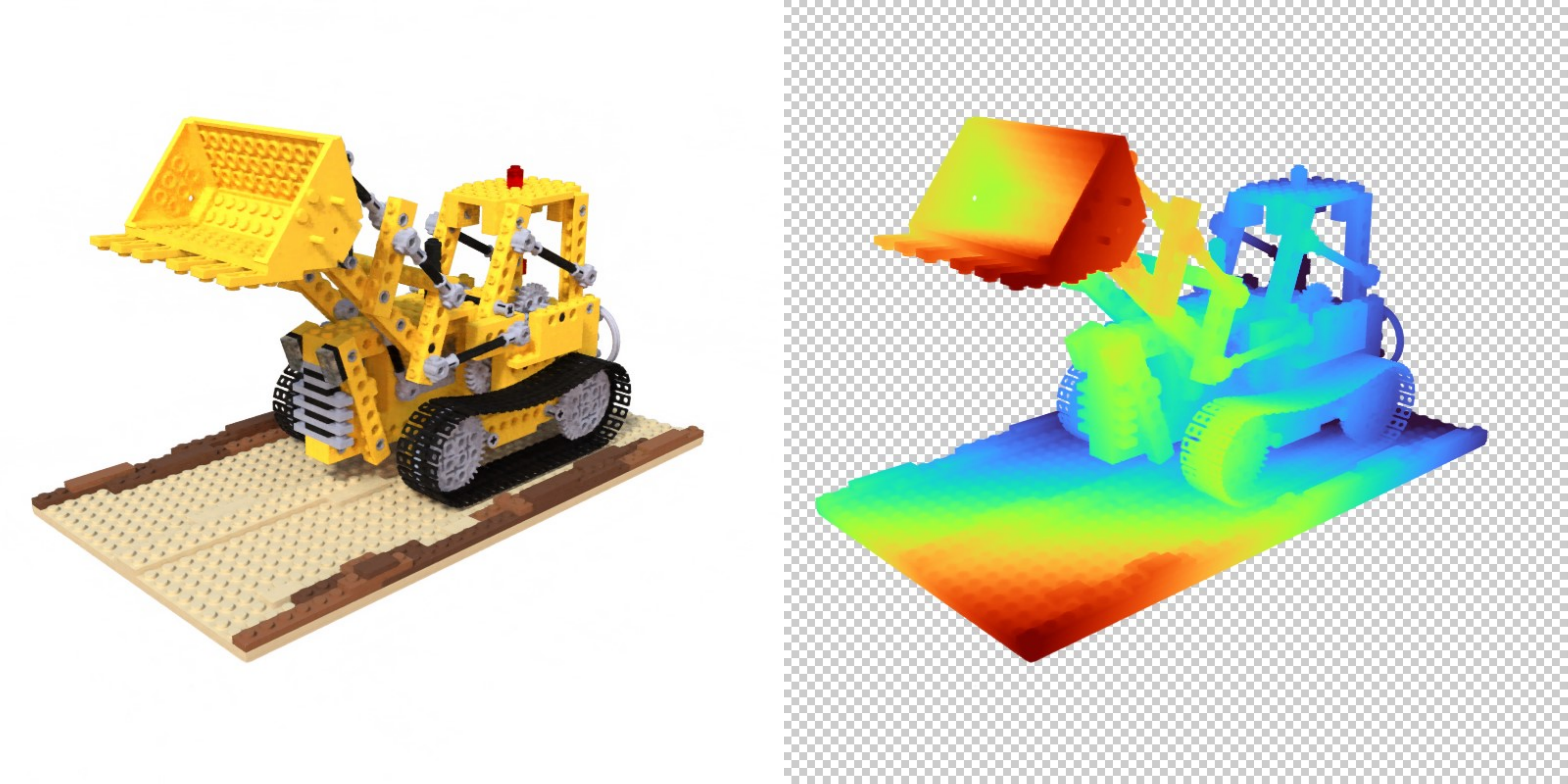}
    \caption{Neural Feature Fields}
  \end{subfigure}
  \hfill
  \begin{subfigure}{0.49\linewidth}
    \includegraphics[width=\linewidth]{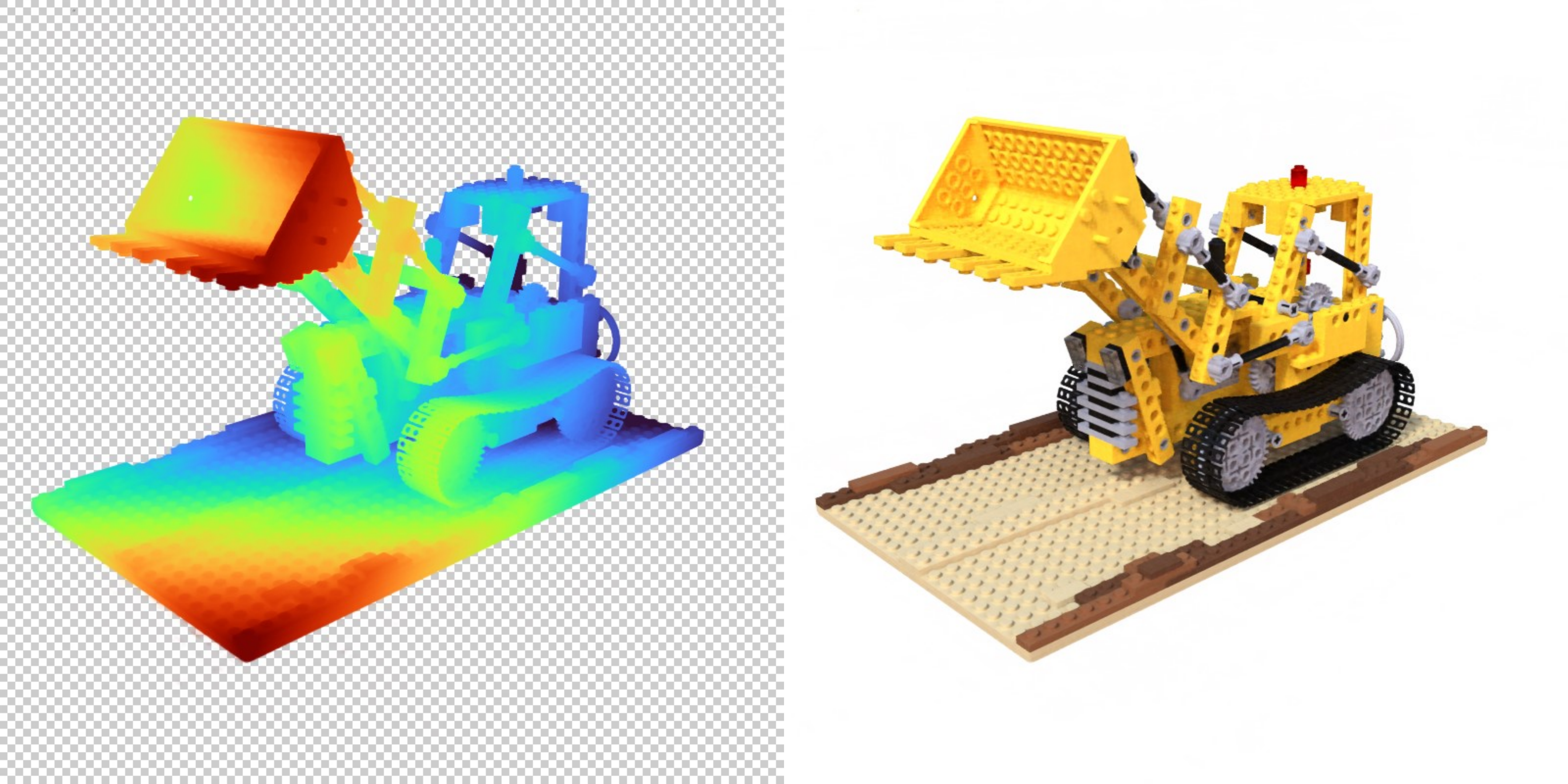}
    \caption{Neural Radiance fields}
  \end{subfigure}
  \vspace{-5pt}
  \caption{
    The rendered depth and RGB map from neural radiance fields and neural feature fields based on Mip-NeRF 360.
  }\vspace{-5pt}
  \label{fig:r_and_f}
\end{figure}

\begin{table}[t!]
  \tablestyle{10.pt}{1}
  \begin{tabular}{cccc}\toprule
    & PSNR$\uparrow$ & SSIM$\uparrow$ & LPIPS$\downarrow$ \\ \midrule
    Mip-NeRF 360 R &  \underline{31.35} &  \underline{0.948} &  \textbf{0.031} \\
    Mip-NeRF 360 F &  \textbf{31.40} &  \textbf{0.948} &  \textbf{0.031} \\
    \midrule
    iNGP R &  \textbf{33.43} &  \textbf{0.964} &  \textbf{0.015} \\
    iNGP F &  \underline{33.25} &  \underline{0.962} &  \underline{0.017} \\
    \bottomrule  
  \end{tabular}
  \caption{Results on the Realistic Synthetic $360^{\circ}$ dataset,
  `R' indicates neural radiance fields, and `F' indicates neural feature fields.
  }\label{tab:r_and_f}
\end{table}

\section{Multi-Space module}
In this section, we revise the neural feature fields in \secref{subsec:feats_fields},
, then we introduce our MLP-based MS module in \secref{subsec:module},
and we integrate out MLP-based MS module into pure MLP-based and grid-based NeRF methods 
to analyze the performance in \secref{sec:grid_msnerf_feats},
finally, we design another hybrid MS module in \secref{subsec:grid_module}.

\begin{figure*}
  \begin{overpic}[width=0.49\linewidth]{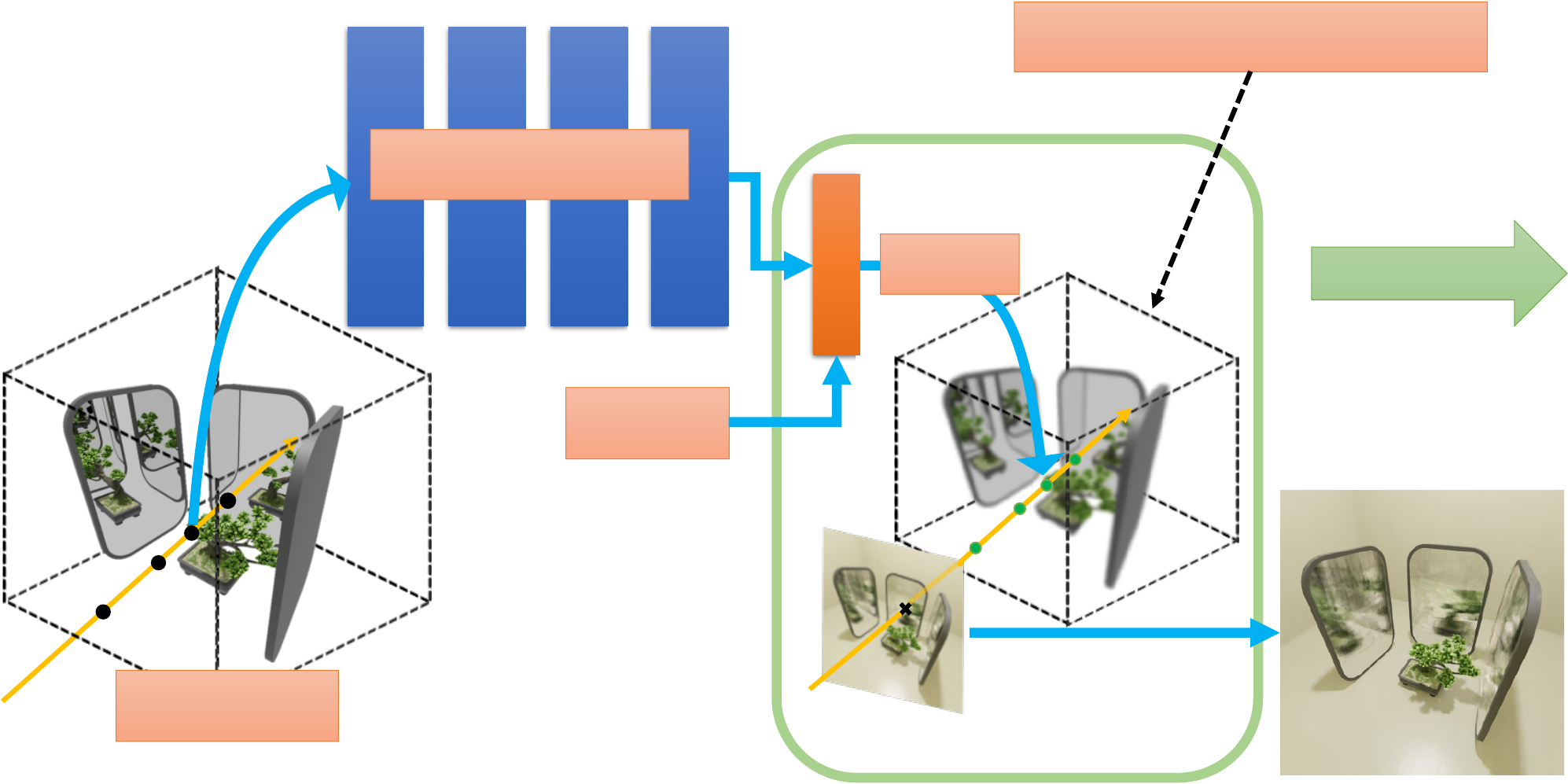} 
    \scriptsize
    \put(9, 4){3D scene}
    \put(24, 38.7){NeRF backbone}
    \put(40, 22.2){$\mathbf{d}$}
    \put(56.9, 32.3){$(\sigma, \mathbf{c})$}
    \put(65.5, 47){Learned radiance fields}
    \put(87, 21){Output}
  \end{overpic}
  %\hfill
  \begin{overpic}[width=0.49\linewidth]{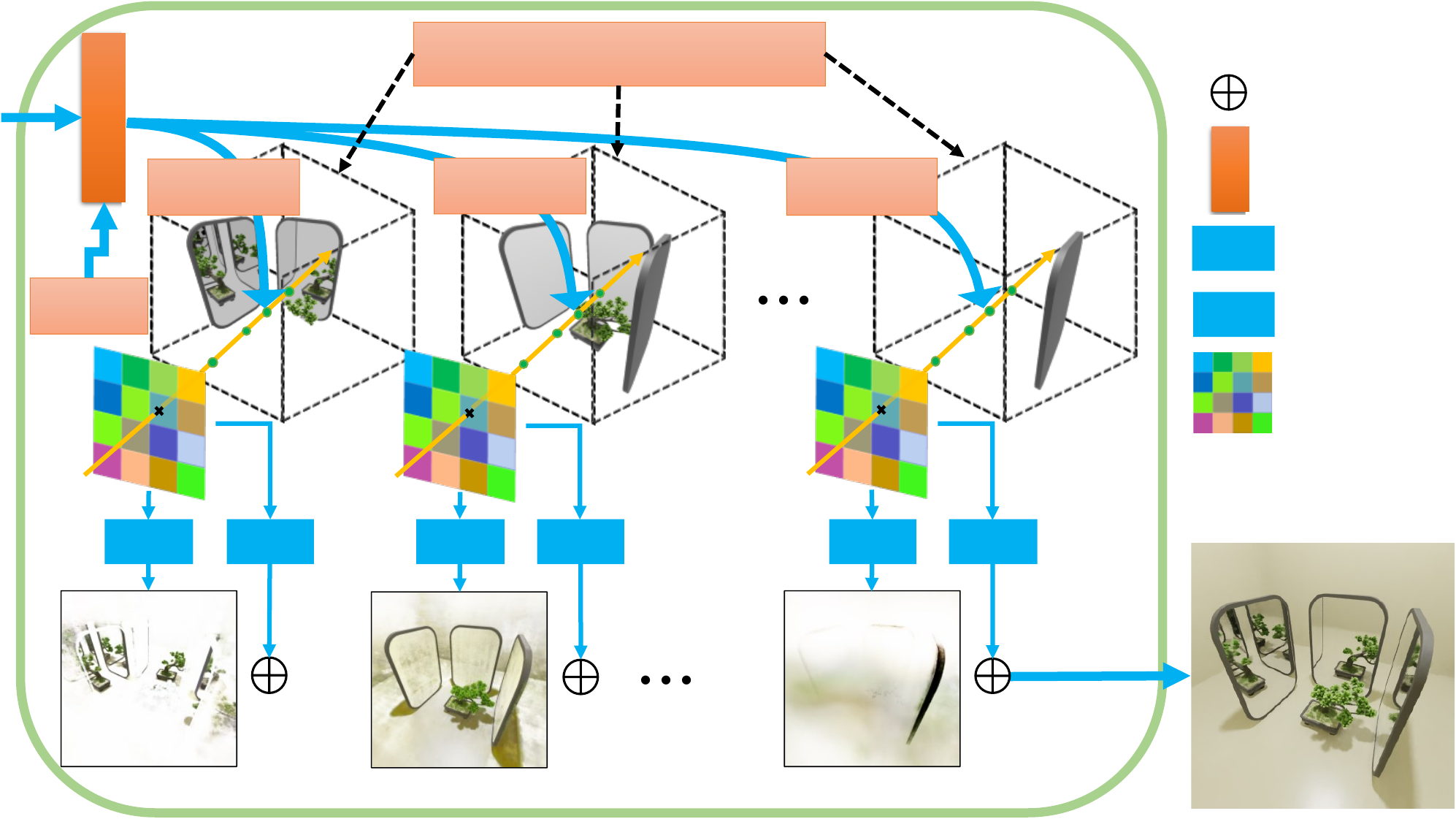} 
    \scriptsize
    \put(5, 34.2){$\mathbf{d}$}
    \put(10.5, 43){\tiny $(\sigma^1, \mathbf{f}^1)$}
    \put(30, 43){\tiny $(\sigma^2, \mathbf{f}^2)$}
    \put(54.1, 43){\tiny $(\sigma^k, \mathbf{f}^k)$}
    \put(29.2, 51.5){Learned feature fields}
    \put(86.5, 49){Weighted sum}
    \put(86.5, 43.5){Output layer}
    \put(82.2, 38.5){$\Theta_D$}
    \put(88.5, 38.5){Decoder MLP}
    \put(82.2, 34){$\Theta_G$}
    \put(88.5, 34){Gate MLP}
    \put(88.5, 28.5){Feature map}
    \put(87, 21){Output}
    \put(7.5, 18.2){$\Theta_D$}
    \put(16, 18.2){$\Theta_G$}
    \put(29, 18.2){$\Theta_D$}
    \put(37.2, 18.2){$\Theta_G$}
    \put(57.5, 18.2){$\Theta_D$}
    \put(65.5, 18.2){$\Theta_G$}
  \end{overpic}
  \caption{Our multi-space module only modifies the output 
    and volumetric rendering part of the network. 
    The original NeRF calculates a pair of density $\sigma$ and 
    radiance $\mathbf{c}$ to get the accumulated color.
    Our output layer produces pairs of densities $\{\sigma^k\}$ and 
    features $\{\mathbf{f}^k\}$, 
    which correspond to multiple parallel feature fields.
    Then, we use volumetric rendering to get multiple feature maps. 
    Two simple MLPs, \ie, Decoder MLP and Gate MLP, are utilized 
    to decode RGB maps and pixel-wise weights from these feature maps.
  }\label{fig:pip}
  \vspace{-5pt}
\end{figure*}

\subsection{Neural Feature Fields}
\label{subsec:feats_fields}
To implement our multi-space neural radiance fields, the underlying field must have the following intrinsics: 
the reconstructed scene must be 3D consistent, as the subspaces are independent scenes, and we try to model each subspace at scene-level;
there must be ways to render 2D images and pixel-level composition maps for each views,
because the model is only supervised by the composed images.
Though we can modify neural radiance fields with an additional channel to output the composition information,
we experimentally verify that adding additional channel is suboptimal in \secref{sec:ablation}.
We can also use neural feature fields to encode both RGB maps and composition maps for rendered views,
which is originally designed for memory saving~\cite{niemeyer2021giraffe} replaces the output colors $\mathbf{c}_i$ of radiance fields 
by features $\{\mathbf{f}^k_i\}$ of $d$ dimension.
When rendering, neural feature fields follow the same 3D points sampling and volumetric rendering scheme as in \secref{subsec:pre}, 
except the rendered map is feature map $\hat{\mathbf{F}}(\mathbf{r})$ instead of $\hat{\mathbf{C}}(\mathbf{r})$, 
then the color map $\hat{\mathbf{C}}(\mathbf{r})$ can be decoded from rendered feature map by a small MLP $\Theta$:
\begin{equation}
  \{\mathcal{F}(\mathbf{r})\}\stackrel{\Theta}{\longrightarrow} \{\mathbf{C}(\mathbf{r})\}
  \label{eq:F_d}
\end{equation}
and the model is supervised by the Ground-Truth images as in \equref{eq:loss}.

Though neural feature fields is capable of encoding both RGB maps and subspace composition maps for each view,
we need to further validate that the neural feature fields are 3D consistent.
With the development of NeRF-based methods, there are two main branches of efforts that improve the rendering quality or rendering speed of NeRF-based methods, 
and the representative methods are Mip-NeRF 360~\cite{barron2022mip} and iNGP~\cite{InstantNGP}.
The former uses better positional parameterization, bigger MLPs, and improved regularization to improve rendering quality,
and the latter designs unique hash positional encoding, occupancy grid-guided sampling strategy, and highly optimized CUDA library to improve the rendering speed.
As we aim to design a unified framework that improves the rendering quality on reflective surfaces of both NeRF-based methods,
we build neural feature fields based on Mip-NeRF 360 and iNGP separately, and carry out experiments on the Realistic Synthetic $360^{\circ}$ dataset from NeRF~\cite{mildenhall2021nerf}.
As shown in \tabref{tab:r_and_f} and \figref{fig:r_and_f}, neural feature fields are equivalent to neural radiance fields in terms of novel view synthesis.

\subsection{Multi-Space module with feature fields}
\label{subsec:module}
Based on the analysis in \secref{subsec:feats_fields},
we propose a compact Multi-Space module (MS module) using the neural feature field scheme~\cite{niemeyer2021giraffe}, 
to sufficiently extract multi-space information from standard NeRF backbone network structures with only small computational overheads. 
Specifically, the MS module will replace the original output layer of the NeRF backbone. 
Below we describe the detailed architecture of our module.

As shown in \figref{fig:pip}, 
our MS module only modifies the output part of vanilla NeRF.
Vanilla NeRF computes single density $\sigma_i$ and radiance $\mathbf{c}_i$ 
for each position along a ray casting through the scene 
and performs volumetric rendering using \equref{eq:intergel} to get
the accumulated color.
On the contrary, our multi-head layer replaces the neural radiance field 
with the neural feature fields~\cite{niemeyer2021giraffe}.
Specifically, the modified output layer gives $K$ densities $\{\sigma^k_i\}$
and features $\{\mathbf{f}^k_i\}$ of $d$ dimension 
for each position along a ray with each pair corresponding to a sub-space, 
where $K$ and $d$ are hyperparameters for the total sub-space number and the feature dimension of the neural feature field, respectively.

We then integrate features along the ray in each sub-space 
to collect $K$ feature maps that encode the color information and 
visibility of each sub-space from a certain viewpoint.
As all pixels are calculated the same way, we denote each pixel as 
$\{\mathcal{F}^k\}$ for simplicity and describe the operation at the pixel level.
Each pixel $\{\mathcal{F}^k\}$ of the feature maps is calculated using:
\begin{equation}
    \hat{\mathcal{F}^k}(\mathbf{r})=\sum_{i=1}^N T^k_i(1-{\rm exp}(-\sigma^k_i \delta_i))\mathbf{f}^k_i,
    \label{eq:intergel_ours}
\end{equation}
where the superscript $k$ indicates the sub-space that the ray casts through. 
The $k$-th density $\sigma^k_i$ and feature $\mathbf{f}^k_i$ correspond to 
the $k$-th sub-space. 
$T^k_i = {\rm exp}(-\sum_{j=1}^{i-1}\sigma^k_j\delta_j)$ and 
$\delta_i = t_i - t_{i-1}$ are similarly computed as in \equref{eq:intergel}.

Then, $\{\mathcal{F}^k\}$ is decoded by two small MLPs, each with just one hidden layer. 
The first is a Decoder MLP that takes $\{\mathcal{F}^k\}$ as input 
and outputs RGB vectors.
The second is a Gate MLP that takes $\{\mathcal{F}^k\}$ as input 
and outputs weights that control the visibility of certain sub-spaces.
Specifically, we use:
\begin{equation}
    \{\mathcal{F}^k\}\stackrel{\Theta_D}{\longrightarrow} \{\mathbf{C}^k\}, \{\mathcal{F}^k\}\stackrel{\Theta_{G}}{\longrightarrow} \{w^k\},
    \label{eq:F_k}
\end{equation}
where $\Theta_D$ represents the Decoder MLP, and $\Theta_{G}$ represents 
the Gate MLP.
In the end, the MS module applies the softmax function to $\{w^k\}$ 
as the color contribution of each subspace to form the final render results:
\begin{equation}
    \hat{\mathbf{C}}(\mathbf{r}) = \frac{1}{\sum_{i=1}^K\mathrm{exp}(w^i)}\sum_{k=1}^{K}\mathrm{exp}(w^k)\mathbf{C}^k.
    \label{eq:C_ours}
\end{equation}
\equref{eq:C_ours} needs no additional loss terms compared with the vanilla NeRF method. 
As a result, the above light-weighted MS module is able to serve as an enhancement module onto the conventional NeRF-like backbone networks, 
and we will show that our approach achieves significant enhancements in \secref{sec:main_res}. 

\subsection{Grid-based Multi-Space NeRF with feature fields}
\label{sec:grid_msnerf_feats}
Grid-based techniques~\cite{InstantNGP} promote the development of NeRF-based methods by a huge step, especially in acceleration fields
where grid-based NeRF methods are able to converge to comparably high quality in just minutes, 
while pure MLP-based methods require hours to days of training.
These methods follow the design convention of small MLPs and learnable explicit parameters organized in 2D/3D grids,
where the grids explicitly encode coordinate-related information, and the MLPs act more like decoders.
Grid-based NeRF methods follow the rendering equation in \equref{eq:intergel}, 
except that the positional encoding function \equref{eq:PosEnc} is replaced 
by a query function $Q(\mathbf{V}, \mathbf{p})$ 
mapping sampled points $\{\mathbf{p}_i\}$ into the values 
from the corresponding positions in the grid 
$\mathbf{V} \in \mathbb{R}^{\mathrm{d}_x \times \mathrm{d}_x \times \mathrm{d}_z \times \mathrm{d}_v}$,
where $\mathrm{d}_x, \mathrm{d}_y, \mathrm{d}_z$ represents the grid resolution along x-, y-, and z-axis 
while $\mathrm{d}_v$ represents the dimension of queried features.
As these methods render images from single radiance fields like previous methods,
they also fail to render high-quality reflection and refraction.

\begin{figure}
  \begin{subfigure}{0.32\linewidth}
    \includegraphics[width=\linewidth]{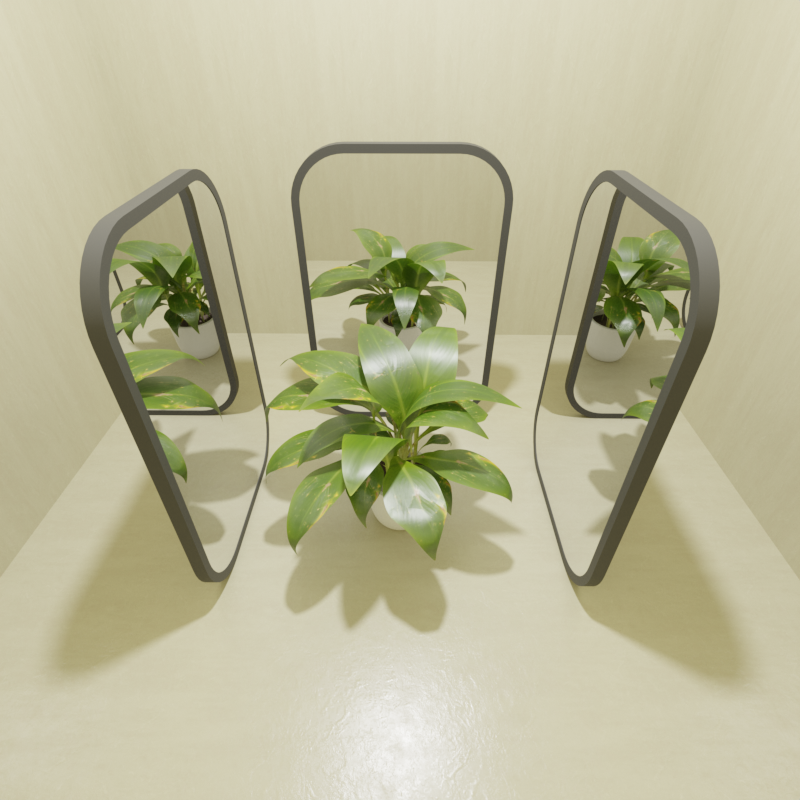}
    \caption{Ground-Truth}
  \end{subfigure}
  \hfill
  \begin{subfigure}{0.32\linewidth}
    \includegraphics[width=\linewidth]{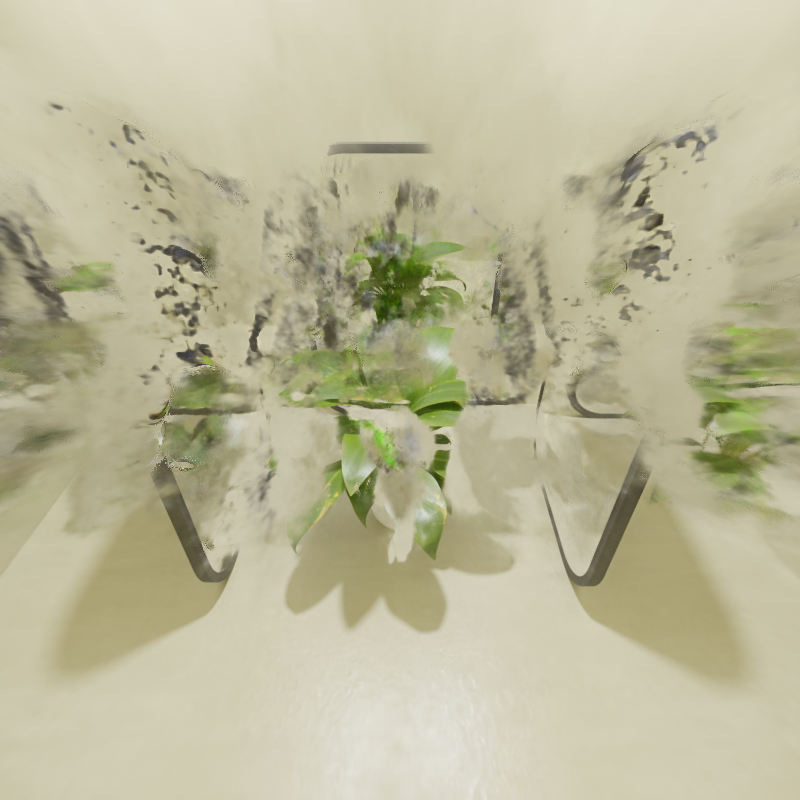}
    \caption{w/o our module}
  \end{subfigure}
  \hfill
  \begin{subfigure}{0.32\linewidth}
    \includegraphics[width=\linewidth]{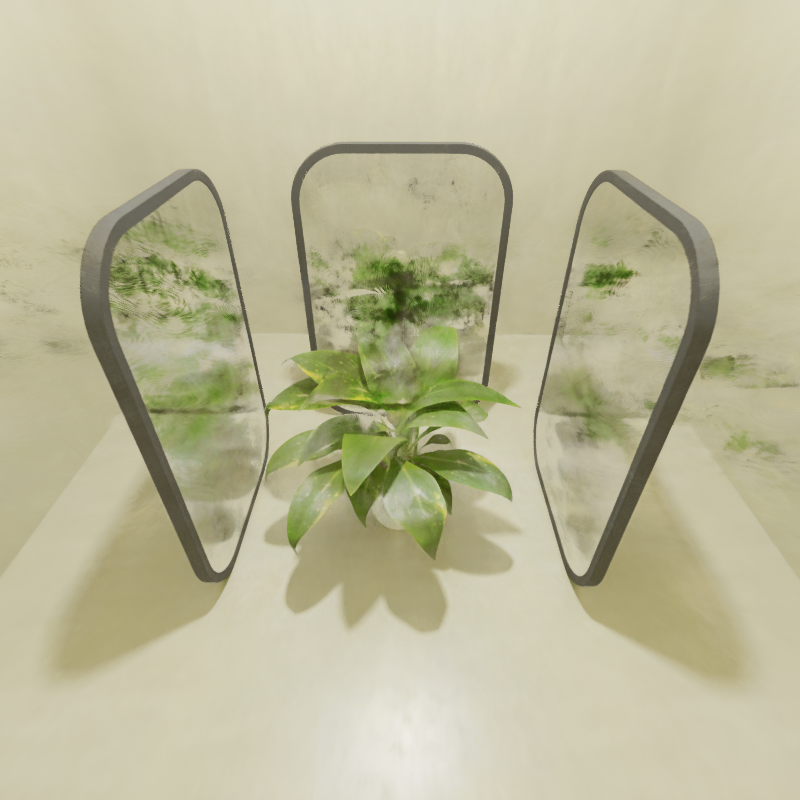}
    \caption{w/ our module}
  \end{subfigure}
  \vspace{-5pt}
  \caption{
  Visual comparison between iNGP with and without our MS module in \secref{subsec:module}.
  }\vspace{-5pt}
  \label{fig:vis_msngp}
\end{figure}

We choose the iNGP~\cite{InstantNGP} implemented with the proposal estimator in \cite{li2023nerfacc} as our baseline 
and simply integrate our MS module in \secref{subsec:module} into it as multi-space iNGP.
We conduct experiments on the Scene04 with circle camera path from our synthesized dataset in \secref{sec:our_dataset},
because this scene contains relatively complex reflections and simple objects.
As shown in \figref{fig:vis_msngp},
though our MS module prevents the model from collapsing,
it still struggles to render high-quality reflections.

\begin{figure}[t]
  \centering
  \begin{overpic}[width=0.99\linewidth]{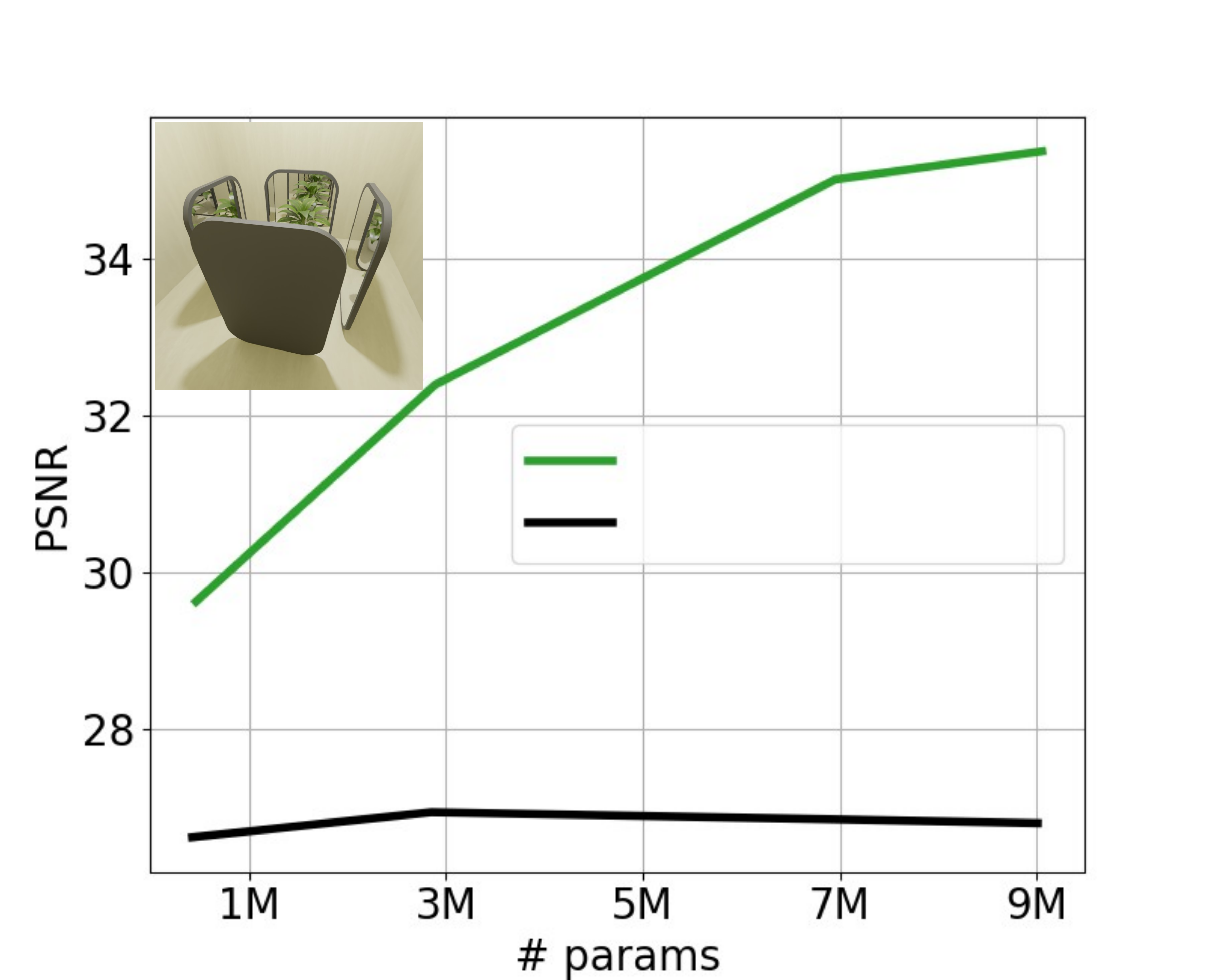} 
    \small
    \put(53, 41.8){MS-Mip-NeRF 360}
    \put(53, 36.8){Mip-NeRF 360}
  \end{overpic}
  \caption{We integrate our MLP-based multi-space module with 6 subspaces into the Mip-NeRF 360 method 
  and scale the size of the NeRF MLP in the model by the network depth and width;
  then, we evaluate the performance by PSNR on the shown scene with the circle camera path.}
  \label{fig:model_size}
\end{figure}

The results in \tabref{tab:r_and_f} also show that the Mip-NeRF 360 gains performance increase from feature fields,
while the performance of iNGP drops when integrated with feature fields,
therefore, we suspect that the performance of our MS module with feature fields should be related to the MLP capacity of NeRF model.
To validate the above speculation, we carry out experiments on Scene05 with the circle camera path from our synthesized dataset in \secref{sec:our_dataset},
which contains infinite reflections and relatively simple objects, therefore, the performance improvements are clearer.
We integrate the MS module in \secref{subsec:feats_fields} with Mip-NeRF 360
and scale the base MLP size by network depth and width;
the results in \figref{fig:model_size} show that bigger MLPs gain more performance increase
from our MS module.

\begin{figure}
  \begin{subfigure}{0.99\linewidth}
    \includegraphics[width=\linewidth]{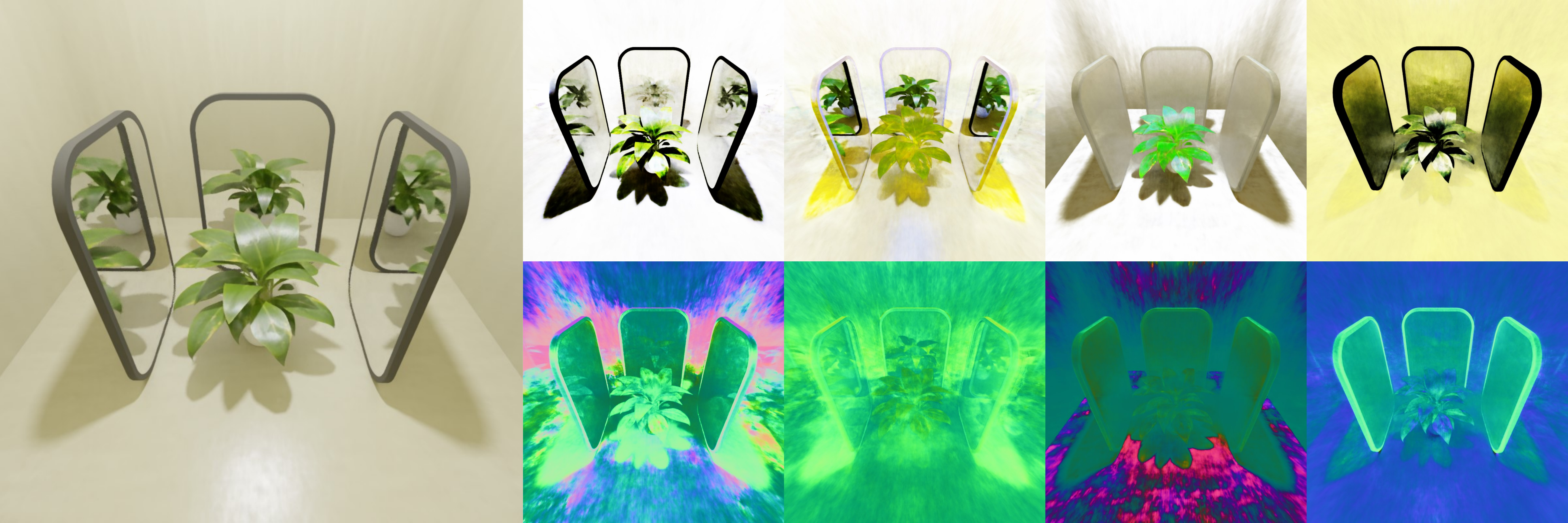}
    \caption{Multi-space Mip-NeRF 360}
  \end{subfigure}
  \vfill
  \begin{subfigure}{0.99\linewidth}
    \includegraphics[width=\linewidth]{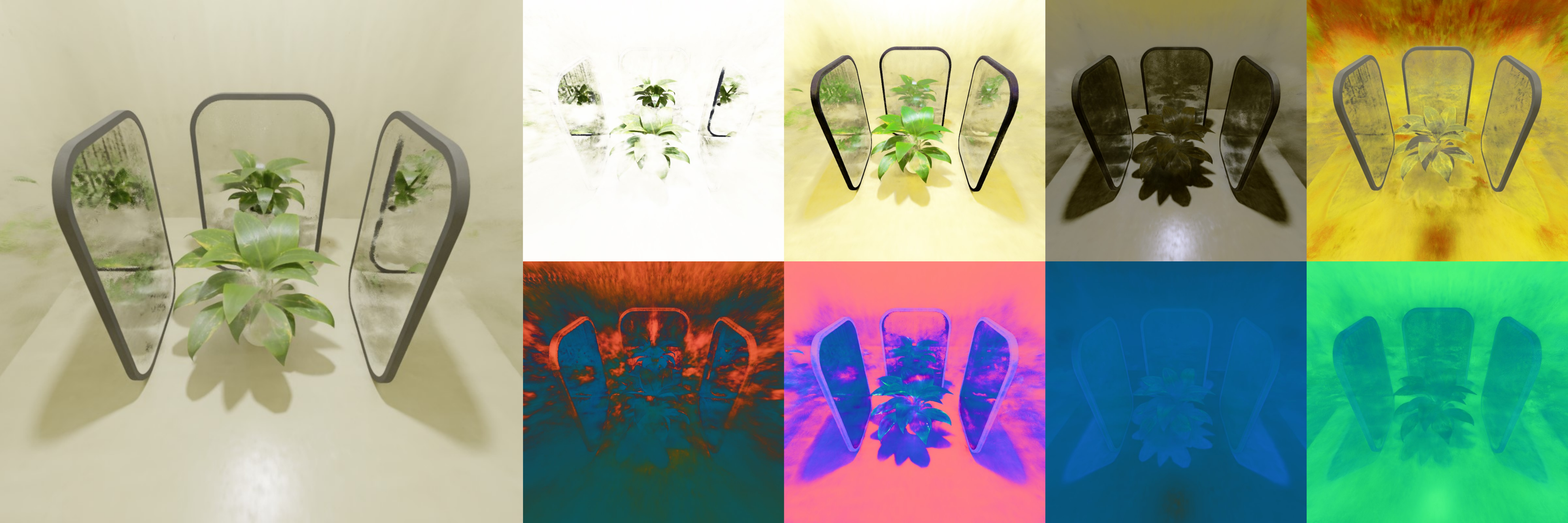}
    \caption{Multi-space iNGP}
  \end{subfigure}
  \vspace{-5pt}
  \caption{
    Visualization of the sub-spaces rendered from Mip-NeRF 360 and iNGP with our MS module in \secref{subsec:module}.
    The left image is the composed image, and the right ones are sub-space RGB maps and sub-space feature maps with PCA transformation.
  }\vspace{-5pt}
  \label{fig:vis_sub}
\end{figure}

Also, we visualize the RGB maps and the feature maps visualized by PCA transformation of four sub-spaces 
rendered by the multi-space iNGP and multi-space Mip-NeRF 360 in \figref{fig:vis_sub},
which shows that bigger MLPs with our MS module can reconstruct cleaner sub-spaces,
while iNGP is designed with small MLPs and directly integratd with MS module from \secref{subsec:module} can only reconstruct suboptimal subspaces.

\begin{figure*}
  \begin{overpic}[width=0.99\linewidth]{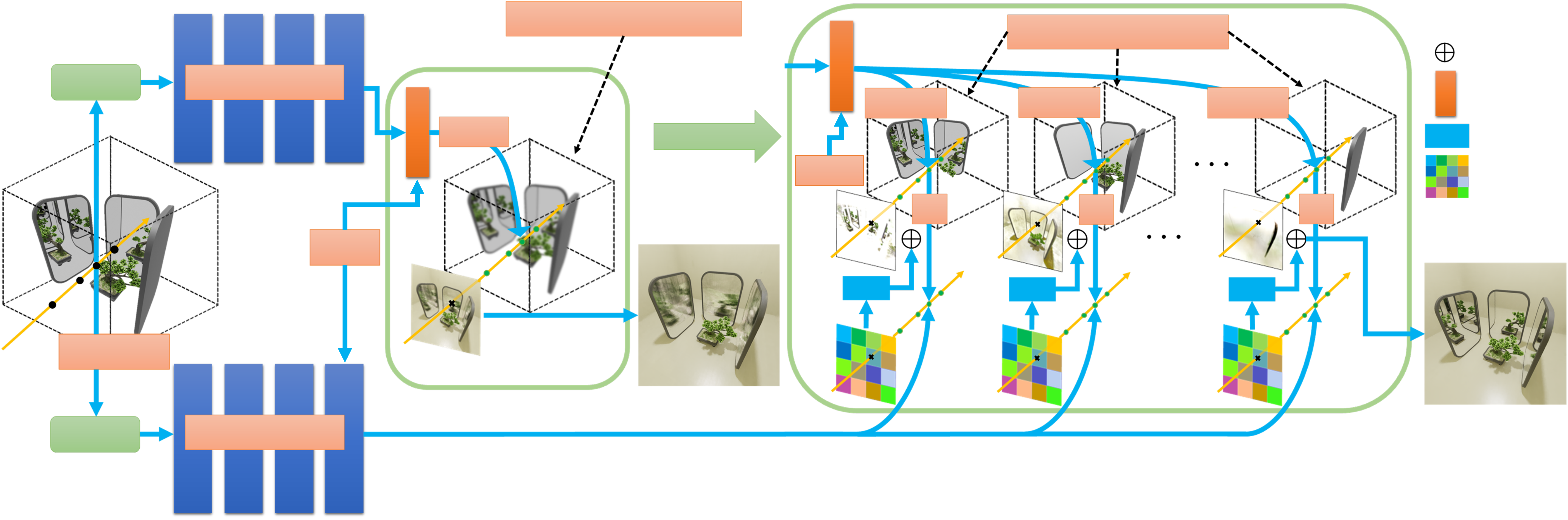} 
    \scriptsize
    \put(21.45, 16.85){$\mathbf{d}$}
    \put(4.3, 4.9){$\gamma({\rm \mathbf{p}})$}
    \put(4.3, 10.2){3D scene}
    \put(3.5, 27.5){\tiny $Q({\rm \mathbf{V}},{\rm \mathbf{p}})$}
    \put(13.5, 5.0){Small MLP}
    \put(12, 27.3){NeRF backbone}
    \put(28.5, 24.2){$(\sigma, \mathbf{c})$}
    \put(32.7, 31.5){Learned radiance fields}
    \put(43, 18){Output}
    
    \put(52.35, 21.75){$\mathbf{d}$}
    \put(55.4, 26.2){\tiny $(\sigma^1, \mathbf{c}^1)$}
    \put(65.2, 26.2){\tiny $(\sigma^2, \mathbf{c}^2)$}
    \put(77.1, 26.2){\tiny $(\sigma^k, \mathbf{c}^k)$}
    \put(65, 30.5){Multi radiance fields}
    \put(94, 29.5){Weighted sum}
    \put(94, 26.5){Output layer}
    \put(91.2, 24){$\Theta_G$}
    \put(94, 24){Gate MLP}
    \put(94, 21.5){Feature map}
    \put(93.5, 17){Output}
    \put(58.8, 19.2){\tiny $\sigma^1$}
    \put(69.2, 19.2){\tiny $\sigma^2$}
    \put(83, 19.2){\tiny $\sigma^k$}
    \put(54.2, 14.2){$\Theta_G$}
    \put(64.7, 14.2){$\Theta_G$}
    \put(78.8, 14.2){$\Theta_G$}
  \end{overpic}
  \caption{The illustration of the proposed hybrid MS module.
  Considering the architecture of grid-based methods, 
  we decompose the multi-space gate information from the radiance fields branch
  and use another small branch
  to model multiple feature fields, from which we decode the gate information.
  Along with the multiple radiance fields, we perform weighted sum to get the final rendering image.}
  \label{fig:grid_pip}
  \vspace{-5pt}
\end{figure*}

\subsection{Hybrid Multi-Space Module}
\label{subsec:grid_module}

Though MS module with feature fields exhibit good performance with MLP-based NeRF methods,
it is not compatible with grid-based methods.
Considering the representation ability of small MLPs, 
we comprise to model the sub-spaces in grid-based methods using neural radiance fields as shown in \figref{fig:grid_pip}.
Given sampled points $\{\mathbf{p}_i\}$ along the ray and the explicit grid parameters $\mathbf{V}$, we modify the output layer 
to map each features vector queried by $Q(\mathbf{V}, \mathbf{p})$ 
and the ray direction $\mathbf{d}$ 
to $K$ densities $\{\sigma^k_i\}$ and colors $\{c^k_i\}$,
which model multiple radiance fields instead of multiple feature fields,
where $K$ represent the total number of sub-spaces.
Then, a multi-space integration in \equref{eq:intergel_ours} is performed, 
except that we integrate K colors $\{c^k_i\}$ instead of features $\{\mathbf{f}^k_i\}$, 
to form $K$ color maps $\{\mathbf{C}^k\}$.
We decompose the multi-space composition information from the NeRF model,
and compress it into a small MLP, as in scenes with reflections and refractions, 
the visibility of virtual images is only related to 3D positions and view directions.
Specifically, we design another branch with a small pure MLP network,
which maps each 3D positions $\{\gamma(\mathbf{p}_i)\}$, 
where $\gamma(\dots)$ is the positional encoding from \equref{eq:PosEnc}, 
and view directions $\mathbf{d}$ to features $\{\mathbf{f}_i\}$ of $d$ dimension.
Note that we record one feature vector for each position due to the model capacity.
Then we put the features $\{\mathbf{f}_i\}$ into each sub-space along the rays,
and perform volumetric rendering using the features $\{\mathbf{f}_i\}$ 
along with each of the $K$ densities $\{\sigma^k_i\}$ from 
the multiple radiance fields branch to form $K$ feature maps $\{\mathcal{F}^k\}$, 
as in \equref{eq:intergel_ours}.
Finally, we decode the pixel-wise composition weight map from $\{\mathcal{F}^k\}$ following \equref{eq:F_k},
and we compose the rendered results using the $K$ color maps $\{\mathbf{C}^k\}$ from NeRF branch and the decoded weight map $\{w^k\}$ as in \equref{eq:C_ours}.

%%%%%%%%%%%%%%%%%method%%%%%%%%%%%%%%%%%%

%%%%%%%%%%%%%%%%%dataset%%%%%%%%%%%%%%%%%%
\section{Dataset}
\label{sec:dataset}

\begin{table*}[t]
  \tablestyle{4.7pt}{1.1}
    \begin{tabular}{cccccccc}
      \toprule
      dataset & origin & applications & Type & viewpoints & properties & number \\
      \midrule
      Realistic Synthetic $360^{\circ}$ & \cite{mildenhall2021nerf} & view synthesis & S & 360-degree & non-Lambertian & 8 \\
      Real Forward-Facing & \cite{mildenhall2021nerf, mildenhall2019local} & view synthesis & R & forward-facing & non-Lambertian & 8 \\
      Shiny & \cite{wizadwongsa2021nex} & view synthesis & R & forward-facing & high-specular, refraction & 8 \\
      Tanks and Temples(T\&T) & \cite{knapitsch2017tanks} & view synthesis & R & 360-degree & unbounded scenes & 4 \\
      Mip-NeRF 360 & \cite{barron2022mip} & view synthesis & R & 360-degree & unbounded scenes & 9 \\
      EikonalFields & \cite{bemana2022eikonal} & view synthesis & R & 360-degree & refraction & 4 \\
      RFFR & \cite{guo2022nerfren} & view synthesis & R & forward-facing & reflection, semi-transparent & 6 \\
      DTU & \cite{jensen2014large} & reconstruction & R & 360-degree & non-Lambertian objects & 15* \\
      BlendedMVS & \cite{yao2020blendedmvs} & reconstruction & S & 360-degree & non-Lambertian scenes & 7* \\
      Shiny Blender & \cite{verbin2022refnerf} & view synthesis & S & 360-degree & glossy materials & 6 \\
      Ref-NeRF Real captured scenes & \cite{verbin2022refnerf,hedman2021baking} & view synthesis & R & 360-degree & glossy materials & 3\\
      \bottomrule 
    \end{tabular}
  \caption{Properties of a commonly used dataset for NeRF-based methods.
    `S' and `R' represent synthesized and real captured, respectively. 
    We denote those unnamed datasets with the name of the methods. 
    `*' refers to the number of scenes commonly used by NeRF-based methods, 
    as the original dataset contains more scenes than noted, and we do not take them into consideration.
  }\label{tab:existing_dataset}
  \vspace{-12pt}
\end{table*}

\subsection{Existing datasets}
\label{sec:exist_dataset}

We briefly revisit the commonly used or most relevant datasets to our task
and list their properties in \tabref{tab:existing_dataset}.
With these well-designed datasets, NeRF-based methods have achieved great improvements 
in many applications under various settings, such as novel view synthesis in unbounded scenes and 
3D reconstruction.
However, most existing dataset fails to cover scenes containing complex light paths
with the camera moving 360-degree around,  \eg, a glass of water in front of a mirror,
which is very common in our daily life.
~\cite{guo2022nerfren} propose the RFFR (Real Forward-Facing with Reflections) dataset,
which contains 6 forward-facing scenes with reflective objects, such as transparent glass
and mirrors.
However, views behind the reflective objects cannot be evaluated, which is crucial 
for understanding reflections.
~\cite{bemana2022eikonal} propose a dataset containing 4 scenes,
where cameras move around the central refractive objects with a large view range up to 360-degree.
However, it is a nearly object-level dataset, and the refractive is rather simple.

\subsection{Our proposed dataset}
\label{sec:our_dataset}

\begin{figure}[t]
  \centering
  \begin{subfigure}{0.99\linewidth}
    \centering
	  \includegraphics[width=\textwidth]{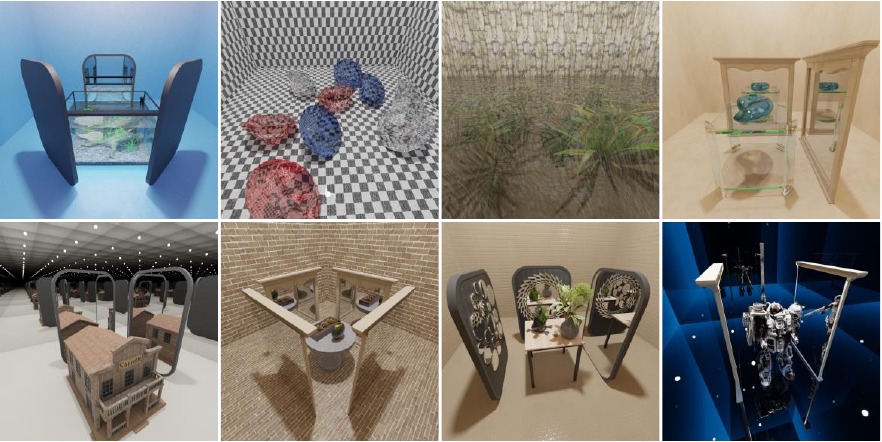}
    \caption{A part of our synthesized dataset.}
    \label{fig:syn_dataset}
  \end{subfigure}
  \vfill
  \begin{subfigure}{0.99\linewidth}
    \centering
	  \includegraphics[width=\textwidth]{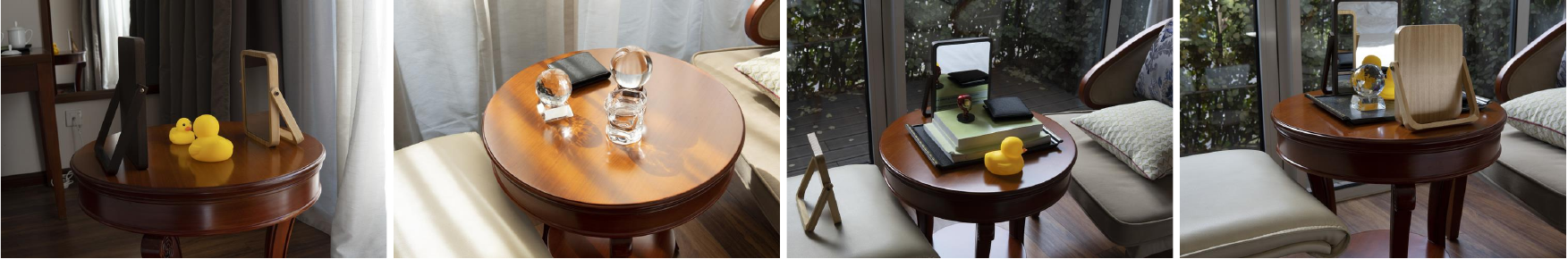}
    \caption{A part of our real captured dataset.}
    \label{fig:real_dataset}
  \end{subfigure}
  \vspace{-5pt}
  \caption{Demo scenes of our datasets (more in the supplementary). 
    Our dataset exhibits diversities of reflection and refraction, 
    which can serve as a benchmark for validating the ability to synthesize 
    novel views with complex light paths.
  }\vspace{-10pt}
  \label{fig:dataset_demo}
\end{figure}

\begin{figure}[t]
  \centering
  \includegraphics[width=0.99\linewidth]{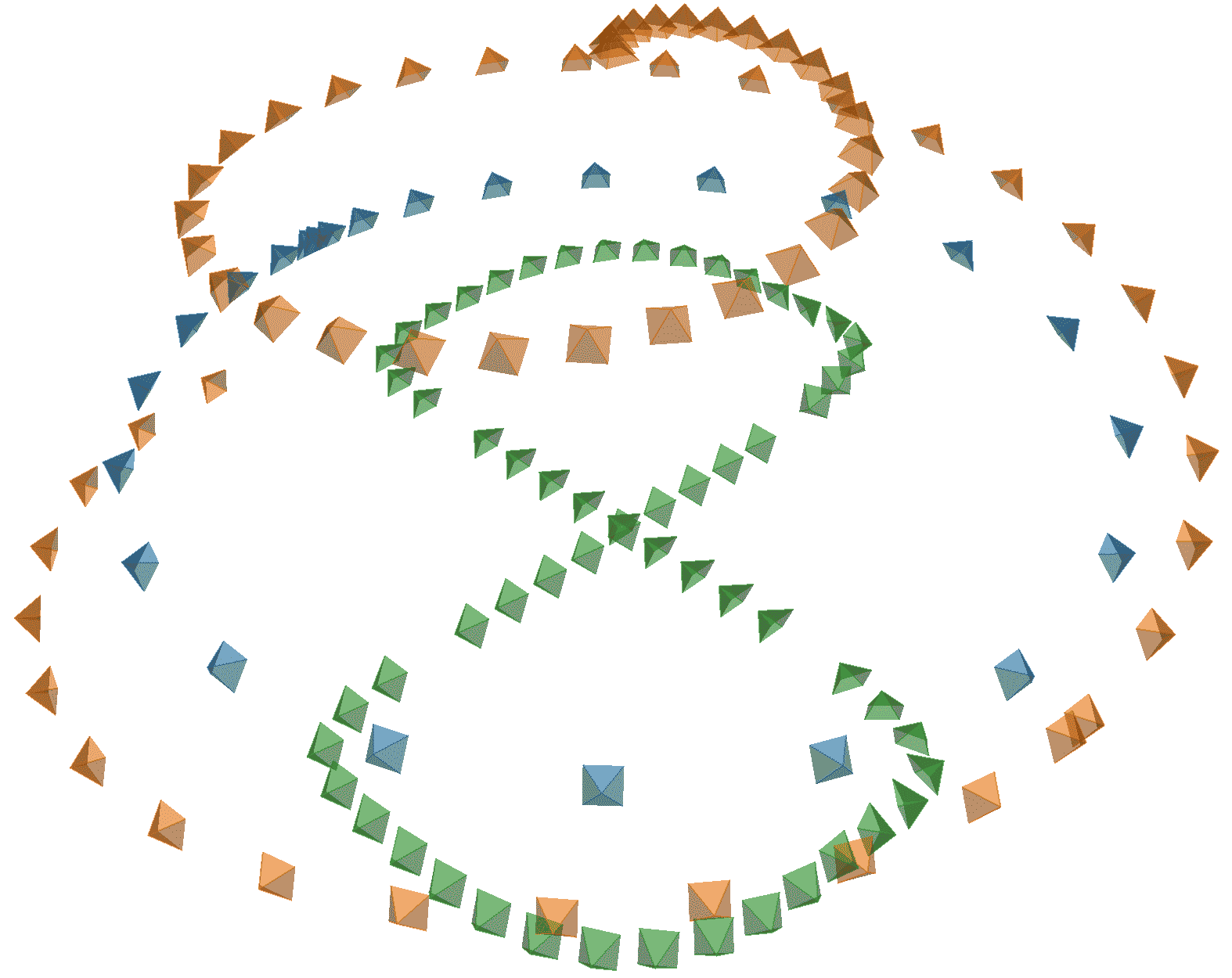}
\caption{Visualization of our designed camera paths.
    Blue cameras represent the circle path, green cameras belong to the mirror-passing-through path,
    and orange cameras represent the spiral path.}
  \label{fig:vis_camera}
\end{figure}

As summarized in \secref{sec:exist_dataset}, there lacks a 360-degree dataset consisting of complex reflection and refraction to facilitate the related research.
Therefore, we collect a 360-degree dataset comprising \SNumber synthetic scenes and \RNumber real captured scenes.

For our synthesized part shown in \figref{fig:syn_dataset}, we use an open source software Blender~\cite{blender}, and design our scenes with 3D models from BlenderKit, a community for sharing 3D models.
As our dataset consists of complete scenes instead of single objects, we design three kinds of camera paths.
The simplest one is a circle path, where we fix the height of our camera position with the camera looking at the center of the scene and moving the camera around a circle to render the whole scene.
We render this path for all our scenes, where we uniformly sample 120 points along the circle and randomly choose 100 images for the training set, 10 for the validation set, and 10 for the test set.
Besides, we design a 360-degree spiral path, where cameras gradually spiral up from the equator to the pole, 
looking at the center of the scene.
To further evaluate the robustness of future related methods, we design a novel mirror-passing-through path,
where the cameras move through the mirrors in the scenes back and forth.
We select 5 simple scenes and 5 difficult scenes from our dataset to render the above two paths, where we uniformly sample 300 points along the path and randomly choose 100 images as the training set and 200 as the test set.
We visualize the three kinds of camera paths in \figref{fig:vis_camera}.

\begin{figure}[t]
  \centering
  \includegraphics[width=0.99\linewidth]{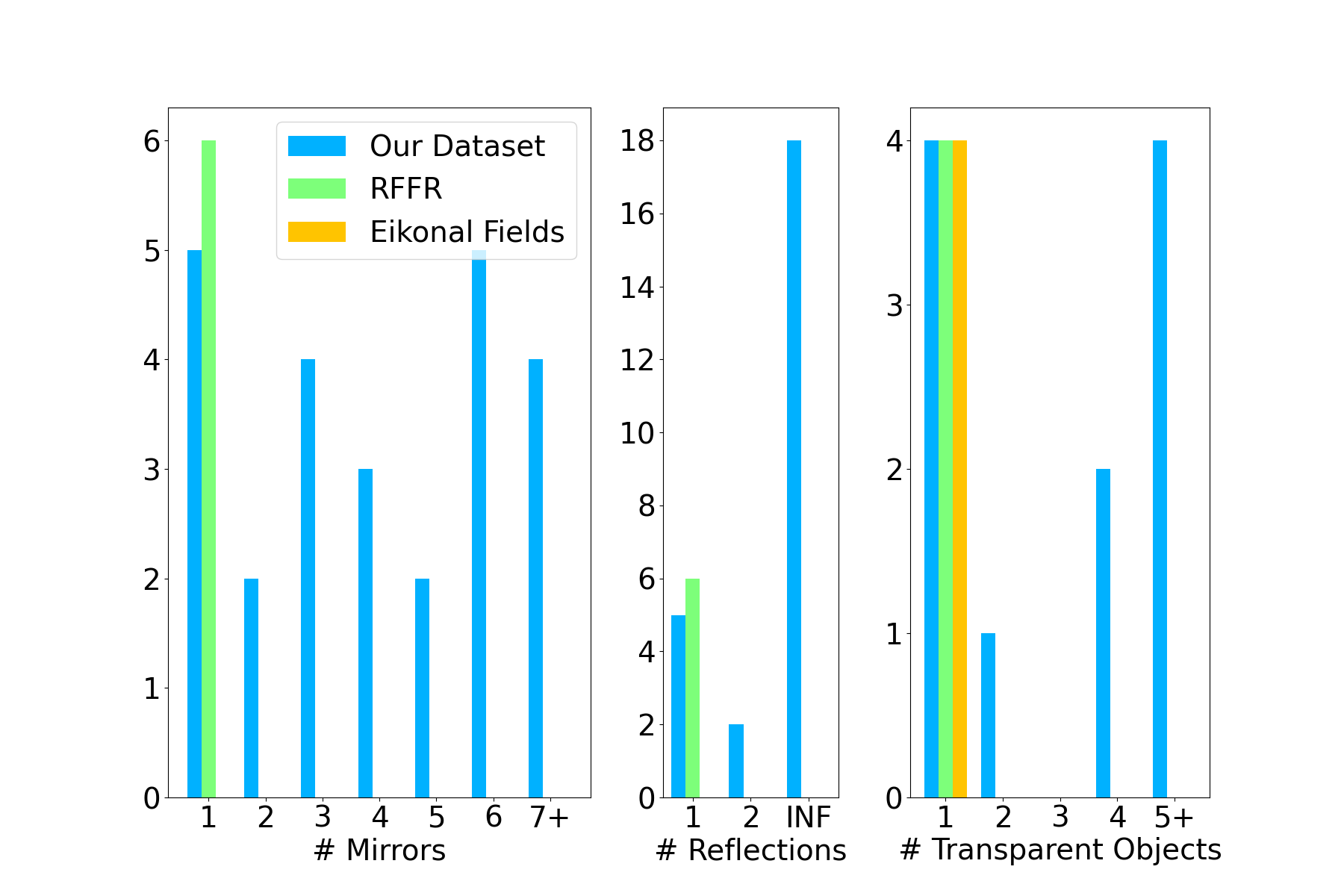}
\caption{We quantitatively compare our dataset with two most related datasets, 
    \ie, RFFR~\cite{guo2022nerfren} and EikonalFields~\cite{bemana2022eikonal}.
    We statistically count the number of mirrors and transparent objects, 
    which are directly related to reflection and refraction, respectively.
    Furthermore, we quantify the complexity of light paths 
    by the maximum number of reflections occurring in the scene.
    Note that in our scenes with more than two mirrors, 
    there must exist two facing mirrors; therefore, 
    the maximum reflections jump from 2 to infinite.
    }
  \label{fig:data_property}
\end{figure}

The constructed dataset features a wide variety of scenes containing reflective and refractive objects. We include a variety of complexity of light paths, controlled by the number and the layout of the mirror(s) in the scene, where the number of mirrors ranges from 1 up to tens of small pieces.
Note that even a scene in our dataset with only one mirror is more challenging than RFFR~\cite{guo2022nerfren}, as our camera moves from the front to the back of the mirror(s).
Besides, we also construct rooms with mirror walls that can essentially be treated as unbounded scenes, where we add mirrors in the center of the room and create unbounded virtual images.
We further build challenging scenes, including a combination of reflection and refraction.
As shown in \figref{fig:data_property}, our dataset exhibits much more challenging properties.
Furthermore, we render instance-level masks for each reflective and refractive object in our synthesized scenes.

We also include \RNumber captured real scenes with complex light conditions shown in \figref{fig:real_dataset}.
We construct our scenes using two mirrors, one glass ball with a smooth surface, one glass ball with a diamond-like surface, a few toys, and a few books.
We capture pictures randomly with 360-degree viewpoints. 

%%%%%%%%%%%%%%%%%dataset%%%%%%%%%%%%%%%%%%

%%%%%%%%%%%%%%%%%experiments%%%%%%%%%%%%%%%%%%

\section{Experiments}
\label{sec:exp}

\begin{table}[t]
  \centering
  \begin{subtable}[t]{\linewidth}
    \tablestyle{1pt}{1}
    \begin{tabular}{ccccc} \toprule
      & PSNR$\uparrow$ & SSIM$\uparrow$ & LPIPS$\downarrow$ & \# Params \\
      \midrule
      NeRF                 & 23.79/35.24 & 0.662/\underline{0.902} & 0.240 &   1.159M \\
      $\textrm{MS-NeRF}_S$ & 26.59/35.52 & 0.737/0.900 & \underline{0.232} &   1.201M \\
      $\textrm{MS-NeRF}_M$ & \underline{26.71}/\underline{35.72} & \underline{0.741}/0.901 & \textbf{0.226} &   1.245M \\
      $\textrm{MS-NeRF}_B$ & \textbf{26.95}/\textbf{35.87} & \textbf{0.748}/\textbf{0.903} & \textbf{0.226} &   1.311M \\
      \midrule
      Mip-NeRF & 24.47/35.97 & 0.693/0.906 & 0.245 &   0.613M  \\
      Ref-NeRF & 25.58/36.65 & 0.716/\textbf{0.911} & \textbf{0.210} &   0.713M  \\
      $\textrm{MS-Mip-NeRF}_S$ & 28.08/36.60 & 0.779/0.906 & 0.224 &   0.634M  \\
      $\textrm{MS-Mip-NeRF}_M$ & \underline{28.44}/\underline{36.76} & \underline{0.788}/\underline{0.907} & 0.222 &   0.656M  \\
      $\textrm{MS-Mip-NeRF}_B$ & \textbf{28.48}/\textbf{36.83} & \textbf{0.789}/\underline{0.907} & \underline{0.220} &   0.689M  \\
      \midrule
      Mip-NeRF 360 & \underline{24.20}/\underline{37.06} & \underline{0.733}/\textbf{0.925} &  \textbf{0.150}   &  9.007M  \\
      MS-Mip-NeRF 360  & \textbf{28.35}/\textbf{37.65} & \textbf{0.822}/\underline{0.923} &  \textbf{0.150}   &  9.007M  \\
      \bottomrule
    \end{tabular}
    \caption{MS module from \secref{subsec:module} on our synthetic dataset with circle paths.}
    \label{tab:main_res} 
  \end{subtable}
  \\ \vspace{7pt}
  \begin{subtable}[t]{\linewidth}
    \tablestyle{2pt}{1}
    \begin{tabular}{ccccc}
      \toprule
           & PSNR$\uparrow$ & SSIM$\uparrow$ & LPIPS$\downarrow$ & \# Params \\
      \midrule
      TensoRF & \underline{24.25}/\underline{33.83} & \underline{0.697}/\underline{0.937} & \underline{0.196} & {\scriptsize 17.34$|$4.00e-2M} \\
      MS-TensoRF & \textbf{26.96}/\textbf{37.74} & \textbf{0.767}/\textbf{0.951} & \textbf{0.147} & {\scriptsize 17.34$|$4.60e-2M} \\
      \midrule
      iNGP & \underline{24.04}/\underline{29.98} & \underline{0.743}/\underline{0.915} & \underline{0.201} & {\scriptsize 14.23$|$3.28e-2M} \\
      MS-iNGP & \textbf{26.59}/\textbf{33.67} & \textbf{0.800}/\textbf{0.930} & \textbf{0.157} & {\scriptsize 14.23$|$3.33e-2M} \\
      \bottomrule
    \end{tabular}
    \caption{MS module from \secref{subsec:grid_module} on our synthetic dataset with circle paths.}
    \label{tab:grid_res} 
  \end{subtable}
  \\ \vspace{7pt}
  \begin{subtable}[t]{\linewidth}
    \tablestyle{1pt}{1}
    \begin{tabular}{ccccc}
      \toprule
           & PSNR$\uparrow$ & SSIM$\uparrow$ & LPIPS$\downarrow$ & \# Params \\
      \midrule
      Mip-NeRF 360  & \underline{25.21}/\underline{39.15} & \underline{0.795}/\textbf{0.969} & \underline{0.102} & 9.007M  \\
      MS-Mip-NeRF 360 & \textbf{31.31}/\textbf{39.98} & \textbf{0.894}/\underline{0.966} & \textbf{0.098} & 9.052M \\
      \midrule
      Mip-NeRF 360  & \underline{24.87}/\underline{34.02} & \underline{0.828}/\textbf{0.915} & \textbf{0.184} & 9.007M  \\
      MS-Mip-NeRF 360 & \textbf{26.74}/\textbf{34.03} & \textbf{0.851}/\underline{0.910} & \underline{0.188} & 9.052M \\
      \bottomrule
    \end{tabular}
    \caption{MS module from \secref{subsec:module} on our synthetic dataset with spiral paths 
      and mirror-passing-through paths. The first two rows are spiral paths,
      and the last two rows are mirror-passing-through paths.
    }\label{tab:main_res_other_paths}
  \end{subtable}
  \\ \vspace{7pt}
  \begin{subtable}[t]{\linewidth}
    \tablestyle{5.5pt}{1}
    \begin{tabular}{ccccc}
      \toprule
           & PSNR$\uparrow$ & SSIM$\uparrow$ & LPIPS$\downarrow$ & \# Params \\
      \midrule
      Mip-NeRF 360  &  \underline{26.70} &  \underline{0.889} &  \textbf{0.113} &  9.007M  \\
      MS-Mip-NeRF 360 &  \textbf{28.14} &  \textbf{0.891} &  \underline{0.119} &   9.052M \\
      \bottomrule 
    \end{tabular}
    \caption{Comparisons on the real captured part of our dataset.}
    \label{tab:comp_realcap}
  \end{subtable}
  \\ \vspace{7pt}
  \begin{subtable}[t]{\linewidth}
    \tablestyle{8.5pt}{1}
    \begin{tabular}{ccccc} \toprule
           & PSNR$\uparrow$ & SSIM$\uparrow$ & LPIPS$\downarrow$ & \# Params \\
      \midrule
      NeRFReN  &  \underline{35.26} &  \underline{0.940} &  \underline{0.081} &  1.264M  \\
      $\textrm{MS-NeRF}_T$ &  \textbf{35.93} &  \textbf{0.948} &  \textbf{0.066} &   1.295M \\
      \bottomrule
    \end{tabular}
    \caption{Comparisons on RFFR dataset.}
    \label{tab:comp_nerfren}
  \end{subtable} \\
  \caption{Quantitative comparisons with existing methods. 
  Parameters with the pattern ``A$|$B'' refer to grid parameters and the MLP parameters;
  others are all MLP parameters. For PSNR and SSIM metrics on our synthesized dataset, 
  we split the rendered images according to the masks of reflective and refractive areas,
  and the metrics are reported as ``reflective and refractive areas/other areas''.
  }
  \vspace{-5pt}
  \label{tab:main_tab}
\end{table}

\begin{table}[t!]
  \tablestyle{6pt}{1}
  \begin{tabular}{ccccc}\toprule
    & PSNR$\uparrow$ & SSIM$\uparrow$ & LPIPS$\downarrow$ & \# Params\\ \midrule
    Mip-NeRF  &  \underline{30.74} &  \underline{0.942} &  \textbf{0.047} & 1.264M  \\
    $\textrm{MS-Mip-NeRF}_B$ &  \textbf{30.81} &  \textbf{0.943} &  \textbf{0.047} & 1.295M \\
    \midrule
    Mip-NeRF  &  \textbf{25.78} &  \textbf{0.775} &  \textbf{0.213} & 1.264M  \\
    $\textrm{MS-Mip-NeRF}_B$ &  \underline{25.59} &  \underline{0.764} &  \underline{0.223} & 1.295M \\
    \bottomrule  
  \end{tabular}
  \caption{Results on the Realistic Synthetic $360^{\circ}$ dataset 
    (first two rows) and Real Forward-Facing dataset (last two rows).
  }\label{tab:nerf_dataset}
\end{table}

\subsection{Baselines, Hyperparameters, and benchmarks}
\label{sec:methods_details}

\begin{figure}
  \begin{subfigure}{0.49\linewidth}
    \includegraphics[width=\linewidth]{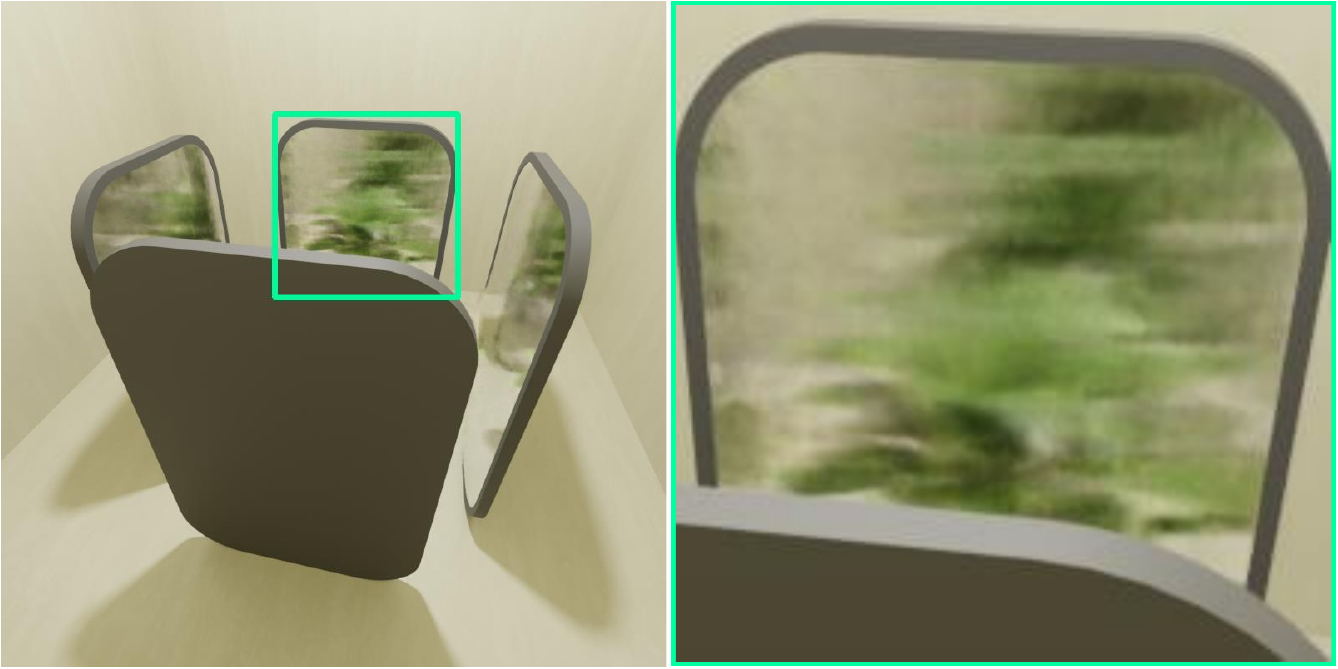}
    \caption{Mip-NeRF 360}
  \end{subfigure}
  \hfill
  \begin{subfigure}{0.49\linewidth}
    \includegraphics[width=\linewidth]{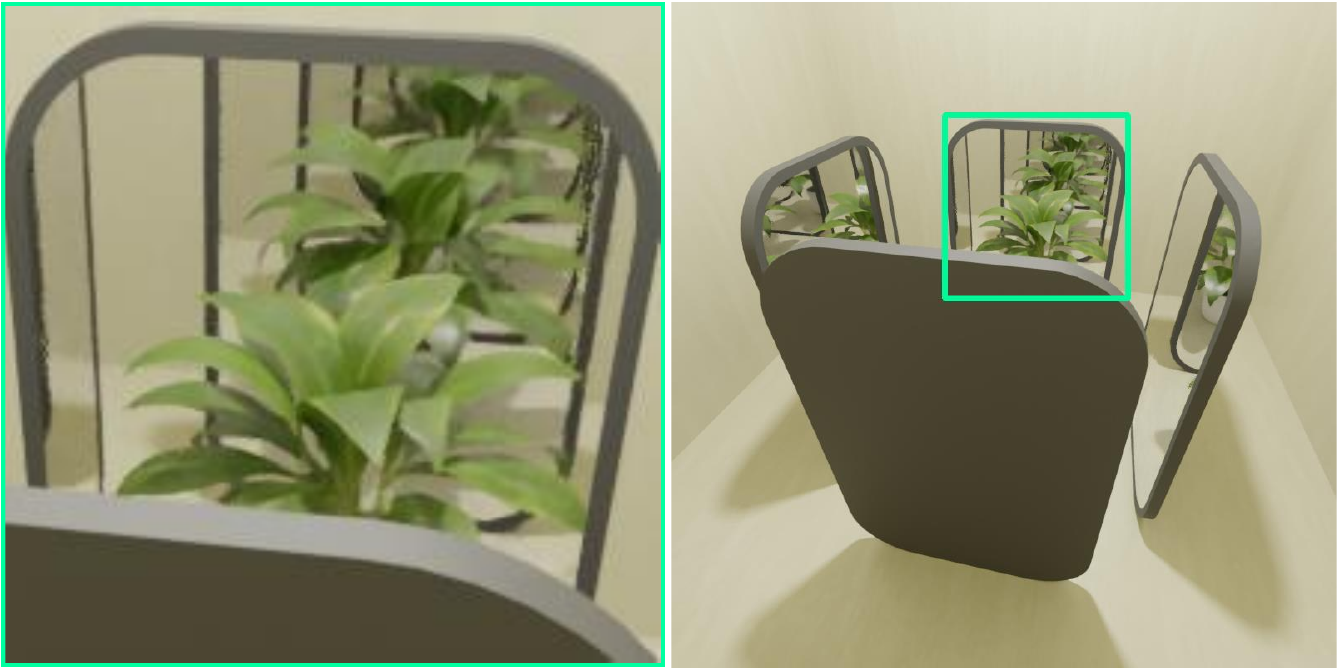}
    \caption{MS-Mip-NeRF 360}
  \end{subfigure}
  \vspace{-5pt}
  \caption{Visual comparison between Mip-NeRF 360 and MS-Mip-NeRF 360. 
    Our module can extend Mip-NeRF 360 to model unbounded virtual scenes.
  }\vspace{-5pt}
  \label{fig:vis_4m}
\end{figure}

To thoroughly evaluate the superiority and robustness and demonstrate the potential of our method, 
we conduct various experiments based on different datasets with different baselines and 
our modules of different scales. 
We select 5 representative NeRF-based methods and categorize them into two parts:
a) MLP-based NeRF methods, including NeRF~\cite{mildenhall2021nerf}, Mip-NeRF~\cite{barron2021mip} and Mip-NeRF 360~\cite{barron2022mip}; 
b) grid-based NeRF method, including TensoRF~\cite{Chen2022ECCV} and iNGP~\cite{InstantNGP}.
MLP-based NeRF methods contain representative methods for novel view synthesis, which typically encode 
the underlying radiance fields into the weights of MLP and aim at rendering high-quality images.
We conduct experiments with our MLP-based MS module in \secref{subsec:module} based on these methods on our dataset to demonstrate the superiority of our scheme.
Grid-based NeRF methods feature hybrid representation, specifically combinations of small MLP 
and discrete learnable/fixed parameters organized in 3D/2D grids,
which heavily reduces the parameters to be optimized in each iteration
and converge to relatively high-quality rendering in a very short time.
TensoRF and iNGP construct the networks using learnable parameters 
in 2D and 3D grids respectively with small MLPs;
therefore, we integrate our hybrid MS module described 
in \secref{subsec:grid_module} into them to validate our methods.
We conduct various experiments with these baselines on different datasets 
to demonstrate the superiority and generalization of our scheme.

As our MLP-based MS module is quite simple, we can scale our module by three hyperparameters, which we refer to as $K$ for the sub-space number, $d$ for the dimension of output features, and $h$ for the hidden layer dimension of Decoder MLP and Gate MLP, respectively.
For the hybrid MS scheme, there are also hyperparameters $K$, $d$, and $h$, except that $d$ and $h$ only control the Gate MLP.
All the training details can be found in the supplementary.
We report our results with three commonly used metrics: PSNR, SSIM~\cite{wang2004image}, and LPIPS~\cite{zhang2018unreasonable}.

\subsection{Experiments on MLP-based NeRF methods}
\label{sec:nerf_details}

We scale our module from \secref{subsec:module} and integrate it with NeRF~\cite{mildenhall2021nerf}, Mip-NeRF~\cite{barron2021mip}, and 
Mip-NeRF 360~\cite{barron2022mip} to conduct experiments, and we follow most default settings from 
\cite{mildenhall2021nerf,barron2021mip,barron2022mip,guo2022nerfren,verbin2022refnerf}, 
except that we use 1024 rays per batch and train 200k iterations for all experiments on all scenes.

For NeRF~\cite{mildenhall2021nerf} and Mip-NeRF~\cite{barron2021mip} based experiments, we build $\textrm{MS-NeRF}_S$ and $\textrm{MS-Mip-NeRF}_S$ with hyperparameters $\{K=6, d=24, h=24\}$, similarly $\textrm{MS-NeRF}_M$ and $\textrm{MS-Mip-NeRF}_M$ with hyperparameters $\{K=6, d=48, h=48\}$, and $\textrm{MS-NeRF}_B$ and $\textrm{MS-Mip-NeRF}_B$ with hyperparameters $\{K=8, d=64, h=64\}$.
For Mip-NeRF 360~\cite{barron2022mip} based experiments, we construct MS-Mip-NeRF 360 with hyperparameters $\{K=8, d=32, h=64\}$.
Moreover, we provide a comparison with Ref-NeRF~\cite{verbin2022refnerf} because it uses Mip-NeRF as a baseline and possesses an outstanding ability to model glossy materials.
As NeRF, Mip-NeRF, and Ref-NeRF are all designed for bounded scenes, 
we evaluate these methods on the first 25 scenes with circle paths while the last 8 scenes are equivalent to unbounded scenes.
For Mip-NeRF 360 based experiments, we evaluate on all scenes with all paths.

%%
%\textbf{(c)} To show the \textbf{generalization}, 
We also compare our method with NeRFReN on the RFFR dataset~\cite{guo2022nerfren}.
NeRFReN is a specially designed two-branch network based on vanilla NeRF for mirror-like surfaces in forward-facing scenes. Thus we construct a tiny version of our method, referred to as $\textrm{MS-NeRF}_T$, based on NeRF with hyperparameters $\{K=2, d=128, h=128\}$.
Here we use two sub-spaces as NeRFReN tries to decompose reflective surfaces into two parts, and we want to show that our space decomposition is more effective.
For a fair comparison, we re-train NeRFReN using the official code under the provided settings on the RFFR dataset, except that we set the number of the used mask to 0 as our method requires no extra mask.

\begin{figure}
  \begin{subfigure}{0.49\linewidth}
    \includegraphics[width=\linewidth]{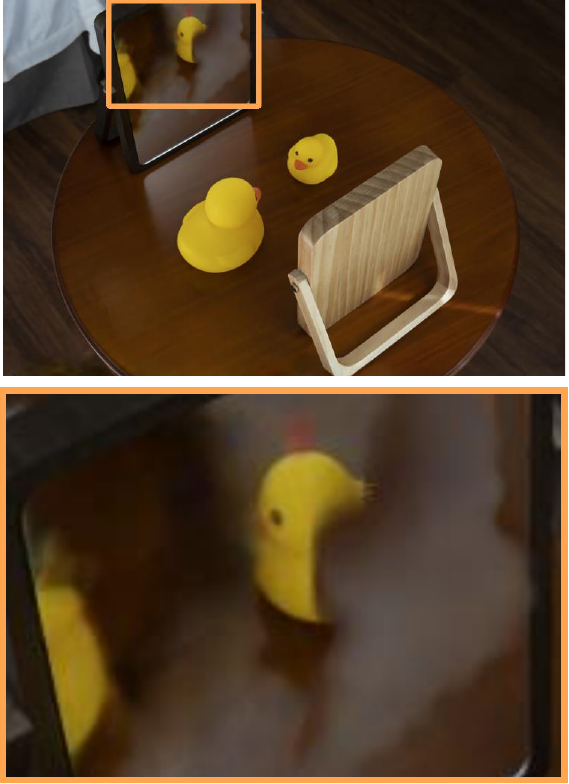}
    \caption{Mip-NeRF 360}
  \end{subfigure}
  \hfill
  \begin{subfigure}{0.49\linewidth}
    \includegraphics[width=\linewidth]{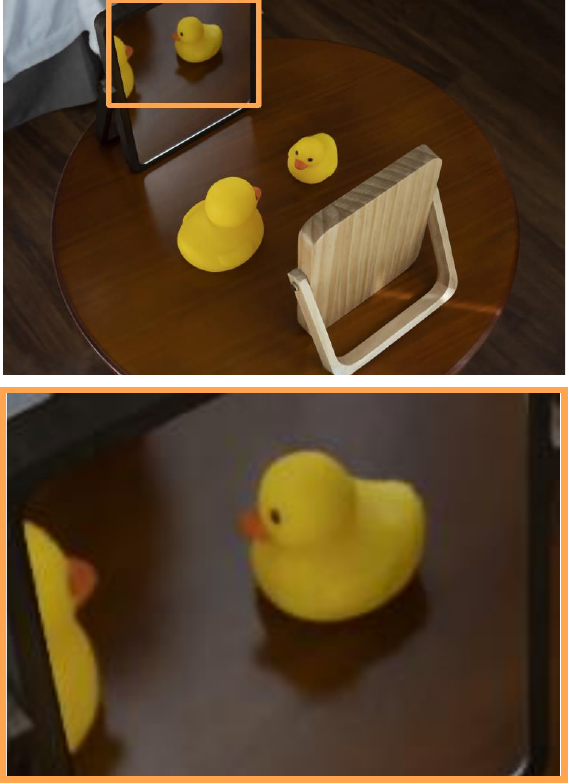}
    \caption{MS-Mip-NeRF 360}
  \end{subfigure}
  \vspace{-5pt}
  \caption{Visual comparison between Mip-NeRF 360 and MS-Mip-NeRF 360 on the real captured part of our dataset. Our method is robust enough to recover virtual images in the real world.}
  \label{fig:vis_real}
  \vspace{-5pt}
\end{figure}

\subsection{Experiments on grid-based NeRF methods}
\label{sec:other_details}

We carry out experiments on grid-based methods to demonstrate the compatibility 
of our scheme with other applications instead of just pure MLP-based rendering methods.

iNGP~\cite{InstantNGP} and TensoRF~\cite{Chen2022ECCV} achieve fast convergence 
using combinations of learnable 3D/2D feature grids and very small MLPs.
We separately construct the MS-iNGP and MS-TensoRF with our hybrid MS module 
proposed in \secref{subsec:grid_module} using hyperparameters $\{K=4, d=8, h=32\}$,
and we initialize the extra branch with a similar MLP but of smaller size to the one in the main branch.
We follow all the default settings~\cite{Chen2022ECCV} for MS-TensoRF experiments,
and we implement the MS-iNGP based on the implementation from \cite{li2023nerfacc} with their proposal estimator.
We evaluate TensoRF on the first 25 scenes with circle paths, as this method is designed for bounded scenes.
On the contrary, iNGP with the proposal estimator~\cite{li2023nerfacc} is able to reconstruct unbounded scenes,
therefore, we evaluate iNGP-related models on all scenes with circle paths.

\subsection{Comparisons}
\label{sec:main_res}

\begin{figure}
  \begin{subfigure}{0.49\linewidth}
    \includegraphics[width=\linewidth]{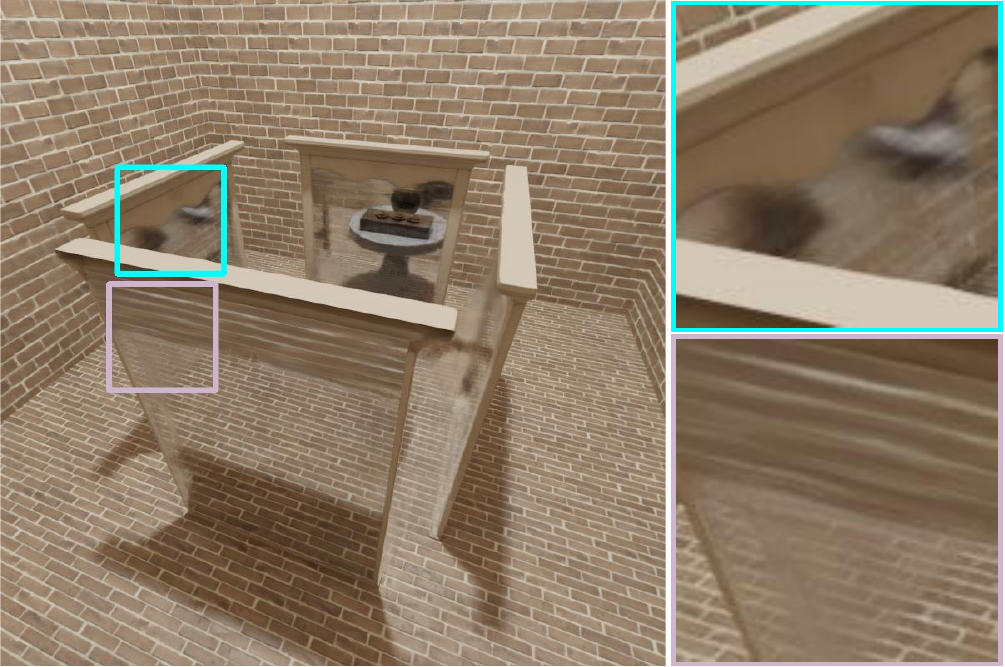}
    \caption{Ref-NeRF}
  \end{subfigure}
  \hfill
  \begin{subfigure}{0.49\linewidth}
    \includegraphics[width=\linewidth]{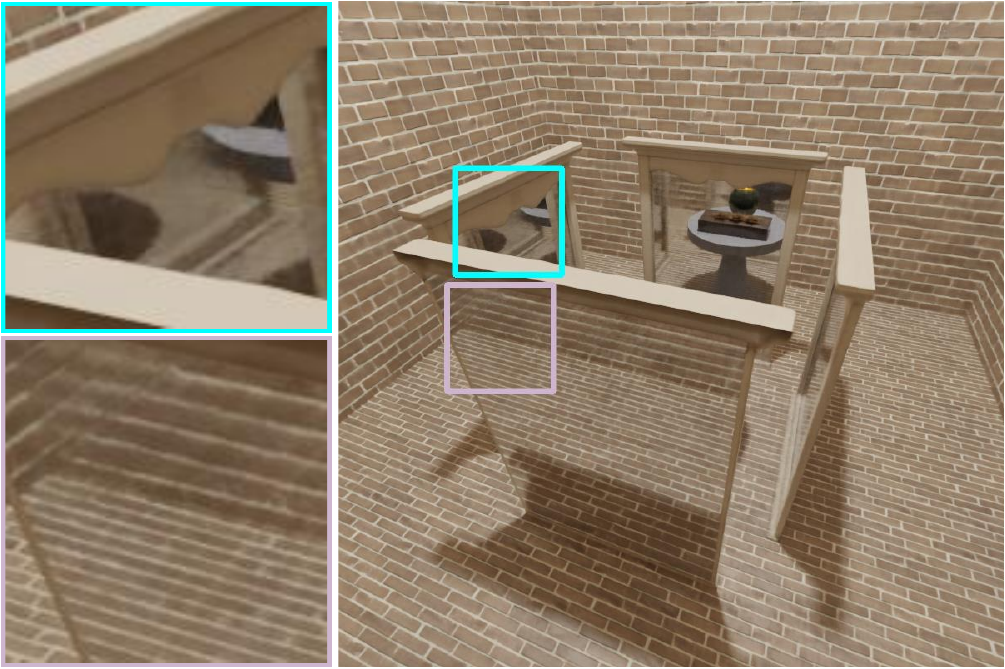}
    \caption{$\textrm{MS-Mip-NeRF}_B$}
  \end{subfigure}
  \vspace{-5pt}
  \caption{Visual comparison between $\textrm{MS-Mip-NeRF}_B$ and Ref-NeRF. Our method significantly outperforms Ref-NeRF on reflective surfaces.}
  \vspace{-10pt}
  \label{fig:vis_v6}
\end{figure}

\myPara{Quantitative comparisons}
As reported in \tabref{tab:main_res}, our MLP-based module can be integrated into most NeRF-like models and improve performance by a large margin with minimal extra cost introduced.
Especially in Mip-NeRF 360-based experiments, our module exhibits better results of 4.15 dB improvement in PSNR using merely 0.5\% extra parameters
without degrading the performance on non-mirror regions.
Besides, our Mip-NeRF-based models also outperform Ref-NeRF~\cite{verbin2022refnerf} by a large margin on mirror, 
which is a variant based on Mip-NeRF with the outstanding ability to model glossy materials.

The synthetic part with mirror-passing-through paths and spiral paths and the real captured scenes are much more challenging, 
therefore, we demonstrate our results compared with the pure MLP-based \sArt method Mip-NeRF 360 in \tabref{tab:main_res_other_paths} 
and \tabref{tab:comp_realcap} on these benchmarks.
Our approach also shows large improvements,
which indicates the robustness of our scheme.
Specifically, the spiral camera paths introduce wider view changes on the virtual images,
while the mirror-passing-through camera paths introduce rapid disappearance of virtual images along the rays,
and the results indicate that our scheme is capable of handling these without degrading the performance on non-mirror regions.

\begin{figure}[t]
  \centering
  
  \begin{subfigure}{0.49\linewidth}
    \includegraphics[width=\linewidth]{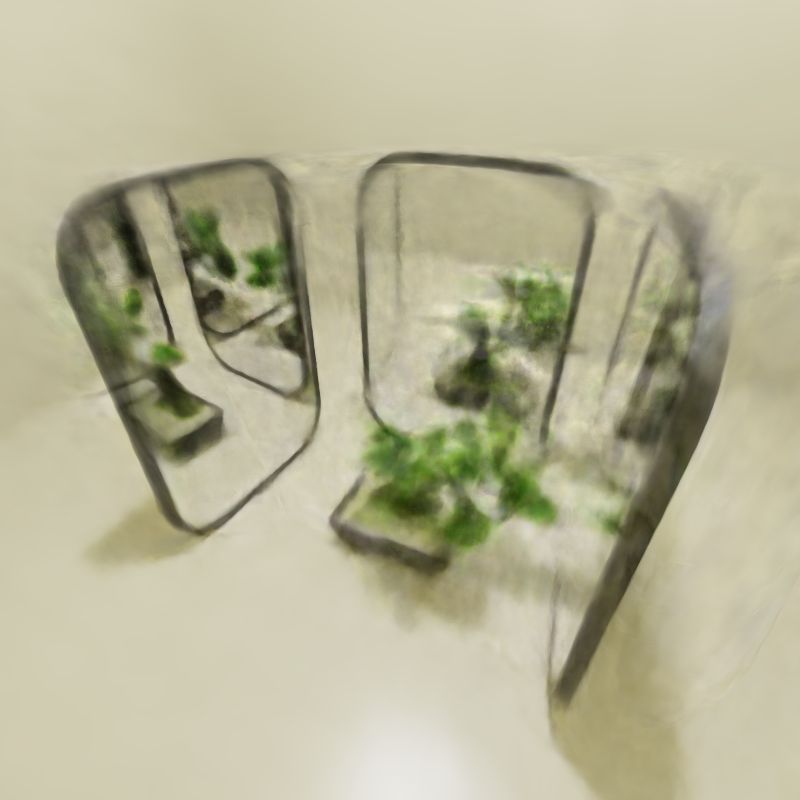}
    \caption{NeRFReN}
  \end{subfigure}
  \hfill
  \begin{subfigure}{0.49\linewidth}
    \includegraphics[width=\linewidth]{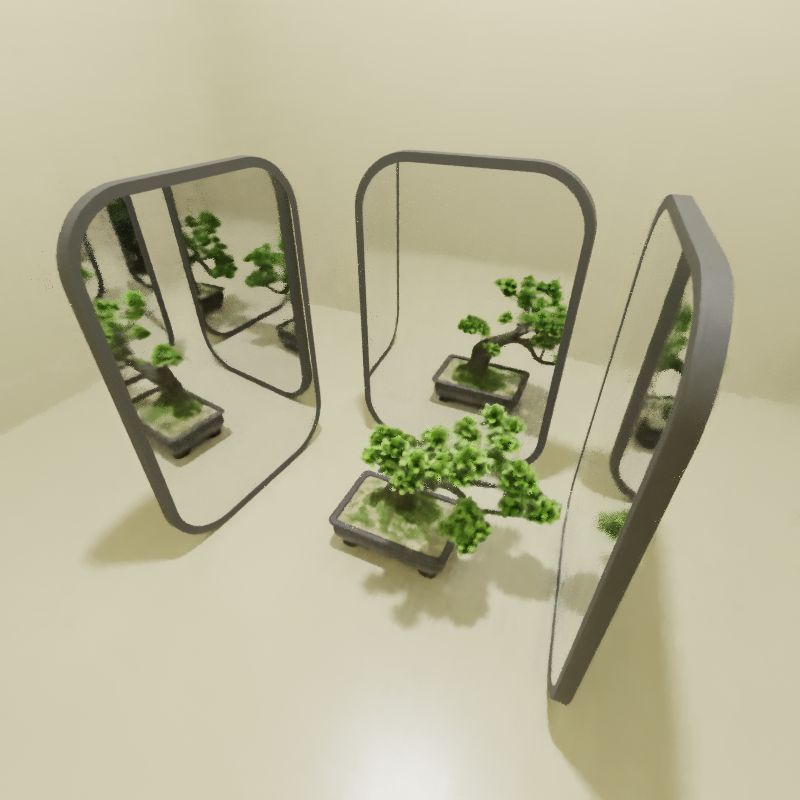}
    \caption{$\textrm{MS-NeRF}_B$}
  \end{subfigure}
  \vspace{-5pt}
  \caption{(a) Trained with accurately labeled masks, NeRFReN even fails to render ordinary parts of the scene in 360-degree scenes with mirrors. 
     (b) Our method requires no extra manually labeled masks and renders high-quality images.
  }
  \label{fig:comp_nerfren}
\end{figure}

Results in \tabref{tab:grid_res} indicate that 
our hybrid MS module is compatible with TensoRF and iNGP.
As shown, our module can improve the performance on reflective regions
and also help stabilize the reconstruction on other regions.

On the RFFR dataset, 
which contains forward-facing reflective surfaces in the scenes, 
our approach achieves better results when no manually labeled masks 
are provided in training as in \tabref{tab:comp_nerfren}.
For NeRFReN, we re-train the model using the official code 
following the provided setting, except the number of masks used 
for reflective surfaces is 0.

The Realistic Synthetic $360^{\circ}$ dataset and Real Forward-Facing dataset are first introduced in \cite{mildenhall2021nerf},
which are commonly used for evaluating the ability of NeRF-based methods in novel view synthesis.
We also train Mip-NeRF and $\textrm{MS-Mip-NeRF}_B$ on these datasets because Mip-NeRF is a commonly used backbone for NeRF-based methods.
The results are reported in \tabref{tab:nerf_dataset},
which demonstrate that our multi-space module has no influence on the representation ability of NeRF-based methods on common materials.

All of the above experiments demonstrate the superiority and compatibility of our method.

\begin{figure*}[t]
  \centering
  \begin{subfigure}{\linewidth}
    \centering
    \begin{overpic}[width=\linewidth]{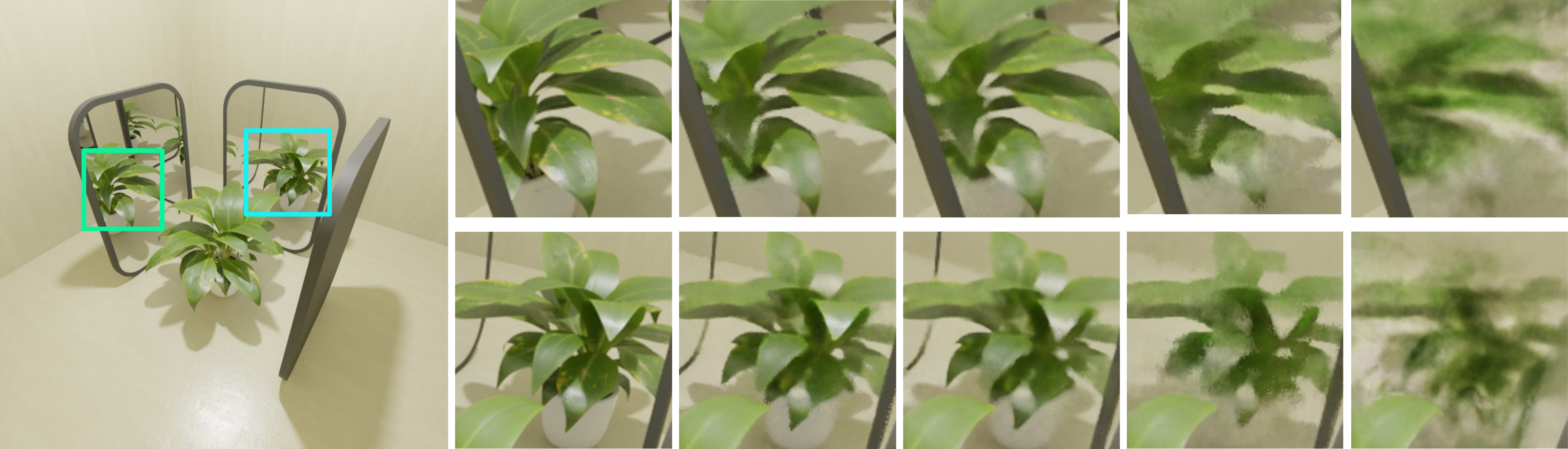} \small
      \put(13, -2){(a)}
      \put(34.5, -2){(b)}
      \put(48.7, -2){(c)}
      \put(63, -2){(d)}
      \put(78, -2){(e)}
      \put(92, -2){(f)}
    \end{overpic}
  \end{subfigure}
  \vspace{10pt}
  \caption{
  Detailed comparisons on rendered images.
  (a) Overview; (b) Ground-Truth; (c) MS-Mip-NeRF 360; (d) $\textrm{MS-Mip NeRF 360}_{p}$; (e) $\textrm{MS-Mip NeRF 360}_{h}$; 
  (f) Mip-NeRF 360.}
  \label{fig:abla_cut}
  \vspace{-5pt}
\end{figure*}

\myPara{Qualitative comparisons and discussions} 
Besides quantitative comparisons, we summarize the advantages of our modules and support them by qualitatively or quantitatively comparing our methods with the corresponding baselines.

Qualitative comparisons with the state-of-the-art MLP-based method, Mip-NeRF 360, are shown in \figref{fig:mirror_through}, \figref{fig:vis_4m}, and \figref{fig:vis_real}. Our method renders high-fidelity virtual images, bounded and unbounded, in both synthetic and real-world scenes. 
Furthermore, our MS module is also robust in sudden changes when the camera moves through the mirror,
as in \figref{fig:mirror_through}.

A qualitative comparison with Ref-NeRF~\cite{verbin2022refnerf}, which understands virtual images as textures using the conventional NeRF backbone, is shown in \figref{fig:vis_v6}. 
As Ref-NeRF is also based on Mip-NeRF, we compare our Mip-NeRF-based variant with Ref-NeRF using the same baseline and use comparable parameters (specifically ours 0.689M and Ref-NeRF 0.713M) in the comparison. 
Again, the qualitative results show our significant improvements in rendering reflective surfaces.

We also compare with the NeRFReN model, which requires accurately labeled masks of the reflective regions during training and handles forward-facing reflective surfaces only. 
In this comparison, we train their model on our synthesized dataset with extra accurate reflection masks provided. 
\figref{fig:comp_nerfren} shows that their model fails to recover 360-degree high-fidelity rendering while our approach succeeds.

\subsection{Ablation studies}
\label{sec:ablation}

\begin{table}[t]
  \centering
  \begin{subtable}[t]{\linewidth}
    \tablestyle{5.pt}{1}
    \begin{tabular}{ccccc}
      \toprule
           & PSNR$\uparrow$ & SSIM$\uparrow$ & LPIPS$\downarrow$ & \# Params \\
      \midrule
        Mip-NeRF 360   &  29.84 &  0.942 &  0.100 &   9.007M \\
        $\textrm{MS-Mip NeRF 360}_{h}$  &  33.59  &  0.958  &  0.081  &  9.060M  \\
        $\textrm{MS-Mip NeRF 360}_{p}$  &  \underline{37.50}  &  \underline{0.972}  &  \underline{0.064}  &  9.018M  \\
        MS-Mip NeRF 360 &  \textbf{38.84} &  \textbf{0.977} &  \textbf{0.054} &   9.052M \\
      \bottomrule
    \end{tabular}
    \caption{
    Ablation study on our MLP-based MS module architecture.}
    \label{tab:abla_naive}
  \end{subtable}
  \\ \vspace{6pt}
  \begin{subtable}[t]{\linewidth}
    \tablestyle{3.8pt}{1}
    \begin{tabular}{ccccc}
      \toprule
           & PSNR$\uparrow$ & SSIM$\uparrow$ & LPIPS$\downarrow$ & \# Params \\
      \midrule
        MS-TensoRF-grid   &  \underline{32.07} &  \underline{0.932} &  \underline{0.130} &   {\scriptsize 21.73$|$5.60e-2M} \\
        MS-TensoRF   &  \textbf{34.72} &  \textbf{0.938} &  \textbf{0.113} &   {\scriptsize 17.23$|$4.60e-2M} \\
      \bottomrule  
    \end{tabular}
  \caption{Ablation study on our hybrid MS module architecture.}
  \label{tab:comp_grid}
  \end{subtable}
  \caption{Ablation study on our module architecture.
  Parameters with the pattern ``A$|$B'' refer to grid parameters and the MLP parameters;
  others are all MLP parameters.}
  \vspace{-10pt}
\end{table}

In this section, we evaluate the design of our module and explore the relation between the number of sub-spaces and the number of virtual images.
\myPara{Ablation on using neural feature field}
We implement a module that simply outputs $K$ scalars $\{\sigma^k_i\}$ and $K$ RGB-G vectors $\{\mathbf{\hat{c}}^k_i\}$ of four dimensions, 
which consists of three RGB channels and one scaler that controls the gating information of multi-space composition.
When rendering, we use the same integral equation as NeRF to accumulate the $K$ RGB-G vectors $\{\mathbf{\hat{c}}^k_i\}$ and get the RGB-G maps of each sub-space,
then we split the RGB-G maps into RGB maps and sub-space composition maps(gating maps),
finally, we apply softmax over the gating maps along the sub-space channels and use it as the weight to compose sub-space RGB maps to form the final rendering results.
We integrate this design into Mip-NeRF 360 noted as $\textrm{MS-Mip NeRF 360}_{p}$, where we set $K=8$. 
We also integrate the hybrid module from \secref{subsec:grid_module} into Mip-NeRF 360 noted as $\textrm{MS-Mip NeRF 360}_{h}$,
the results are in \tabref{tab:abla_naive}.
We also exhibit a few visual results in \figref{fig:abla_cut}, 
which indicate that a simple multi-space radiance field assumption can help the model partially overcome the violation of reflections, 
but will also introduce the over-smoothing problem because of the lack of an efficient multi-space composition strategy.

\begin{figure}[t]  \small
  \begin{overpic}[width=.49\linewidth]{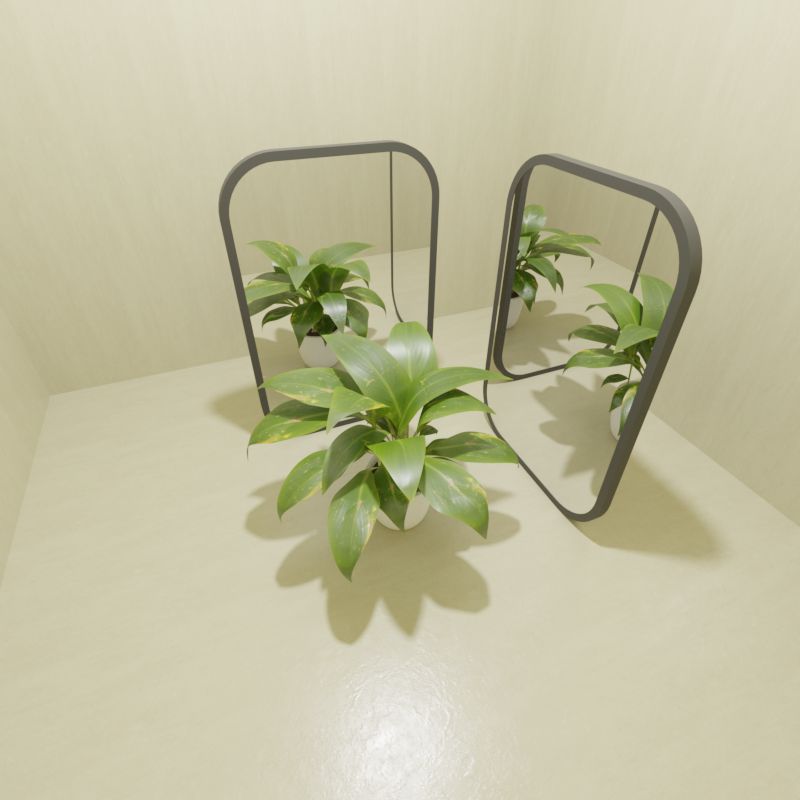} 
    \put(2,2){scene01}
  \end{overpic} \hfill
  \begin{overpic}[width=.49\linewidth]{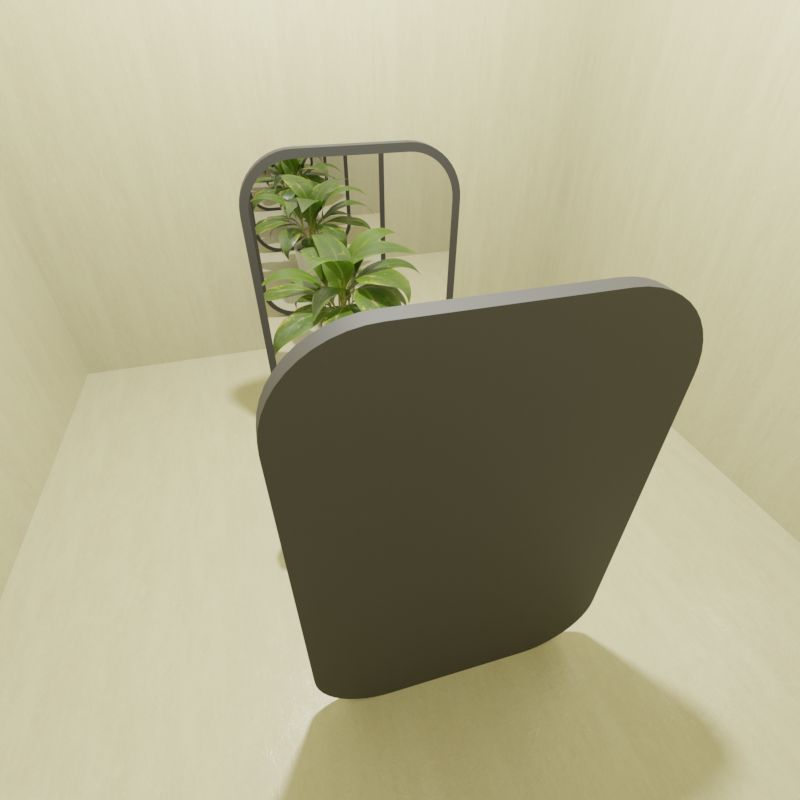} 
    \put(2,2){scene02}
  \end{overpic}
  \vspace{5pt} \\
  \begin{overpic}[width=\linewidth]{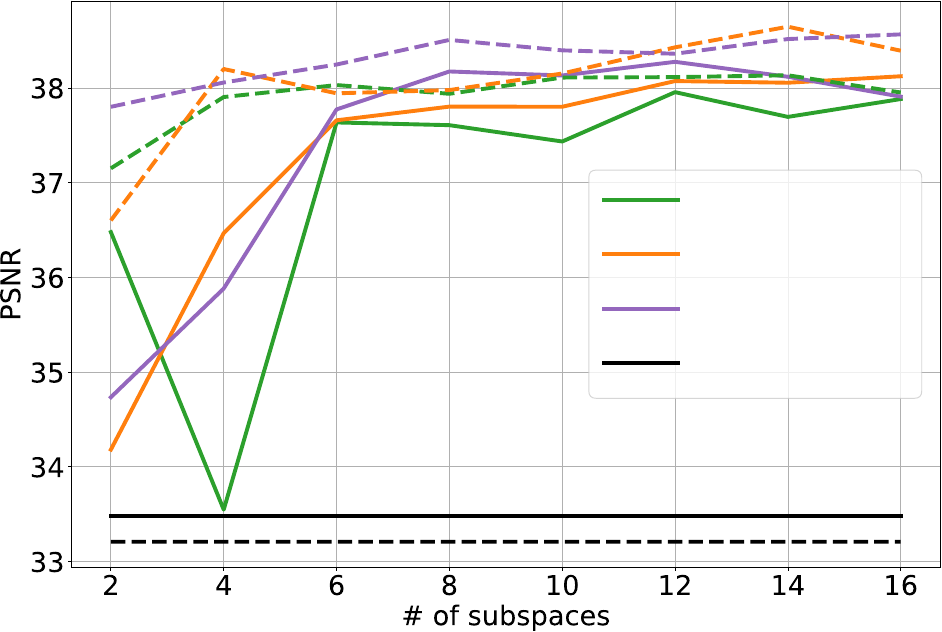} 
    \put(73, 44.7){$\textrm{MS-NeRF}_{d=24}$}
    \put(73, 39.0){$\textrm{MS-NeRF}_{d=48}$}
    \put(73, 33.3){$\textrm{MS-NeRF}_{d=64}$}
    \put(73, 27.5){NeRF}
  \end{overpic}
  \caption{We use PSNR to quantitatively evaluate the ablation experiments on
    scene01 and scene02 and plot the results with solid and dotted lines, respectively.
  }\label{fig:abla_num_res}
  \vspace{-5pt}
\end{figure}

\myPara{Ablation on the sub-space number}
In our Euclidean space, one can control the number of virtual sub-spaces by the number and the layout of the mirror(s).
For example, when two mirrors are facing each other, there could be infinitely recursive virtual image spaces, but when two mirrors are placed back against each other, there will be just one virtual image behind each mirror.
To provide a guideline for the design of our module, we choose two scenes consisting of two mirrors with different layouts from our synthesized part of the dataset and train NeRF-based variants of different sub-space numbers and different feature dimensions.
We construct our variants based on NeRF with the output feature dimensions $d\in \{24, 48, 64\}$ and the number of sub-spaces $K\in \{2, 4, 6, ..., 16\}$, then we train our models on the two scenes and report the results using PSNR.
Our results in \figref{fig:abla_num_res} show that the number of sub-spaces is not required to match the actual number of virtual image spaces, and 6 sub-spaces can guarantee stable learning for multi-space radiance fields. 
Moreover, feature fields with dimension $d=24$ already encode enough information for composition, but for stable performance $d=48$ is better.

\myPara{Ablation on the hybrid MS module}
We further design a grid-based MS module 
to provide a guideline for the MS module in grid-based NeRF methods.
Specifically, we replace the positional encoding $\gamma({\rm \mathbf{p}})$
in our hybrid MS module from \secref{subsec:grid_module}
with features queried by $Q(\mathbf{V}_{vis}, \mathbf{p})$,
where $\mathbf{V}_{vis}$ is a grid similar to the $\mathbf{V}$ 
in the main branch but of smaller size.
We refer to the model integrated with grid-based MS module as MS-TensoRF-grid,
and we conduct comparisons with the MS-TensoRF model in \secref{sec:other_details}
on the first 10 scenes in our synthesized scenes with circle paths.
As in \tabref{tab:comp_grid}, the results indicate that the multi-space scheme is more related to the positions,
therefore, our hybrid MS module reaches better performance with less cost.

\begin{figure}[t]
  \centering
  \begin{subfigure}{0.49\linewidth}
    \includegraphics[width=\linewidth]{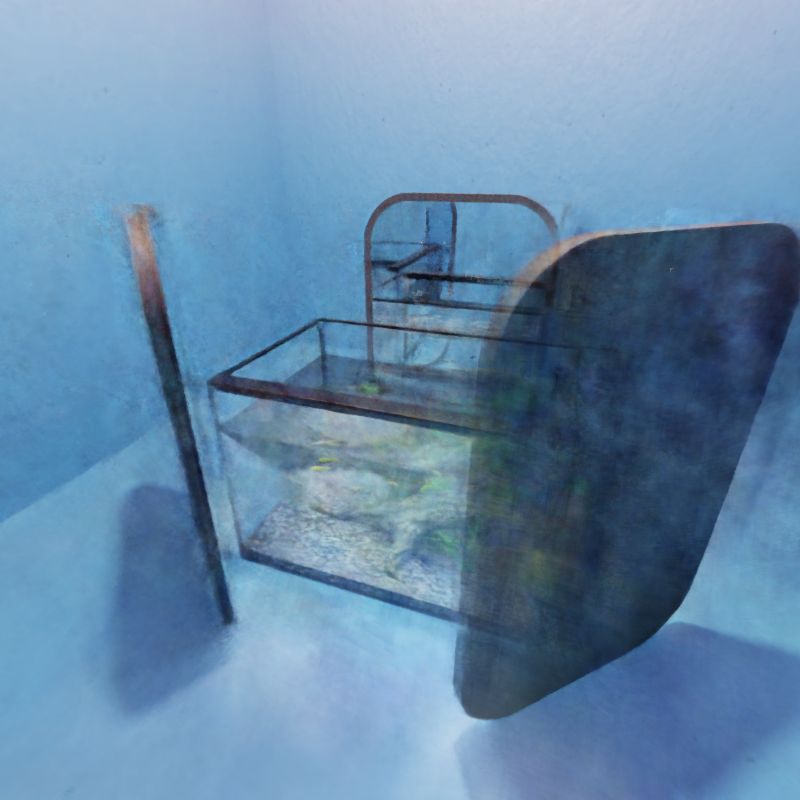}
    \caption{Rendered by NeuS}
  \end{subfigure}
  \hfill
  \begin{subfigure}{0.49\linewidth}
    \includegraphics[width=\linewidth]{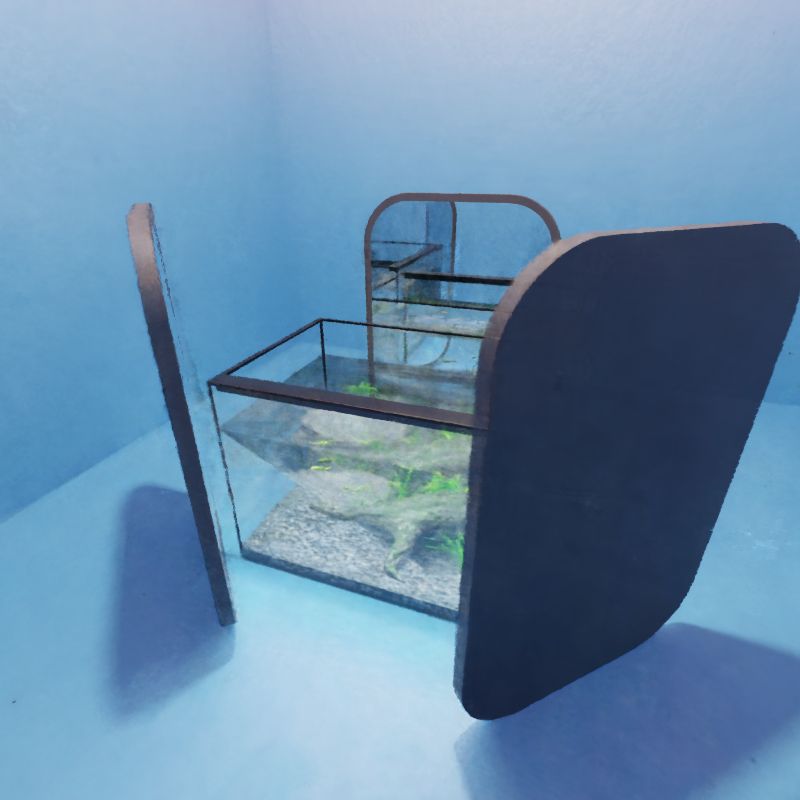}
    \caption{Rendered by MS-NeuS}
  \end{subfigure}
  \begin{subfigure}{0.49\linewidth}
    \includegraphics[width=\linewidth]{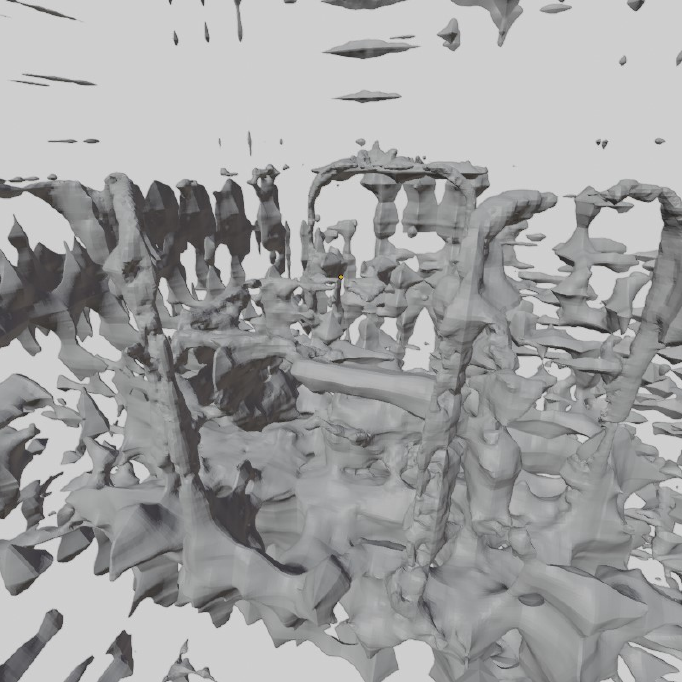}
    \caption{Mesh from NeuS}
  \end{subfigure}
  \hfill
  \begin{subfigure}{0.49\linewidth}
    \includegraphics[width=\linewidth]{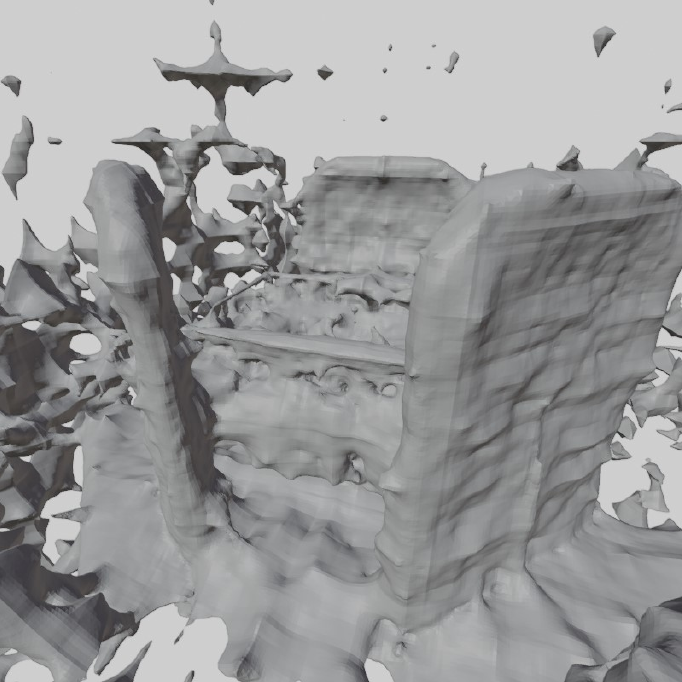}
    \caption{Mesh from MS-NeuS}
  \end{subfigure}
  \vspace{-5pt}
  \caption{
    Our MS module helps NeuS render high quality novel views for reflections,
    and the reconstructed geometry of mirrors is improved.
  }
  \label{fig:comp_neus}
\end{figure}

\myPara{Ablation on the input image number}
To further validate the robustness of our scheme,
we compare the MS-Mip-NeRF 360 model with the Mip-NeRF 360 baseline on Scene01-Scene05 from our synthesized dataset
with all three camera paths with 100, 75, or 30 randomly selected training views.
We report the detailed results in \tabref{tab:n_input_images}, which indicates that 
our scheme robustly improves the performance of the baseline model on reflective areas with varying numbers of input images.
We provide more visual comparisons in the Supplementary to better explore the behavior of our scheme.

\begin{table}[t]
  \centering
  \tablestyle{5.pt}{1}
  \begin{tabular}{ccccc}
    \toprule
        & PSNR$\uparrow$ & SSIM$\uparrow$ & LPIPS$\downarrow$ & \# Params \\
    \midrule
    NeuS   &  \underline{25.86}  &  \underline{0.831}  &  \underline{0.258}  &  {\scriptsize 27.96$|$0.16e-1M}  \\
    MS-NeuS   &  \textbf{28.65}  &  \textbf{0.859}  &  \textbf{0.195}  &  {\scriptsize 27.96$|$0.37e-1M}  \\
    \bottomrule
  \end{tabular}
  \caption{MS module from \secref{subsec:module} based on NeuS on our synthesized dataset with circle paths.
  Parameters with the pattern ``A$|$B'' refer to grid parameters and the MLP parameters.}
  \label{tab:neus}
\end{table}

\begin{table}
  \centering
  \begin{subtable}[t]{\linewidth}
    \tablestyle{4.pt}{1}
    \begin{tabular}{cccc} \toprule
      Model (\# input images) & PSNR$\uparrow$ & SSIM$\uparrow$ & LPIPS$\downarrow$ \\
       \midrule
       Mip-NeRF 360 (100) & \underline{21.84}/\underline{40.17} & \underline{0.660}/\underline{0.980} & \underline{0.087} \\
       MS-Mip-NeRF 360 (100) & \textbf{32.11}/\textbf{43.04} & \textbf{0.913}/\textbf{0.984} & \textbf{0.051} \\
       \midrule
       Mip-NeRF 360 (75) & \underline{21.46}/\underline{39.37} & \underline{0.643}/\underline{0.978} & \underline{0.088} \\
       MS-Mip-NeRF 360 (75) & \textbf{31.44}/\textbf{42.20} & \textbf{0.890}/\textbf{0.982} & \textbf{0.055} \\
       \midrule
       Mip-NeRF 360 (30) & \underline{19.08}/\underline{34.51} & \underline{0.549}/\underline{0.966} & \underline{0.102} \\
       MS-Mip-NeRF 360 (30) & \textbf{28.37}/\textbf{39.84} & \textbf{0.854}/\textbf{0.979} & \textbf{0.060} \\
       \bottomrule
    \end{tabular}
    \caption{
      Circle camera path results.}
    \label{tab:circle_res} 
  \end{subtable}
  \\ \vspace{7pt}
  \begin{subtable}[t]{\linewidth}
    \tablestyle{4.pt}{1}
    \begin{tabular}{cccc} \toprule
      Model (\# input images) & PSNR$\uparrow$ & SSIM$\uparrow$ & LPIPS$\downarrow$ \\
       \midrule
       Mip-NeRF 360 (100) & \underline{22.26}/\underline{40.46} & \underline{0.711}/\underline{0.982} & \underline{0.083} \\
       MS-Mip-NeRF 360 (100) & \textbf{32.30}/\textbf{42.39} & \textbf{0.905}/\textbf{0.984} & \textbf{0.049} \\
       \midrule
       Mip-NeRF 360 (75) & \underline{21.69}/\underline{39.60} & \underline{0.692}/\underline{0.980} & \underline{0.084} \\
       MS-Mip-NeRF 360 (75) & \textbf{31.12}/\textbf{41.78} & \textbf{0.893}/\textbf{0.983} & \textbf{0.052} \\
       \midrule
       Mip-NeRF 360 (30) & \underline{18.86}/\underline{33.40} & \underline{0.596}/\underline{0.960} & \underline{0.116} \\
       MS-Mip-NeRF 360 (30) & \textbf{25.14}/\textbf{36.49} & \textbf{0.792}/\textbf{0.969} & \textbf{0.086} \\
       \bottomrule
    \end{tabular}
    \caption{
      Spiral camera path results.}
    \label{tab:spiral_res} 
  \end{subtable}
  \\ \vspace{7pt}
  \begin{subtable}[t]{\linewidth}
    \tablestyle{4.pt}{1}
    \begin{tabular}{cccc} \toprule
      Model (\# input images) & PSNR$\uparrow$ & SSIM$\uparrow$ & LPIPS$\downarrow$ \\
       \midrule
       Mip-NeRF 360 (100) & \underline{26.69}/\underline{35.63} & \underline{0.857}/\textbf{0.962} & \underline{0.112} \\
       MS-Mip-NeRF 360 (100) & \textbf{29.39}/\textbf{35.83} & \textbf{0.896}/\textbf{0.962} & \textbf{0.104} \\
       \midrule
       Mip-NeRF 360 (75) & \underline{25.25}/\textbf{34.93} & \underline{0.841}/\textbf{0.958} & \underline{0.118} \\
       MS-Mip-NeRF 360 (75) & \textbf{28.21}/\underline{34.60} & \textbf{0.883}/\underline{0.955} & \textbf{0.113} \\
       \midrule
       Mip-NeRF 360 (30) & \textbf{21.60}/\textbf{25.11} & \underline{0.766}/\textbf{0.889} & \textbf{0.237} \\
       MS-Mip-NeRF 360 (30) & \underline{21.58}/\underline{23.04} & \textbf{0.782}/\underline{0.864} & \underline{0.287} \\
       \bottomrule
    \end{tabular}
    \caption{
    Mirror-passing-through camera path results.}
    \label{tab:mt_res} 
  \end{subtable}
  \caption{
  Quantitative comparisons between MS-Mip-NeRF 360 and the baseline model on Scene01-Scene05 of all camera paths 
  from our synthesized dataset with varying input images. We also separately report the metrics as done in \tabref{tab:main_tab}.
  }
  \label{tab:n_input_images}
\end{table}

\subsection{Reconstructed geometry}
\label{subsec:geo}
Though NeuS~\cite{wang2021neus} integrates SDF fields with volumetric rendering for 3D reconstruction,
it inherits the weaknesses of density-based volumetric rendering on the reconstruction of mirror surfaces.
To validate the generalization of our scheme in solving multi-view inconsistency for novel view synthesis, 
we construct MS-NeuS by integrating our MS module from \secref{subsec:module} into NeuS implemented in \cite{instant-nsr-pl}
with the hyperparameters $\{K=4, d=16, h=32\}$.
We conduct experiments on our synthesized dataset with circle paths, 
and the quantitative and qualitative comparisons on rendered views are in \tabref{tab:neus} and \figref{fig:comp_neus}.
As shown in \figref{fig:comp_neus}, our scheme helps NeuS render better novel view images
and reconstructs better geometry for the mirrors,
but it still struggles to reconstruct clean geometry, 
because our MS module is designed for novel view synthesis
and there is no geometric constraints on sub-spaces.

\subsection{Rendering speed}
\label{subsec:fps}

\begin{table}[t]
  \centering
  \tablestyle{1pt}{1}
  \begin{tabular}{ccc}
    \toprule
      method  & training time$\downarrow$ & rendering time(per frame)$\downarrow$ \\
    \midrule
    Mip-NeRF & ~5.20 h & ~21 s \\
    $\textrm{MS-Mip-NeRF}_B$ & ~6.10 h & ~23 s \\
    \midrule
    Mip-NeRF 360 & ~12.56 h & ~32 s \\
    MS-Mip-NeRF 360 & ~13.30 h & ~39 s \\
    \midrule
    TensoRF & ~0.42 h & ~3.5 s \\
    MS-TensoRF & ~1.25 h & ~11.3 s \\
    \midrule
    iNGP & ~0.56 h & ~3.1 s \\
    MS-iNGP & ~0.67 h & ~3.6 s \\
    \bottomrule
  \end{tabular}
  \caption{
  The average training and rendering time comparisons on a single GeForce RTX 3090 GPU.}
  \label{tab:time}
\end{table}

Our scheme performs multiple volumetric rendering operations for each pixel,
therefore, the time consumption is related to the importance sampling strategy and the backbone networks.
We report the average training and rendering times in \tabref{tab:time}.
For pure MLP-based methods, the bottleneck of time consumption lies in the network inference,
and the number of sampled points for each ray is constant,
therefore, integrating our scheme has a relatively small inference on the speed.
TensoRF uses occupancy grid guided sampling strategy,
and in our multi-space version, we record all occupied positions across all sub-spaces into one occupancy grid,
therefore, the sampled points for each ray are much denser,
therefore, the time consumption increases by a large margin.
On the contrary, our iNGP version is constructed based on the NerfAcc framework 
with the proposal network-guided sampling strategy,
and the number of sampled points is constant,
therefore, the time consumption increases by less than one second.
For all our modules, there are redundant points along the rays,
and the possible solution is to sample points in different sub-spaces adaptively,
which we leave for future research.

%%%%%%%%%%%%%%%%%experiments%%%%%%%%%%%%%%%%%%

%%%%%%%%%%%%%%%%%discussion%%%%%%%%%%%%%%%%%%s
\section{Discussion}
\label{sec:dissc}

Along with our motivation in \secref{subsec:hyp} and extensive experiments in \secref{sec:exp},
we conduct more experimental analysis to investigate the mechanism of sub-space decomposition.
The experiments in \secref{subsec:ms-d} reveal that our multi-space scheme equips the density fields
with the ability to handle multi-view inconsistency without photometric supervision.
The experiments in \secref{subsec:loss} further demonstrate that our multi-space scheme successfully
decomposes multi-view inconsistent parts into different sub-spaces under the supervision of different losses.
The above experiments confirm that the multi-space decomposition is only determined by the virtual or real images,
and requires no additional regularizations.

\subsection{Multi-space scheme with density fields}
\label{subsec:ms-d}

\begin{figure}[t]
  \centering
  \begin{subfigure}{0.99\linewidth}
    \includegraphics[width=\linewidth]{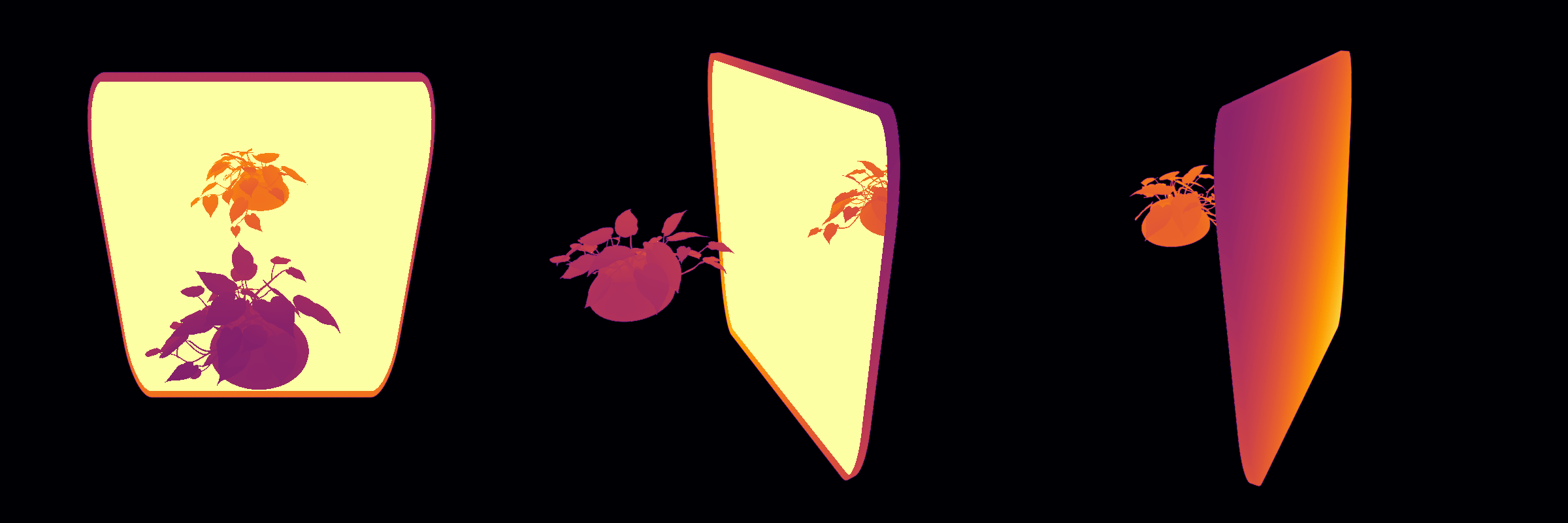}
    \caption{
    Ground-Truth rendered ray length maps.}
    \label{fig:ray_length_gt}
  \end{subfigure}
  \vspace{5pt}
  \vfill
  \begin{subfigure}{0.99\linewidth}
    \begin{overpic}[width=\linewidth]{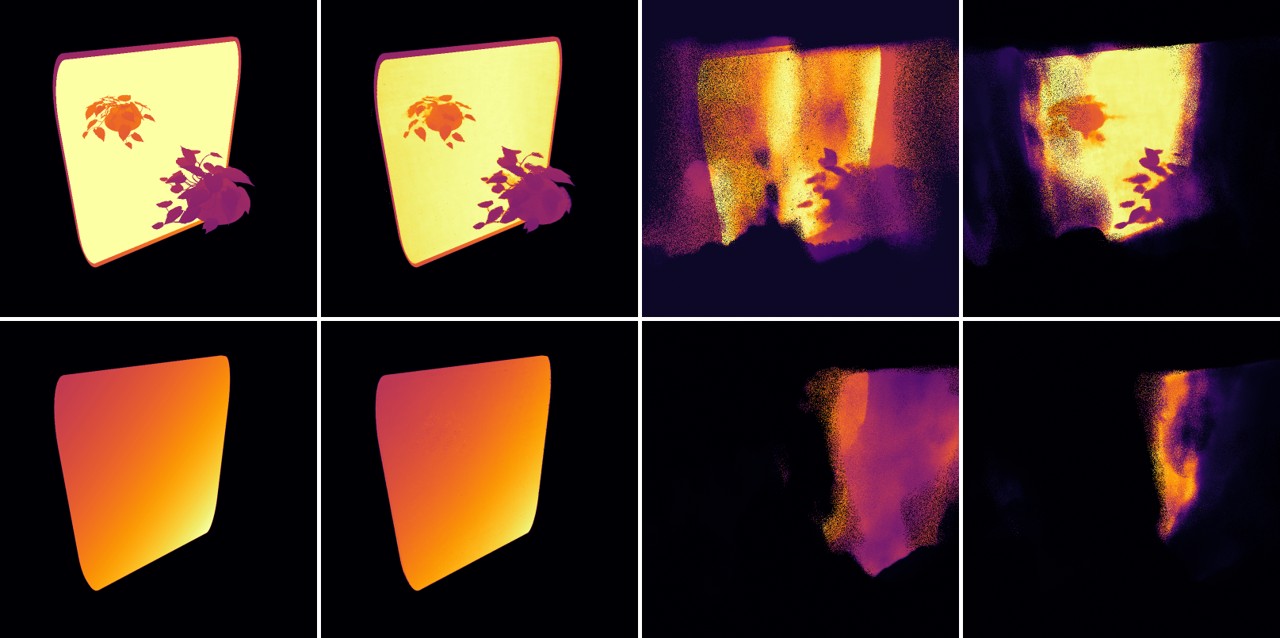} 
      \put(10, -5){\small 
      (i)}
      \put(35, -5){\small 
      (ii)}
      \put(60.2, -5){\small 
      (iii)}
      \put(84, -5){\small 
      (iv)}
    \end{overpic}
    \vspace{5pt}
    \caption{
    Visual comparison of rendered ray length maps.
    (i) GT; (ii) MS-NeRF-depth;
    (iii) NeRF-depth; (iv) NeRF-depth-v.}
    \label{fig:ray_length_vis}
  \end{subfigure}
  \vspace{5pt}
  \vfill
  \begin{subfigure}{0.99\linewidth} 
    \begin{overpic}[width=\linewidth]{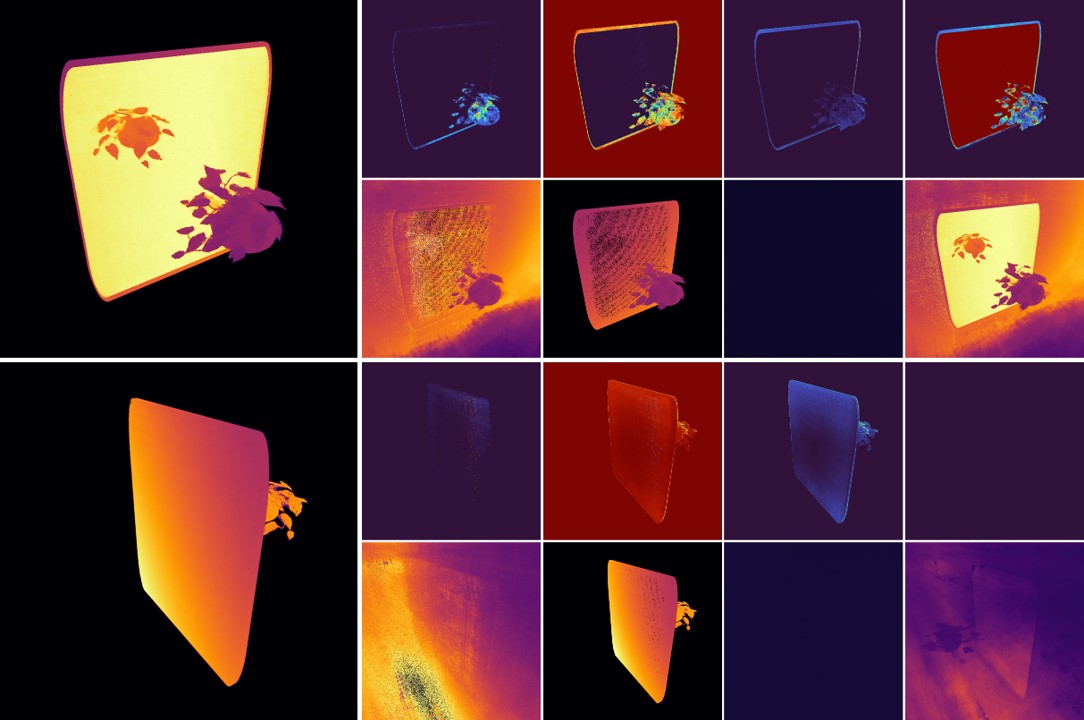}
      \put(0, -5){\small 
      (i) Full render result.}
      \put(34.5, -5){\small 
      (ii) Weight and ray length of sub-spaces.}
    \end{overpic}
    \vspace{5pt}
    \caption{
    Visualization of full rendered results and the decomposition results.}
    \label{fig:ray_length_decomp}
  \end{subfigure}
  \vspace{-5pt}
  \caption{
    Qualitative experiments of the multi-space scheme with density fields only.
  }
  \label{fig:ray_length}
\end{figure}

To investigate the mechanism of the multi-space scheme decomposing multi-view inconsistent space into sub-spaces, 
we conduct experiments on the density field without the radiance field head, 
which demonstrates clearer decomposition.
Specifically, we choose the original NeRF model with only the density output head, referred to as NeRF-depth,
therefore, the model is supervised by rendered depth maps via the rendering equation \equref{eq:intergel} as:
\begin{equation}\label{eq:intergel_d}
  \hat{\mathbf{D}}(\mathbf{r})=\sum_{i=1}^N T_i(1-{\rm exp}(-\sigma_i \delta_i))t_i    
\end{equation}
with $T_i = {\rm exp}(-\sum_{j=1}^{i-1}\sigma_j\delta_j)$,
$\delta_i = t_i - t_{i-1}$, and $\hat{\mathbf{D}}(\mathbf{r})$ is the rendered depth map.
We construct the MS-NeRF-depth model by integrating the multi-space module in \secref{subsec:module} into NeRF-depth
with hyperparameters $\{K=4, d=8, h=32\}$, and similarly only depth map is composed by:
\begin{equation}
  \hat{\mathbf{D}}(\mathbf{r}) = \frac{1}{\sum_{i=1}^K\mathrm{exp}(w^i)}\sum_{k=1}^{K}\mathrm{exp}(w^k)\hat{\mathbf{D}}^k.
  \label{eq:D_ours}
\end{equation}
where $w^k$ is the sub-space composition weights as in \equref{eq:F_k}, and $\hat{\mathbf{D}}^k$ is the sub-space depth map.
For fair comparison, we also build the NeRF-depth-v model based on NeRF-depth, which takes view directions as input before the output layer in addition to the positions because our module requires view directions as input.
All the models are supervised by the depth map, and we directly utilize the MSE loss.
We train these models on rendered ray length maps, where each pixel contains the physical transportation length of the ray 
traveling from the camera to the first non-reflective surfaces, following the circle camera path scene configuration as in \figref{fig:ray_length_gt}.

As in \figref{fig:ray_length_vis}(iii) and \figref{fig:ray_length_vis}(iv),
both the NeRF-depth and NeRF-depth-v fail to reconstruct the underlying density fields
due to the multi-view inconsistency of the ray transportation length 
caused by reflected rays.
On the contrary, our multi-space scheme successfully handles reflected rays 
as in \figref{fig:ray_length_vis}(ii), and the visualization in \figref{fig:ray_length_decomp} proves that successful handling is accomplished by
automatically separating the virtual images from the real ones.
The experiments confirm that our scheme can handle multi-view inconsistency
not only in the appearance domain but also in the geometry domain, and the sub-space decomposition 
follows the principle of separating multi-view inconsistent parts to different sub-spaces where multi-view inconsistency is well-preserved.

\begin{figure*}
  \begin{subfigure}{0.49\linewidth}
    \includegraphics[width=\linewidth]{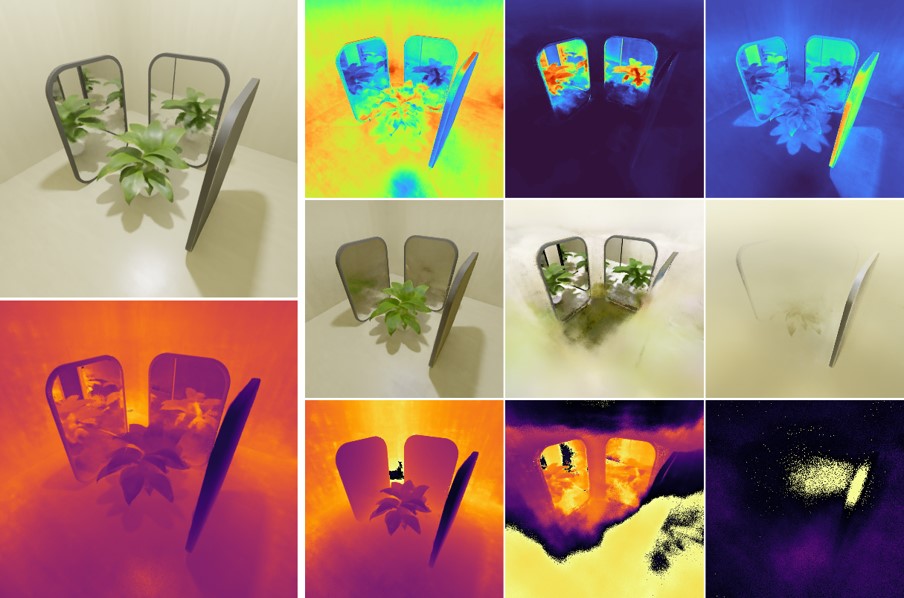}
    \caption{
    $\textrm{MS-NeRF}_B$ w/ loss (a)}
  \end{subfigure}
  \hfill
  \begin{subfigure}{0.49\linewidth}
    \includegraphics[width=\linewidth]{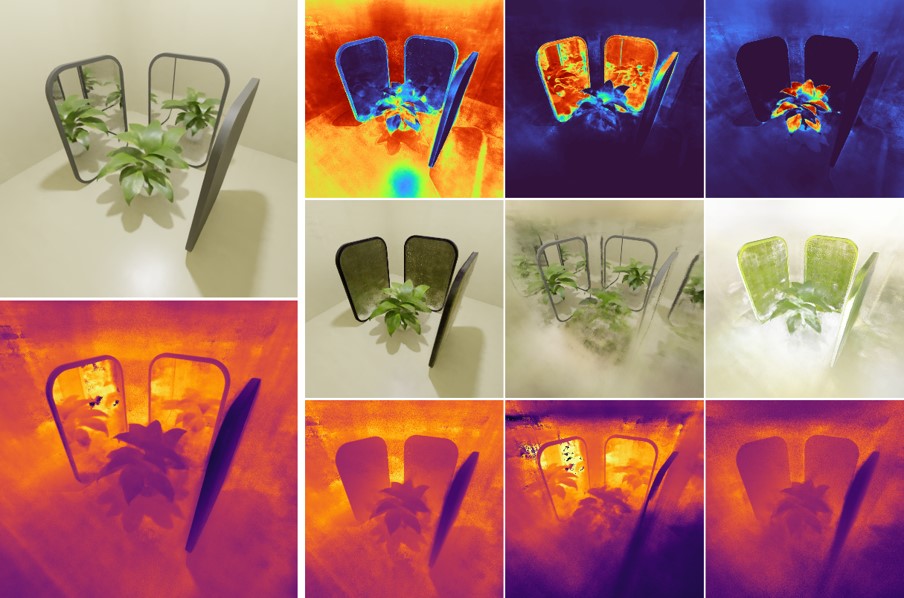}
    \caption{
    $\textrm{MS-NeRF}_B$ w/ loss (b)}
  \end{subfigure}
  \vfill
  \begin{subfigure}{0.49\linewidth}
    \includegraphics[width=\linewidth]{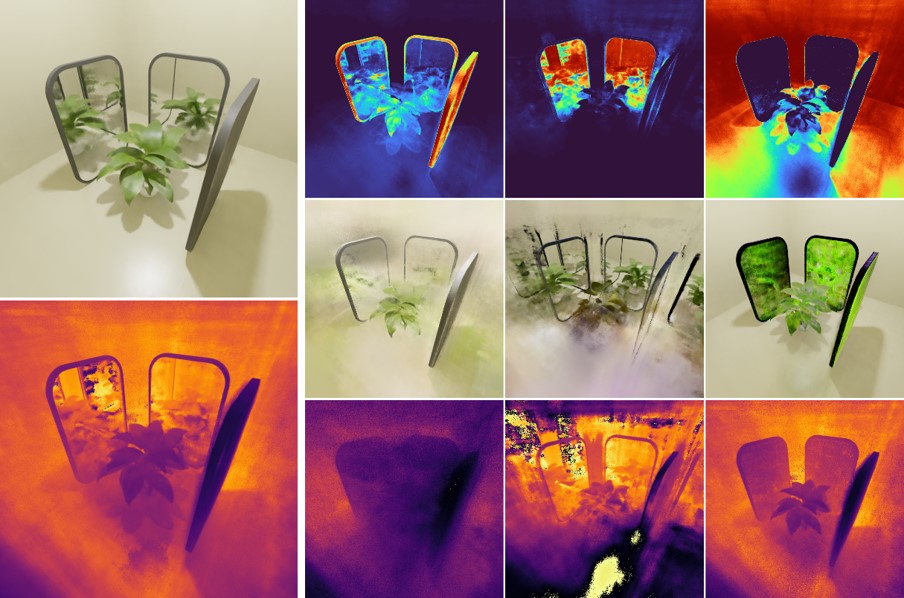}
    \caption{
    $\textrm{MS-NeRF}_B$ w/ loss (c)}
  \end{subfigure}
  \hfill
  \begin{subfigure}{0.49\linewidth}
    \includegraphics[width=\linewidth]{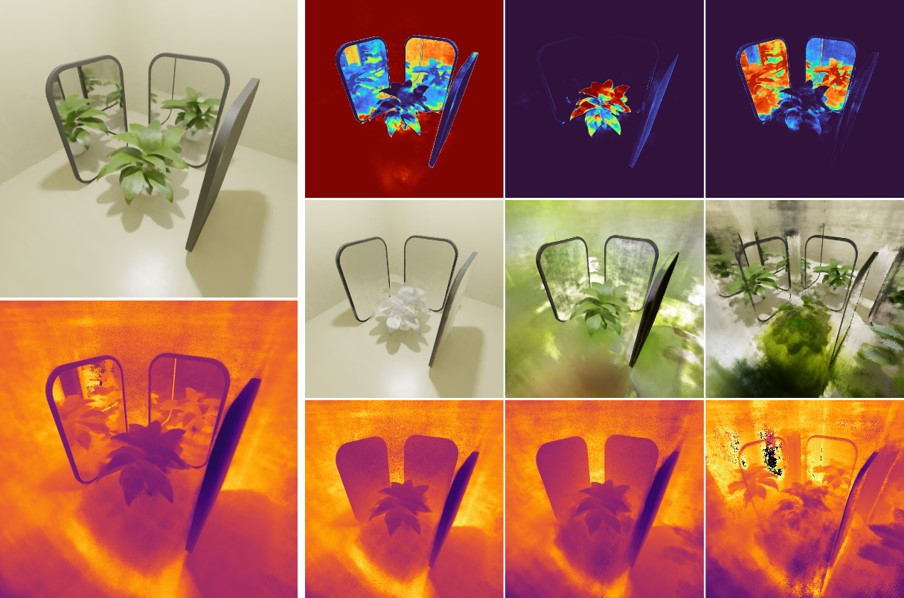}
    \caption{
    $\textrm{MS-NeRF}_B$ w/ loss (d)}
  \end{subfigure}
  \vspace{-5pt}
  \caption{
    Visualization of the sub-space decomposition from $\textrm{MS-NeRF}_B$ supervised by different losses.
    The left image contains the final rendered RGB and depth map, and the right one contains weight maps, RGB images, and depth maps of sub-spaces.
  }\vspace{-5pt}
  \label{fig:qual_diff_loss}
\end{figure*}

\begin{table}[t]
  \centering
  \begin{tabular}{cccc} \toprule
    & PSNR$\uparrow$ & SSIM$\uparrow$ & LPIPS$\downarrow$\\
    \midrule
      NeRF w/ loss (b)           & 22.00/38.88 & 0.638/0.972 & 0.109 \\
    \midrule
    $\textrm{MS-NeRF}_B$ w/ loss (a)  & 24.31/39.24 & 0.737/\textbf{0.976} & 0.093 \\
    $\textrm{MS-NeRF}_B$ w/ loss (b)  & \textbf{28.09}/\underline{39.94} & \textbf{0.831}/\underline{0.974} & \textbf{0.075} \\
    $\textrm{MS-NeRF}_B$ w/ loss (c)  & \underline{27.00}/\textbf{38.97} & \underline{0.803}/0.971 & \underline{0.082} \\
    $\textrm{MS-NeRF}_B$ w/ loss (d)  & 26.06/38.35 & 0.763/0.969 & 0.096 \\
    \bottomrule
  \end{tabular}
  \caption{
  Quantitative evaluation with different photometric losses. The training loss combinations are
  (a) $\mathcal{L}_{MAE}$; (b) $\mathcal{L}_{MSE}$; 
  (c) $0.5\times \mathcal{L}_{MSE} + 0.5\times \mathcal{L}_{SSIM}$; and (d) $0.9\times \mathcal{L}_{MSE} + 0.1\times \mathcal{L}_{LPIPS}$.}
  \label{tab:quant_diff_loss}
\end{table}

\subsection{Multi-space scheme with different supervision}
\label{subsec:loss}

To investigate the relation between sub-space decomposition and different supervision losses,
we choose the commonly used photometric losses, including Mean Absolute Error (MAE) loss $\mathcal{L}_{MAE}$, 
Mean Squared Error (MSE) loss $\mathcal{L}_{MSE}$, 
Structural Similarity Index Measure (SSIM) loss $\mathcal{L}_{SSIM}$, and 
Learned Perceptual Image Patch Similarity (LPIPS) loss $\mathcal{L}_{LPIPS}$,
to supervise the $\textrm{MS-NeRF}_B$ model on Scene01-Scene05 with circle paths from our synthesized dataset.

We provide the quantitative evaluation in \tabref{tab:quant_diff_loss}, which demonstrates that
our multi-space scheme is compatible with different supervision losses.
Besides, the visualization of decomposition in \figref{fig:qual_diff_loss} confirms our motivation from \secref{subsec:hyp} that
the sub-space decomposition is only determined by whether the images are real or virtual.

%%%%%%%%%%%%%%%%%discussion%%%%%%%%%%%%%%%%%%

%%%%%%%%%%%%%%%%%conclusion%%%%%%%%%%%%%%%%%%

\section{Conclusion}
\label{sec:conclusion}

In this paper, we tackle the long-standing problem of rendering reflective surfaces in NeRF-based methods.
We introduce a multi-space NeRF method that decomposes the Euclidean space into multiple virtual sub-spaces.
Our proposed MS-NeRF approach achieves significantly better results compared with conventional NeRF-based methods.
Moreover, a light-weighted design of the MS module allows our approach to serve as an enhancement to the conventional NeRF-based methods. 
We also constructed a novel dataset for the evaluation of similar tasks, hopefully helping future research in the community. 

%%%%%%%%%%%%%%%%%conclusion%%%%%%%%%%%%%%%%%%

%%%%%%%%%%%%%%%%%acknowledgement%%%%%%%%%%%%%%%%%%

%%
\myPara{Acknowledgement} This work is supported by the National Natural Science Foundation of China (62132012),
the Fundamental Research Funds for the Central Universities (Nankai University, No. 63233080), and the Tianjin
science and technology projects (22JCYBJC01270). Computation is supported by the Supercomputing Center of Nankai University (NKCS).

%%%%%%%%%%%%%%%%%acknowledgement%%%%%%%%%%%%%%%%%%

\ifCLASSOPTIONcaptionsoff
  \newpage
\fi

\bibliographystyle{IEEEtran}
\bibliography{main}

\vspace{-.4in}
\begin{IEEEbiography}[{\includegraphics[width=1in]{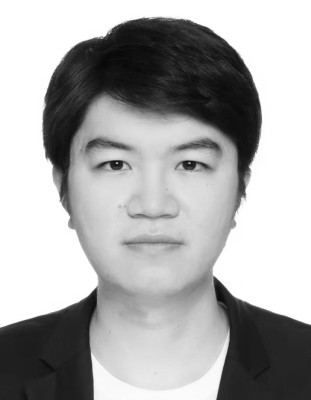}}]
{Ze-Xin Yin} received the master`s degree under the supervision of 
Prof. Ming-Ming Cheng and Asst. Prof. Bo Ren. 
He is currently working toward the PhD degree under the supervision of 
Prof. Jin Xie 
with the College of Computer Science, Nankai University. 
His research interests mainly focus on neural radiance fields and 3D computer vision.
\end{IEEEbiography}

\vspace{-.4in}
\begin{IEEEbiography}[{\includegraphics[width=1in]{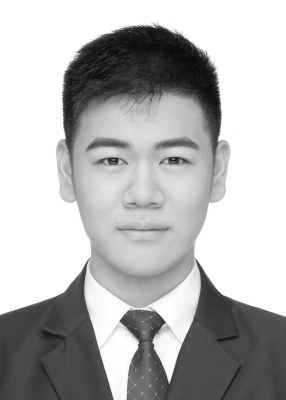}}]
{Peng-Yi Jiao} received the bachelor`s degree from 
the Beijing Institute of Technology, in 2019. 
He is currently working toward the master`s degree 
with the College of Computer Science, Nankai University. 
His research interests include neural radiance fields, computer graphics, and computer vision.
\end{IEEEbiography}

\vspace{-.4in}
\begin{IEEEbiography}[{\includegraphics[width=1in]{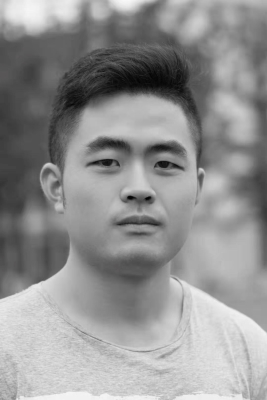}}]
{Jiaxiong Qiu} received the bachelor`s degree 
from Dalian Maritime University, in 2017, 
the master degree from the University of Electronic Science 
and Technology of China,  in 2020, 
and the PhD degree from the College of Computer Science, 
Nankai University, in 2024. 
His research interests include computer vision, computer graphics, robotics, and deep learning.
\end{IEEEbiography}

\vspace{-.4in}
\begin{IEEEbiography}[{\includegraphics[width=1in]{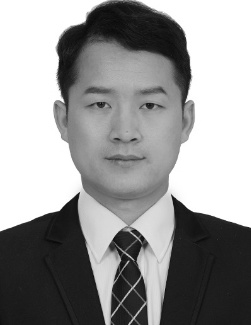}}]
{Ming-Ming Cheng} received the PhD degree from Tsinghua University in 2012.
He is currently a research fellow in The University of Oxford,
working with Prof. Philip Torr. His research interests includes computer graphics,
computer vision, image processing, and image retrieval.
He has received the Google PhD fellowship award, the IBM PhD fellowship award, and the “New
PhD Researcher Award” from Chinese Ministry of Education.
\end{IEEEbiography}

\vspace{-.4in}
\begin{IEEEbiography}[{\includegraphics[width=1in]{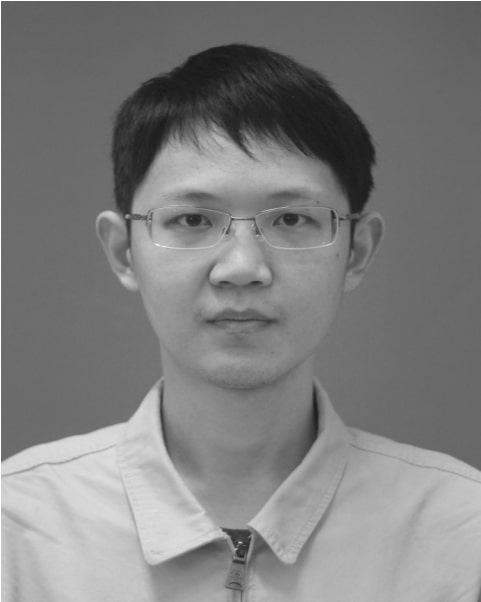}}]
{Bo Ren} received the B.S. and Ph.D. degrees from Tsinghua University in 2010 and 2015 respectively.
He is currently an associate professor at College of Computer Science, Nankai University.
His research interests lies in computer graphics, computer vision and artificial intelligence. 
Current researches involve learning-based/physically-based simulation, 3D scene geometry reconstruction and analysis.
\end{IEEEbiography}
\vspace{-.45in}

\vfill

\end{document}